%% file: _main.tex
\input{_constants}
\arxiv
\pdfoutput=1
\documentclass[10pt,twocolumn,letterpaper]{article}
\input{cvpr_header}

\begin{document}
\title{Automatic Controllable Colorization via Imagination}
\author{Xiaoyan Cong$^3$ \quad Yue Wu$^4$ \quad    Qifeng Chen$^2$ \quad  Chenyang Lei$^1$\thanks{Corresponding author}\\
$^1$CAIR, HKISI-CAS  \qquad $^2$HKUST  \qquad   $^3$Zhejiang University \qquad $^4$Huawei Noah’s Ark Lab \\
}

\newcommand{\yue}[1]{\textcolor{red}{[Yue: #1]}}

\maketitle
\input{00_abstract}

\input{01_intro}
\input{02_related}

\input{04_method}
\input{05_experiment}

\input{10_conclusion}


{\small
\bibliographystyle{ieeenat_fullname}
\bibliography{11_references}
}


\end{document}


\title{Automatic Controllable Colorization via Imagination}
\author{\authorBlock}
\maketitlesupplementary

\input{12_appendix}

{\small
\bibliographystyle{ieee_fullname}
\bibliography{11_references}
}

%% file: _constants.tex
\def\paperID{7681}
\def\confName{CVPR}
\def\confYear{2024}

\newif\ifreview 
\newif\ifarxiv \newcommand{\arxiv}{\arxivtrue}
\newif\ifcamera 
\newif\ifrebuttal 

%% file: cvpr_header.tex
\ifreview \usepackage[review]{cvpr} \fi
\ifarxiv \usepackage[pagenumbers]{cvpr} \fi
\ifrebuttal \usepackage[rebuttal]{cvpr} \fi
\ifcamera \usepackage{cvpr} \fi

\input{_macros}  

\usepackage{xr-hyper}

\makeatletter
\newcommand*{\addFileDependency}[1]{
  \typeout{(#1)}
  \@addtofilelist{#1}
  \IfFileExists{#1}{}{\typeout{No file #1.}}
}

\makeatother

\definecolor{cvprblue}{rgb}{0.21,0.49,0.74}
\usepackage[pagebackref,breaklinks,colorlinks,citecolor=cvprblue]{hyperref}
\usepackage[capitalize]{cleveref}
\crefname{section}{Sec.}{Secs.}
\crefname{table}{Table}{Tables}
\crefname{figure}{Fig.}{Figs.}

%% file: _macros.tex
\usepackage{graphicx}	
\usepackage{amsmath}	
\usepackage{amssymb}	
\usepackage{booktabs}
\usepackage{times}
\usepackage{microtype}
\usepackage{epsfig}
\usepackage{tabularray}
\usepackage[table,xcdraw,dvipsnames]{xcolor}
\usepackage{caption}
\usepackage{float}
\usepackage{placeins}
\usepackage{color, colortbl}
\usepackage{stfloats}
\usepackage{enumitem}
\usepackage{tabularx}
\usepackage{xstring}
\usepackage{multirow}
\usepackage{makecell}
\usepackage{xspace}
\usepackage{url}
\usepackage{subcaption}
\usepackage{xcolor}
\usepackage[hang,flushmargin]{footmisc}
\usepackage{overpic}
\usepackage{algorithm}
\usepackage{algpseudocode}
\usepackage[accsupp]{axessibility}  

\usepackage{diagbox}
\usepackage{pifont}
\usepackage{graphbox} 
\usepackage{colortbl}
\usepackage{xcolor}
\usepackage{collcell}
\definecolor{mygray}{gray}{.9}

\usepackage{scalefnt}

\ifcamera \usepackage[accsupp]{axessibility} \fi

% ...

\ifarxiv  \fi

\newcommand{\R}[1]{{%
    \textbf{%
        \ifstrequal{#1}{1}{\textcolor{red}{R#1}}{%
        \ifstrequal{#1}{2}{\textcolor{blue}{R#1}}{%
        \ifstrequal{#1}{3}{\textcolor{magenta}{R#1}}{%
        \ifstrequal{#1}{4}{\textcolor{teal}{R#1}}{%
                           \textcolor{cyan}{R#1}%
        }}}}%
    }%
}}

%% file: 00_abstract.tex
\begin{abstract}

We propose a framework for automatic colorization that allows for iterative editing and modifications. The core of our framework lies in an imagination module: by understanding the content within a grayscale image, we utilize a pre-trained image generation model to generate multiple images that contain the same content. These images serve as references for coloring, mimicking the process of human experts. As the synthesized images can be imperfect or different from the original grayscale image, we propose a Reference Refinement Module to select the optimal reference composition. Unlike most previous end-to-end automatic colorization algorithms, our framework allows for iterative and localized modifications of the colorization results because we explicitly model the coloring samples. Extensive experiments demonstrate the superiority of our framework over existing automatic colorization algorithms in editability and flexibility. Project page: \href{https://xy-cong.github.io/imagine-colorization/}{https://xy-cong.github.io/imagine-colorization/}.

\end{abstract}


%% file: 01_intro.tex
\vspace{0em}
\section{Introduction}
\label{sec:intro}

Countless black-and-white photographs, relics of bygone eras, exist today, often requiring the deft touch of skilled artists to breathe color into them. However, such a manual process is labor-intensive and time-consuming. Given a grayscale image, colorization methods aim to estimate the corresponding color automatically or with other guidance. Achieving expert-level colorization is challenging due to the multi-modality nature of the task: there exist multiple plausible colorization results for one grayscale image.

Automatic colorization algorithms~\cite{Su-CVPR-2020, CT2, zhang2016colorful, zhang2017real, Kim2022BigColor, wu2021towards, huang2022unicolor, XiaHWW22} have made significant progress in recent years and have demonstrated impressive results in many common scenarios. However, they are still hard to match the colorization proficiency of human experts. Many data-driven methods struggle with issues such as grayish and desaturation because of the color ambiguity that one object may present multiple plausible colors in datasets. GAN-based approaches tend to produce unpleasant artifacts due to the limited representation capacity in GAN's prior. DDColor~\cite{kang2022ddcolor} alleviates these shortcomings to some extent but is unable to achieve controllable and diverse colorization. Human experts excel in colorizing black-and-white photos due to their rich and imaginative priors, enabling them to transform grayscale images into visually pleasant renditions. They possess instance-aware and semantic-aware understanding both locally and globally, imagining diverse plausible colors and assigning the most appropriate ones based on context. Exemplar-based colorization methods~\cite{yin2021yes, li2022eliminating, bai2022semantic, chen2023spcolor}  achieve pleasing results through emulating the human experts' priors, wherein a highly-corresponding reference is manually provided. However, identifying a perfect reference can be both labor-intensive and detract from an optimal user experience.



Another challenge with most automatic algorithms is that the obtained results are often difficult to modify and improve. For example, Su et al.~\cite{Su-CVPR-2020} and Zhang et al.~\cite{zhang2016colorful} can only produce one deterministic result. While UniColor~\cite{huang2022unicolor} offers an interactive colorization, it demands users to provide additional information, such as text or strokes.

\begin{figure}
  \centering
  \includegraphics[width=\linewidth]{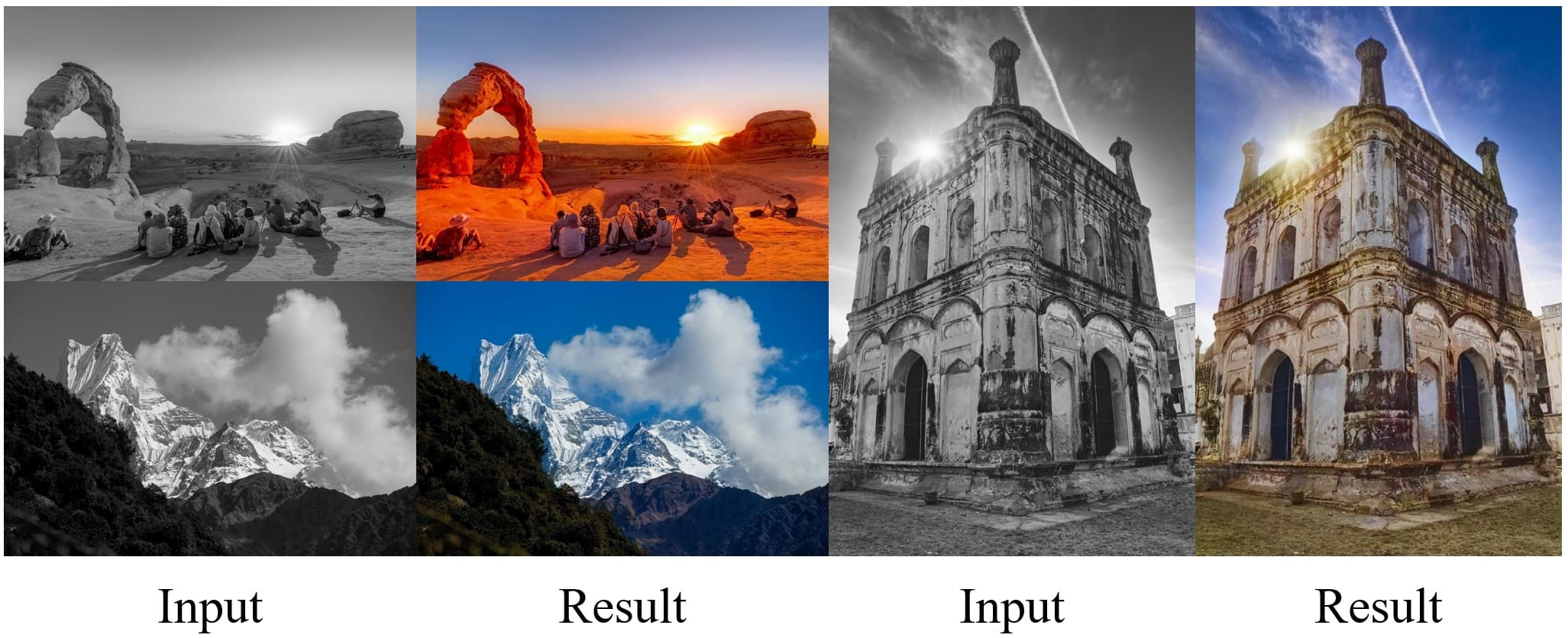}
\vspace{-0.2in}
\caption{\textbf{Our colorization results.} By leveraging our designed Imagination Module, our framework can achieve photorealistic and vivid colorization results.}
\vspace{-0.1in}
\label{fig:teaser}
\end{figure}

\begin{figure}
  \centering
  \includegraphics[width=\linewidth]{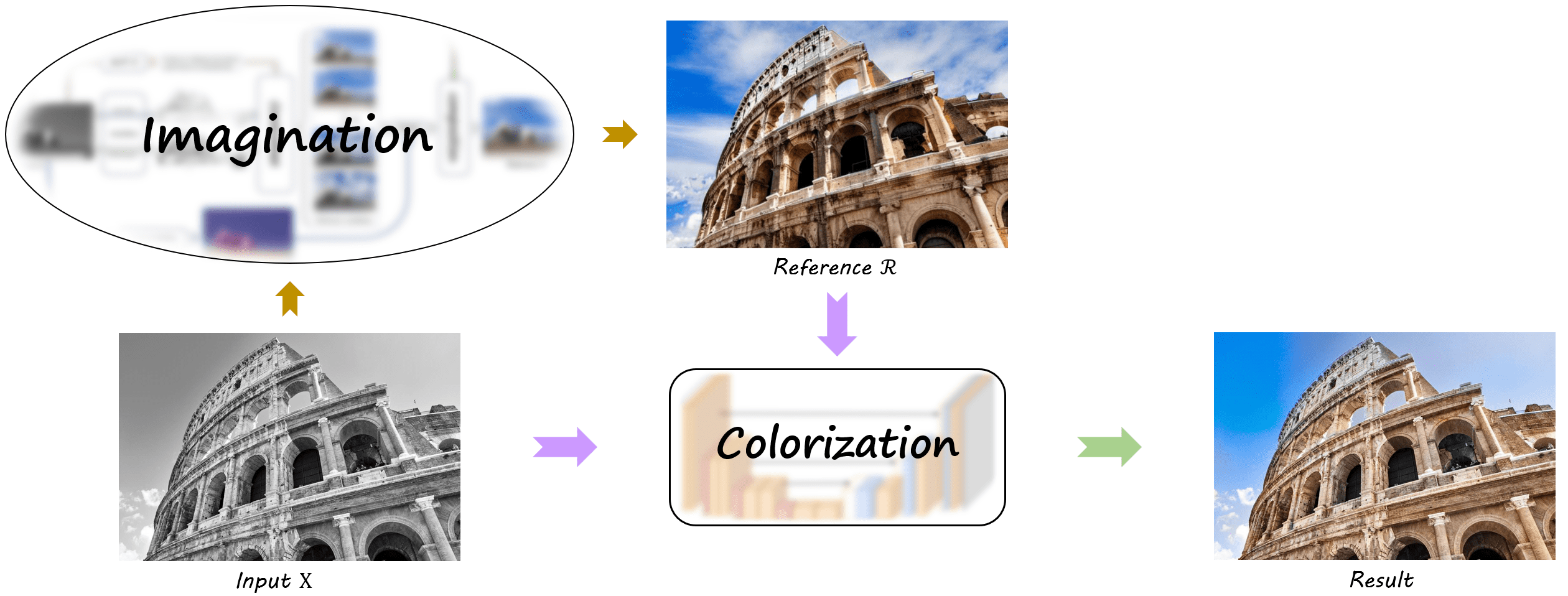}
\vspace{-0.2in}
\caption{\textbf{Framework Overview.} Given a black-and-white input, our framework first synthesizes a semantically similar, spatially aligned, and instance-aware reference by mimicking the imagination process of human experts. Then the colorization module colorizes the black-and-white image with the guidance of reference.}
\label{fig:Framework}
\end{figure}

In this paper, we propose a novel automatic colorization framework that allows iterative editing and modifications from users. Our framework is designed to mimic the process of human expert imagination colorization, as shown in Figure \ref{fig:Framework}. We introduce an imagination module designed to provide semantically similar, structurally aligned, and instance-aware references, mirroring human imagination. Our framework begins by composing a refined reference from multiple candidates generated by pretrained cross-modality image generation models, such as ControlNet~\cite{zhang2023adding}. The rich color priors embedded in large pretrained Diffusion models can be harnessed to synthesize references with diverse and colorful colors. Then, our colorization module colorizes the black-and-white input under the guidance of the reference.

Our composition strategy involves two steps: First, we carry out the semantic segmentation of the input. Second, we compose the refined reference by selecting the segment most similar to all potential reference candidates. The composition strategy enables our framework to exhibit diverse colorization results, as well as provides the flexibility for controllable and editable colorization tailored to the unique desires and preferences of the user. Specifically, if users find certain segments unsatisfactory, they can effortlessly substitute those segments with alternatives from the pool of reference candidates or use an image of their choosing. Consequently, the colorization results are accordingly modified to reflect these changes.

\begin{figure*}[t]
  \centering
  \includegraphics[width=\linewidth]{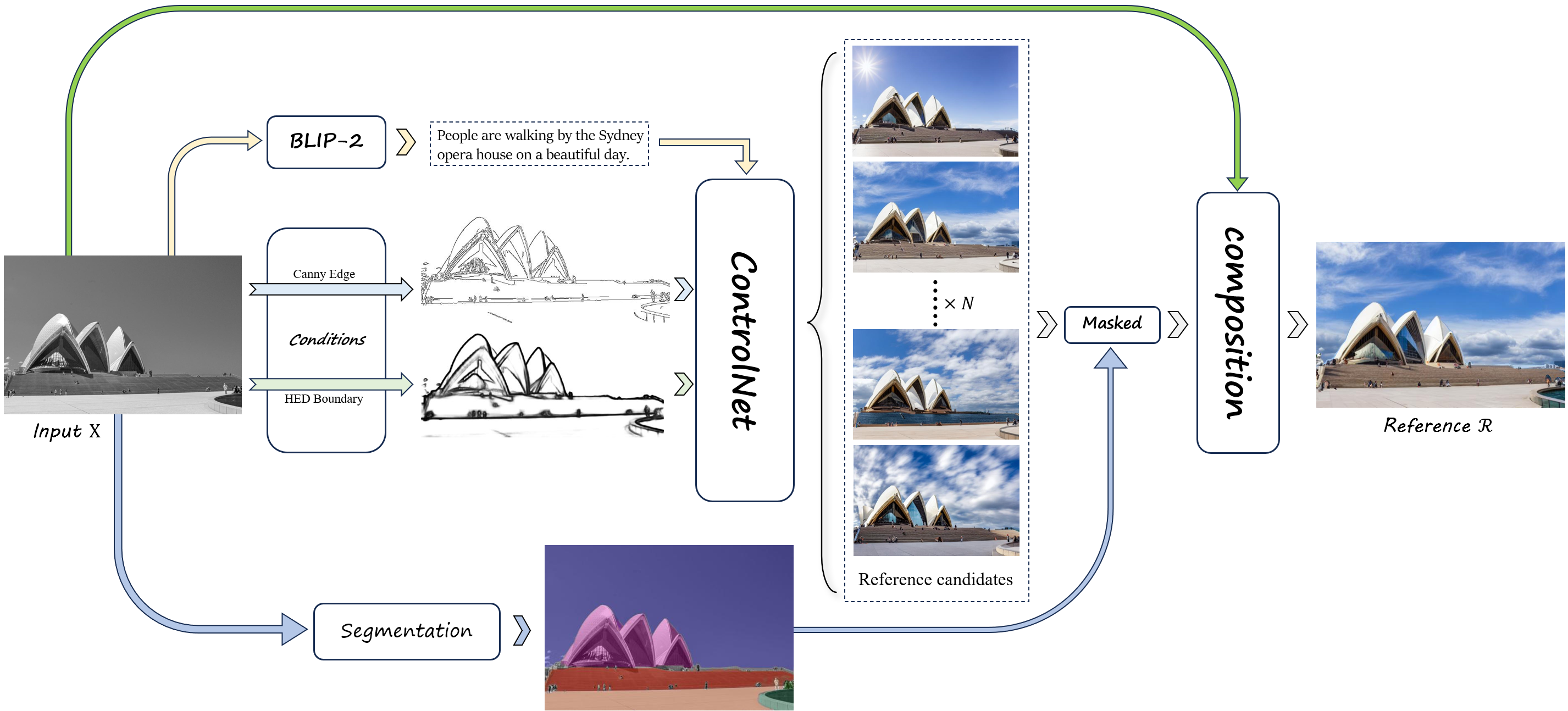}
\caption{\textbf{Imagination Module and Reference Refinement Module.} In Imagination Module \ref{subsec:Imagination-Module}, given a grayscale input, we generate $N$ reference candidates $\mathbf{C}$, $N_1$ of which are conditioned on the canny edge of $\mathbf{X}$, and $N_2$ of which are conditioned on the HED boundary of $\mathbf{X}$, $N = N_1 + N_2$. In Reference Refinement Module \ref{subsec:Reference-Refinement-Module}, we first extract the segmentation $\mathbf{S}$ of $\mathbf{X}$. For each segment $\mathbf{S}^j$, we look for the best reference segment $\mathbf{R}\, \odot\, \mathbf{S}^j$ for the optimal reference $\mathbf{R}$ by selecting the nearest neighbor for $\mathbf{X} \, \odot \, \mathbf{S}^j$ among all reference candidates $\mathbf{C}_i\, \odot\, \mathbf{S}^j$ in terms of the difference in the robust and universe  DINOv2~\cite{oquab2023dinov2} feature space.}
\vspace{-0.1in}
\label{fig:Reference_synthesis}
\end{figure*}

Our contributions can be summarized as follows:\begin{itemize}
\setlength{\itemsep}{0pt}
\setlength{\parsep}{0pt}
\setlength{\parskip}{0pt}
\item We propose a novel automatic colorization framework that leverages the pre-trained diffusion models. We introduce an imagination module that emulates human experts to synthesize semantically similar, structurally aligned, and instance-aware colorful references, with potential applications beyond colorization.
\item We demonstrate our novel automatic colorization framework exhibits remarkable controllable and user-interactive capabilities. We can also present diverse colorization results. 
\item Compared to previous automatic colorization methods, our framework achieves state-of-the-art performance and generalization. 

\end{itemize}

%% file: 02_related.tex
\section{Related Work}
\label{sec:related}

\noindent \textbf{Automatic Colorization.}
Deep learning based methods~\cite{zhang2019deepcolor, deoldify2019, Su-CVPR-2020, CT2, XiaHWW22, wu2021towards, zhang2016colorful, Larsson2016Learning, vitoria2020chromagan, guadarrama2017pixcolor, zhao2019pixelated, iizuka2016let, huang2022unicolor, kumar2021colorization, kang2022ddcolor,Lei_2019_CVPR} achieve automatic colorization by learning the relationship between semantic and color on large datasets. Additional semantic information is incorporated to improve colorfulness, such as classification~\cite{zhang2016colorful, zhang2017real, Larsson2016Learning, iizuka2016let}, semantic segmentation~\cite{guadarrama2017pixcolor, zhao2019pixelated} and instance awareness~\cite{Su-CVPR-2020}. The performance of the above methods is usually good when the color of the object (e.g., a green tree) is closely related to the semantic, while they tend to produce muted or grayish results for objects with a wide spectrum of potential colors (e.g., T-shirt).  In contrast, our framework can alleviate the limitation and generate more diverse and colorful results thanks to our innovative imagination module.

\begin{figure}[t]
\centering
\begin{tabular}{@{}c@{\hspace{1mm}}c@{\hspace{0mm}}c@{\hspace{0mm}}c@{}}

\rotatebox{90}{\small \hspace{8.5mm} \scriptsize{Input} }&
\includegraphics[width=0.3\linewidth, height=0.3\linewidth]{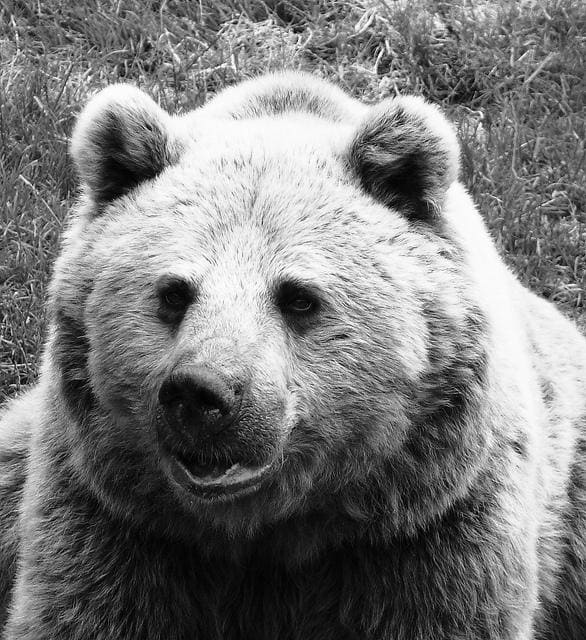}&
\includegraphics[width=0.3\linewidth, height=0.3\linewidth]{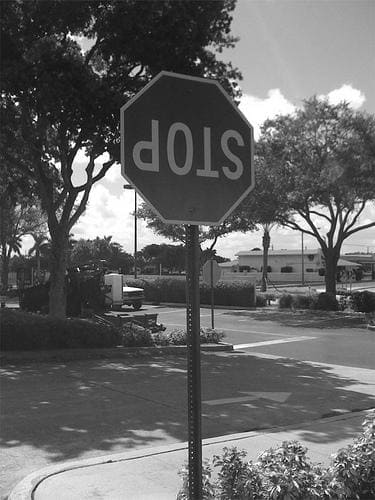}&
\includegraphics[width=0.3\linewidth, height=0.3\linewidth]{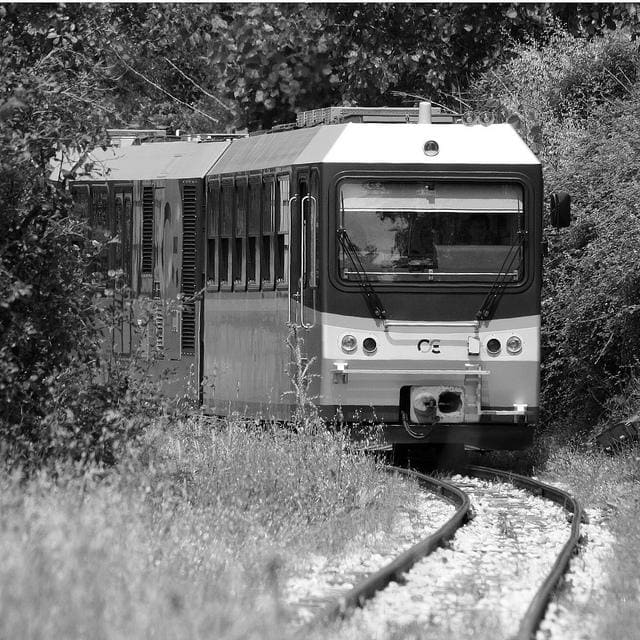}\\

\rotatebox{90}{\small \hspace{3mm} \scriptsize{ControlNet~\cite{zhang2023adding}}}&
\includegraphics[width=0.3\linewidth, height=0.3\linewidth]{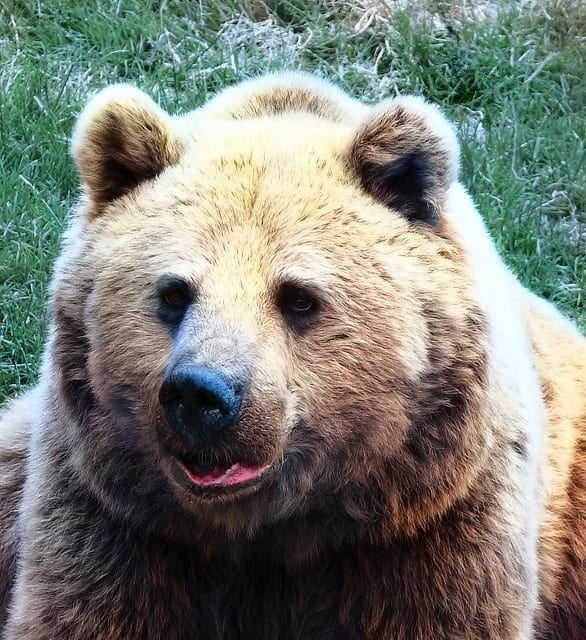}&
\includegraphics[width=0.3\linewidth, height=0.3\linewidth]{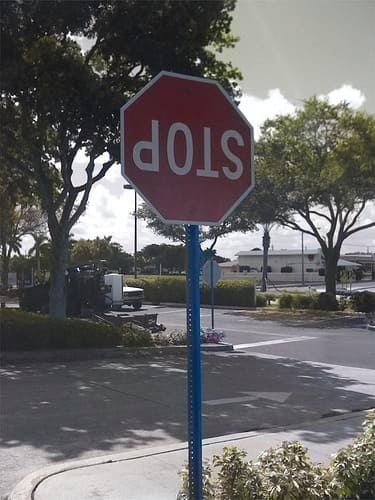}&
\includegraphics[width=0.3\linewidth, height=0.3\linewidth]{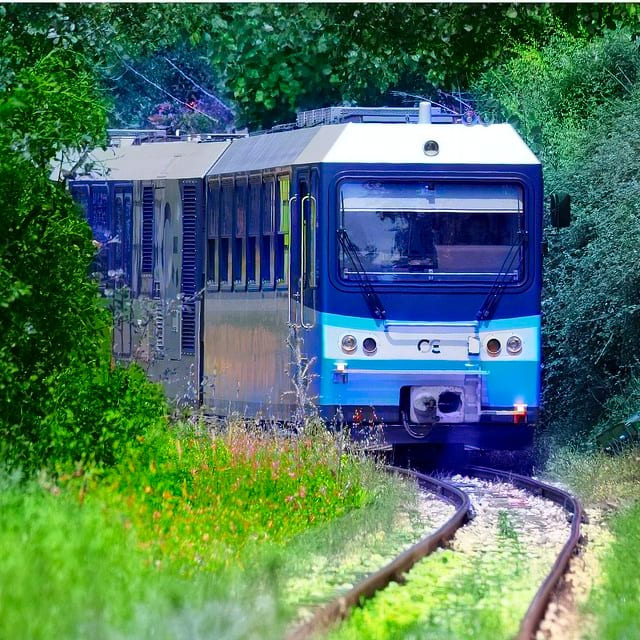}\\


& \multicolumn{1}{c}{\footnotesize Inconsistency} & \multicolumn{1}{c}{\footnotesize Grayish} & \multicolumn{1}{c}{\footnotesize Color bleeding} \\
\end{tabular}
\\
\vspace{-0.1in}
\caption{\textbf{ControlNet Issues.} Directly applying ControlNet\cite{zhang2023adding} to colorization might cause issues like inconsistency, grayish, and color bleeding. We use gray images as conditions of ControlNet and corresponding color images as ground truth. A pretrained stabel diffusion model is used as the initialization. We then train the model on the commonly used colorization datasets.}

\label{fig:controlnet_failure}
\vspace{-0.1in}
\end{figure}

\noindent \textbf{User-guided Colorization.}
Most classical approaches rely on user guidance for colorization, including scribble-based colorization, text-based colorization, and example-based colorization. Scribble-based methods~\cite{ColorizationOptimization, MangaColorization, luan2007natural, sangkloy2017scribbler, zhang2018two, Filling2021zhang} usually formulate colorization as an optimization problem with some assumptions. For example, Levin et al.~\cite{ColorizationOptimization} assume pixels with similar intensity share similar colors. Some follow-up methods work on achieving more efficient color propagation by leveraging texture similarity~\cite{MangaColorization, luan2007natural} and intrinsic distance~\cite{yatziv2006fast}. These approaches often require highly labor-intensive manual processes. Text-based methods~\cite{manjunatha2018learning, xie2018language, bahng2018coloring, Weng_2022_AAAI, lcoder, lcoins,chang2023lcad} offer a reprieve from this obstacle by leveraging user-friendly language descriptions delineating objects and their corresponding colors to guild colorization. Some recent approaches aim to enhance the robustness and consistency of colorization by the grouping mechanism~\cite{lcoins} aggregating similar image patches and assigning vivid colors for unmentioned instances~\cite{chang2023lcad}.

Example-based methods~\cite{Transferring2002, chia2011semantic, ironi2005colorization, tai2005local, Intrinsic2008, charpiat2008automatic, chia2011semantic, gupta2012image, bugeau2013variational, he2018deep, zhang2019deepcolor, xu2020stylization, huang2020learning, Gray2ColorNet2020, SuperpixelVar2020, zhao2021color2embed, li2021globally, yin2021yes, li2022eliminating, bai2022semantic, chen2023spcolor} further reduce the difficulty of colorization for normal users. Only one reference image is required for automatic colorization. Some approaches~\cite{xu2020stylization, zhao2021color2embed, huang2020learning, bai2022semantic} focus on leveraging AdaIN~\cite{huang2017arbitrary} to transfer the global color distribution of reference image, while others~\cite{tai2005local, ironi2005colorization, chia2011semantic, bugeau2013variational, he2018deep, zhang2019deepcolor, li2021globally, bai2022semantic, chen2023spcolor} transfer color according to pixel-level semantic correspondence. Although these methods have improved colorization performance a lot, it is still non-trivial, time-consuming, and not user-friendly for users to find a perfect reference by searching on the Internet or providing descriptions and segmentation cues to an image retrieval system~\cite{chia2011semantic}. In our work, the imagination module avoids such efforts by automatically synthesizing a semantically similar and structurally aligned colorful reference image.

\noindent \textbf{Generative Priors Colorization.}
Generative priors embedded in pretrained GANs and Diffusions have been pivotal in achieving photorealistic colorization. GANs~\cite{goodfellow2014generative, wang2018high, sun2019image, li2019diverse, park2019semantic, karras2019style, karras2020analyzing, sushko2020you, sun2021learning} are a huge family of generative models for conditional and unconditional image synthesis. Some specific conditions are utilized to control better image synthesis, such as layout~\cite{sun2019image, sun2021learning} and semantic map~\cite{li2019diverse, wang2018high, park2019semantic, sushko2020you} et al. StyleGAN-family~\cite{karras2019style, karras2020analyzing} models have shown impressive performance in unconditional high-resolution image synthesis. Diffusion probabilistic models~\cite{sohl2015deep, ho2020denoising, song2020denoising, kingma2021variational, dhariwal2021diffusion, san2021noise, kong2021fast, saharia2022photorealistic, rombach2022high, zhang2023adding} have emerged as another formidable family. Thanks to the Denoising Diffusion Probabilistic Model (DDPM)~\cite{ho2020denoising} and the Denoising Diffusion Implicit Model (DDIM)~\cite{song2020denoising}, Latent Diffusion Model (LDM) like Stable Diffusion~\cite{rombach2022high} achieves state-of-the-art performance in text-to-image generation tasks. Recognizing the vast potential of LDM, Controlnet~\cite{zhang2023adding} controls pretrained diffusion models with task-specific conditions, accommodating multi-modal inputs, and thereby expanding a wider range of application scenarios. While these models struggle to maintain the local spatial structures of grayscale inputs, their inherent capability to harness the diverse color priors from pretrained models offers a promising avenue for synthesizing reference images.

%% file: 04_method.tex
\section{Method}
\label{sec:Method}

\subsection{Overview}
\label{subsec:overview}

Given a grayscale image $\mathbf{X} \in \mathbb{R}^{H \times W \times 1}$ as input, our framework automatically predicts its color channels $\mathbf{Y} \in \mathbb{R}^{H \times W \times 2}$. $\mathbf{X}$ and $\mathbf{Y}$ are the lightness and $ab$ channel in the CIE $Lab$ color space. $H$ and $W$ are the height and width of the grayscale image. The key difference between our approach and existing automatic colorization methods is that: we leverage diverse color priors embedded in large pretrained generative models, such as Stable Diffusion, to synthesize colorful images based on the semantic context and spatial structure of the input grayscale image as a reference for colorization.

Figure~\ref{fig:Framework} shows the architecture of our framework. We first utilize ControlNet\cite{zhang2023adding} to control a large diffusion model to synthesize semantically similar and structurally aligned reference candidates $\mathbf{C} \in \mathbb{R}^{N \times H \times W \times 3}$ based on the grayscale input $\mathbf{X}$ from our imagination module $\mathcal{I}$:
\begin{equation}
    \mathbf{C} = \mathcal{I}(\mathbf{X}, \mathbf{c}(\mathbf{\mathbf{X}})),
\end{equation}
where $\mathbf{c}(\mathbf{X})$ means various optional conditions based on $\mathbf{X}$. The reference $\mathbf{R} \in \mathbb{R}^{H \times W \times 3}$ is synthesized by a reference refinement module $\mathcal{P}$ from multiple reference candidates $\mathbf{C}$. Then, our example-based colorization module $\mathcal{C}$ takes the grayscale image $\mathbf{X}$ and synthesized reference $\mathbf{R}$ as input, and outputs the corresponding photorealistic colorization results:
\begin{align}
    (\mathbf{X}, \mathbf{Y}) &= \mathcal{C}(\mathbf{X}, \mathbf{R}), \\
    \mathbf{R} &= \mathcal{P}(\mathbf{C}).
\end{align}
The details of the imagination module, reference refinement module and colorization module are illustrated in Section~\ref{subsec:Imagination-Module}, Section~\ref{subsec:Reference-Refinement-Module} and Section~\ref{subsec:Colorization-Module}.

\begin{figure}[t] 
  \centering
  \includegraphics[width=1\linewidth]{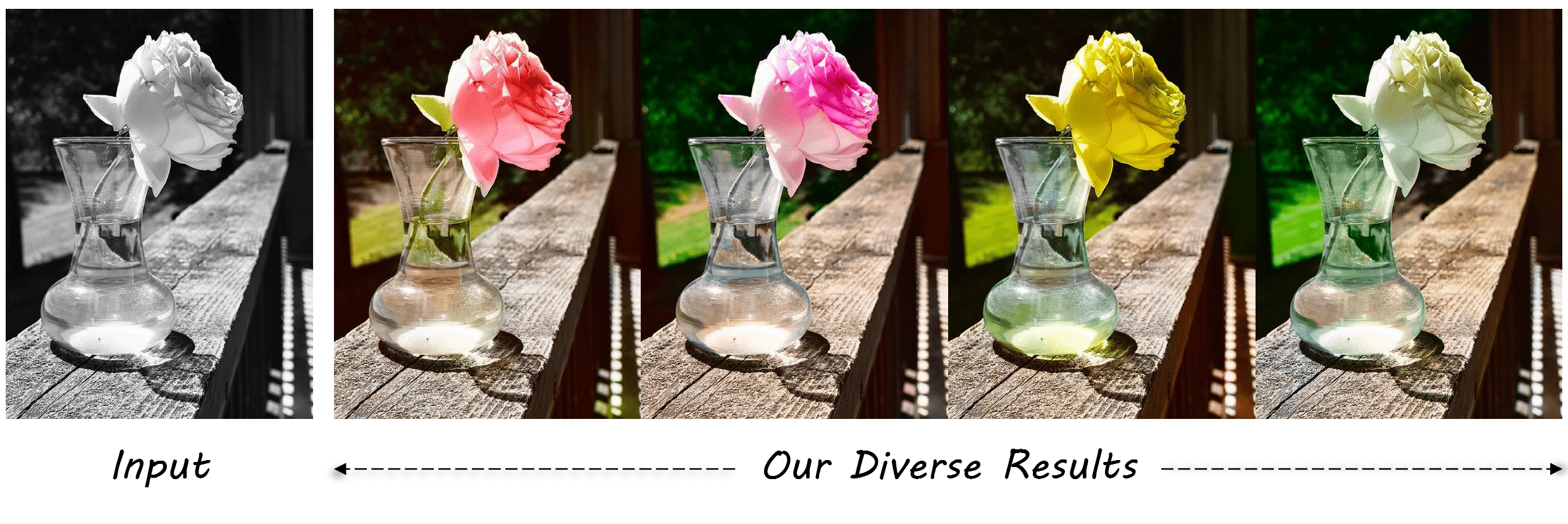}
  \vspace{-0.3in}
  \caption{\textbf{Diverse Colorization.} We can synthesize diverse colorful references from Imagination Module \ref{subsec:Imagination-Module}, yielding diverse colorzation results.}
  \label{fig:Diverse_Colorization}
\end{figure}

\subsection{Preliminary}
\label{subsec:Preliminary}

\paragraph{Image Diffusion Models.} Image Diffusion Models address the image generation task by learning to perform progressive denoising~\cite{ho2020denoising, dhariwal2021diffusion, saharia2022image, rombach2022high}. Given an image $\mathbf{I}_0$, noise is progressively added to produce $\mathbf{I}_t$, a noisy version of $\mathbf{I}_0$. Along with $t$ increasing, the image is gradually approaching pure noise. Image diffusion models learn a neural network $\epsilon_{\theta}$ to predict the noise added to the noisy image $\mathbf{I}_t$ and the objective function can be simplified as
\begin{equation}
    \mathcal{L} = \mathbb{E}_{\mathbf{I}_0, t, \mathbf{c}, \epsilon \in \mathcal{N}(0,1)}[ \| \epsilon - \epsilon_{\theta}(\mathbf{I}_t, t, \mathbf{c})\|_2^2 ].
\end{equation}
Successful image diffusion models like Stable Diffusion~\cite{rombach2022high} operate in the latent space of an autoencoder, making convergency more stable and faster.

\subsection{Imagination Module}
\label{subsec:Imagination-Module}
Given a grayscale image $\mathbf{X}$, our imagination module aims to synthesize multiple semantically similar and structurally aligned reference candidates $\mathbf{C}$. The motivation is to imitate human imagination: humans can achieve photorealistic colorization because they possess sufficient and realistic color priors regarding our colorful world. For each part (e.g. instance) within the grayscale image, we inherently harbor diverse and colorful examples, which empowers us to perform a locally and globally consistent photorealistic composition by selectively choosing from various examples for each part. Similarly, the reference candidates synthesized by our imagination module should encapsulate such color diversity and generalization attributes. Thus, pre-trained large image diffusion models such as Stable Diffusion are ideally suited as our synthesis model, as they inherently contain diverse color priors. 

In this work, we utilize ControlNet~\cite{zhang2023adding} to control large pretrained image diffusion models to synthesize a total of $N$ reference candidates given colorization-task-specific input conditions, with the help of captions generated by BLIP-2~\cite{li2023blip}. Numerous conditional inputs are available, including canny edge~\cite{canny1986computational}, HED boundary~\cite{xie2015holistically}, semantic segmentation map and so on. The overarching objective, irrespective of the chosen conditional input methods, is to furnish a diverse, colorful, and photorealistic color space. We provide an empirical approach: Given the grayscale input $\mathbf{X}$, we generate a total of $N$ reference candidates $\mathbf{C}$, $N_1$ of which are conditioned on the canny edge of $\mathbf{X}$, and $N_2$ of which are conditioned on the HED boundary of $\mathbf{X}$, $N = N_1 + N_2$. Notably, the direct application of ControlNet to colorization leads to issues like inconsistency, grayish, and color bleeding, as shown in Figure \ref{fig:controlnet_failure}. In our imagination module, however, ControlNet shows extremely superior performance on synthesizing diverse and colorful reference candidates, providing quite strong priors for the following colorization module.

  

\begin{figure}[t]
  \centering
  \includegraphics[width=1\linewidth]{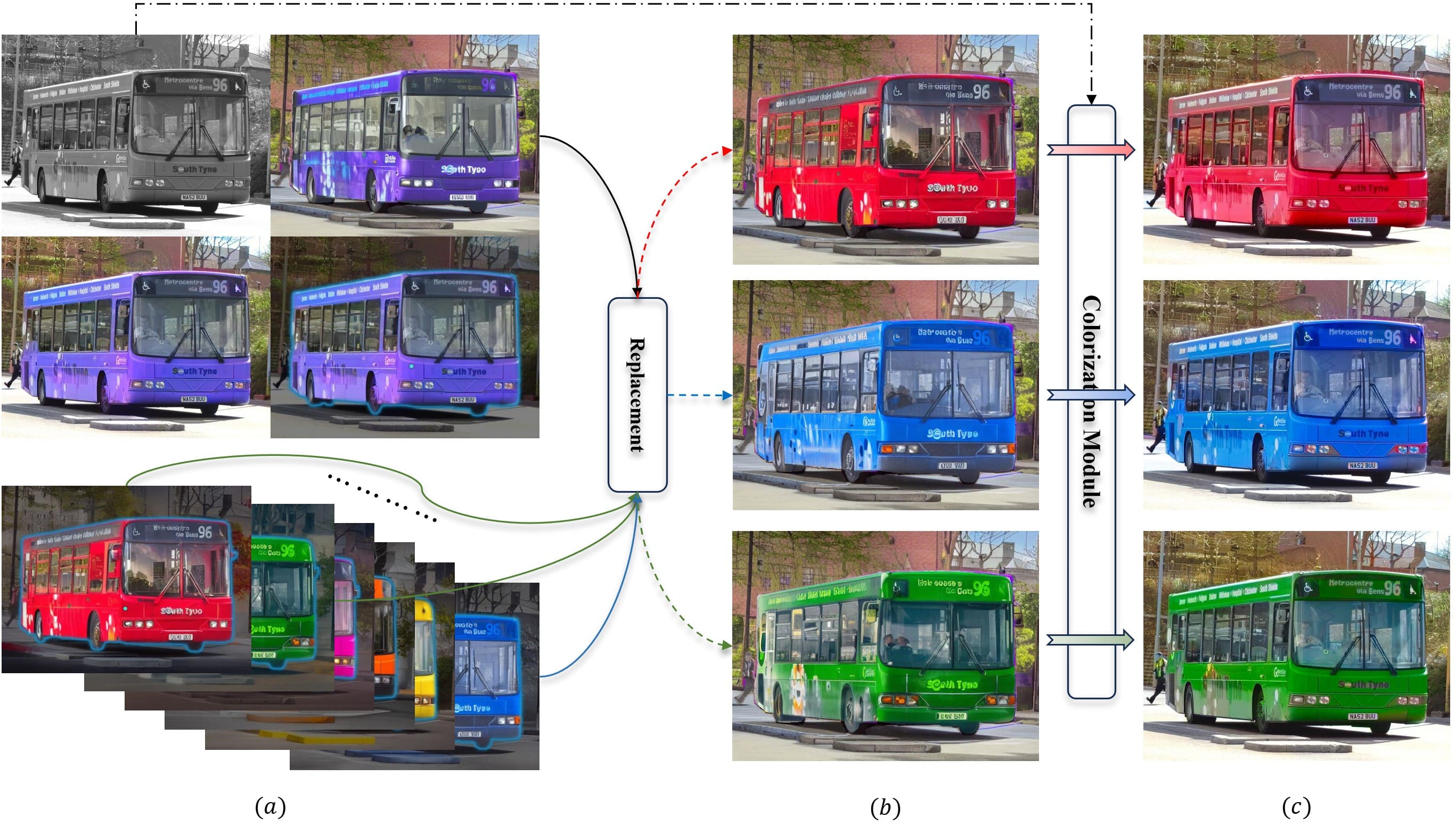}
  \vspace{-0.2in}
  \caption{\textbf{User Interaction.} The top panel of (a) shows the black-and-white input, the reference, the colorization result, and the area marked by the user's mouse click indicating dissatisfaction with the colorization. The bottom panel of (a) displays multiple reference candidates generated by our Imagination Module. (b) Users select a preferred segment from the reference candidates to replace the unsatisfactory part of the reference, consisting of a new reference. (c) The colorization result after the adjustment.}
  \label{fig:user_interactive}
  
\end{figure}

\subsection{Reference Refinement Module}
\label{subsec:Reference-Refinement-Module}
Given $N$ reference candidates $\mathbf{C} \in \mathbb{R}^{N \times H \times W \times 3}$ generated from our imagination module and the grayscale input $\mathbf{X}$, our reference refinement module $\mathcal{P}$ aims to synthesize the most semantically similar and structurally aligned refined reference $\mathbf{R}\in \mathbb{R}^{\times H \times W \times 3}$ through aggregating information across the diverse and colorful reference candidates space. Our motivation is that although a single reference candidate $\mathbf{R} = \mathcal{C}_i$ is enough to provide required priors and guidance in many cases, it is still worth performance improvement further by aggregating diverse information in multiple reference candidates to solve the problem: while $\mathcal{C}_i$ is semantically similar and structurally aligned with $\mathbf{X}$, there might be some incompatible mismatched segmentations in terms of luminance. For example, a desk with high luminance tends to be a bright color, but the generative model may offer a dark one. Both of them share the same grayscale input and ControlNet conditions like canny edge and HED boundary images.

We address this problem by composing a refined reference $\mathbf{R}$ from all multiple reference candidates $\mathbf{C}$. Specifically, we first use an off-the-shelf segmentation model~\cite{hamilton2022unsupervised, kirillov2023segment, zou2023segment, li2023semanticsam} to obtain the segmentation of $\mathbf{X}$. In our implementation, we employ the Semantic-SAM~\cite{li2023semanticsam} to generate semantic-aware and instance-aware segments $\mathbf{S}=\text{Semantic-SAM} (\mathbf{X})=\{ \mathbf{S}^j, j=1,2,\dots, N_{\mathbf{S}}\}$, where we denote each segment as $\mathbf{S}^j$. As $N_{\mathbf{S}}$ decreases, the composed reference becomes more inclined to capture the global information embedded in the reference candidates. Conversely, as $N_{\mathbf{S}}$ increases, the composed reference tends to capture more subtle details among the reference candidates. For each segment $\mathbf{S}^j$, we look for the best reference segment $\mathbf{R}\, \odot\, \mathbf{S}^j$ by selecting the nearest neighbor for $\mathbf{X} \, \odot \, \mathbf{S}^j$ among all reference candidates $\{\mathbf{C}_i \odot \mathbf{S}^j \}_{i=0}^N$ in terms of the difference in the universe feature space:
\begin{equation}
    \alpha(j) = \underset{i}{\mathrm{argmin}}\, \mathcal{M}(\mathcal{D}(\mathbf{X} \odot \mathbf{S}^j), \mathcal{D}(\mathbf{C}_i \odot \mathbf{S}^j )),
\end{equation}
where $\odot$ denotes operation to select some regions of an image, $\alpha(j)$ denotes the best reference candidate index for segment $\mathbf{S}^j$, $\mathcal{M}$ denotes distance measurement metrics ($\mathcal{L}_1, \mathcal{L}_2, \mathcal{L}_{\text{cosine}}$, etc.), and $\mathcal{D}(\cdot)$ denotes the robust and universe image feature DINOv2~\cite{oquab2023dinov2}. Then, our refined semantically similar and structurally aligned reference $R$ can be composed as
\begin{equation}
     \mathbf{R} = \bigcup_{j=0}^{N_\mathbf{S}}(S^j \odot C_{\alpha(j)}).
\end{equation}

\paragraph{User-interactive refinement.} So far, a semantically similar and structurally aligned reference can be synthesized automatically. Although such a reference is capable of achieving visually pleasing colorization, the color of each segment may not be the color that users are most satisfied with. Here we provide a user-interactive reference adjustment method: When users are unsatisfied about some segments $\mathbf{S}^j$, they can effortlessly substitute those segments $\mathbf{R}\, \odot\, \mathbf{S}^j$ with alternatives from the pool of reference candidates $\{\mathbf{C}_i \odot \mathbf{S}^j \}_{i=0}^N$ or use an image of their choosing. For example, in Figure \ref{fig:user_interactive}, we replace the bus with buses with different colors and preserve the colors in other areas. The colorization results are accordingly adjusted. In this case, we achieve controllable and editable colorization, which is non-trivial in the automatic colorization field.

\subsection{Colorization Module}
\label{subsec:Colorization-Module}
Our colorization module draws inspiration from the unified colorization framework UniColor~\cite{huang2022unicolor} to colorize the grayscale input $\mathbf{X} \in \mathbb{R}^{H\times W}$ by utilizing the reference image $\mathbf{R}\in \mathbb{R}^{H\times W \times 3}$ synthesized by our imagination module. UniColor is a unified colorization framework based on multi-modal conditions. Multi-modal conditions, such as strokes, reference images, and text, are converted into hint colors which are leveraged as a guidance for subsequent colorization. Specifically, the colorization process is divided into two steps: coarse-to-fine hint colors generation and hint colors propagation.

\begin{figure}[t]
  \centering
  \includegraphics[width=\linewidth]{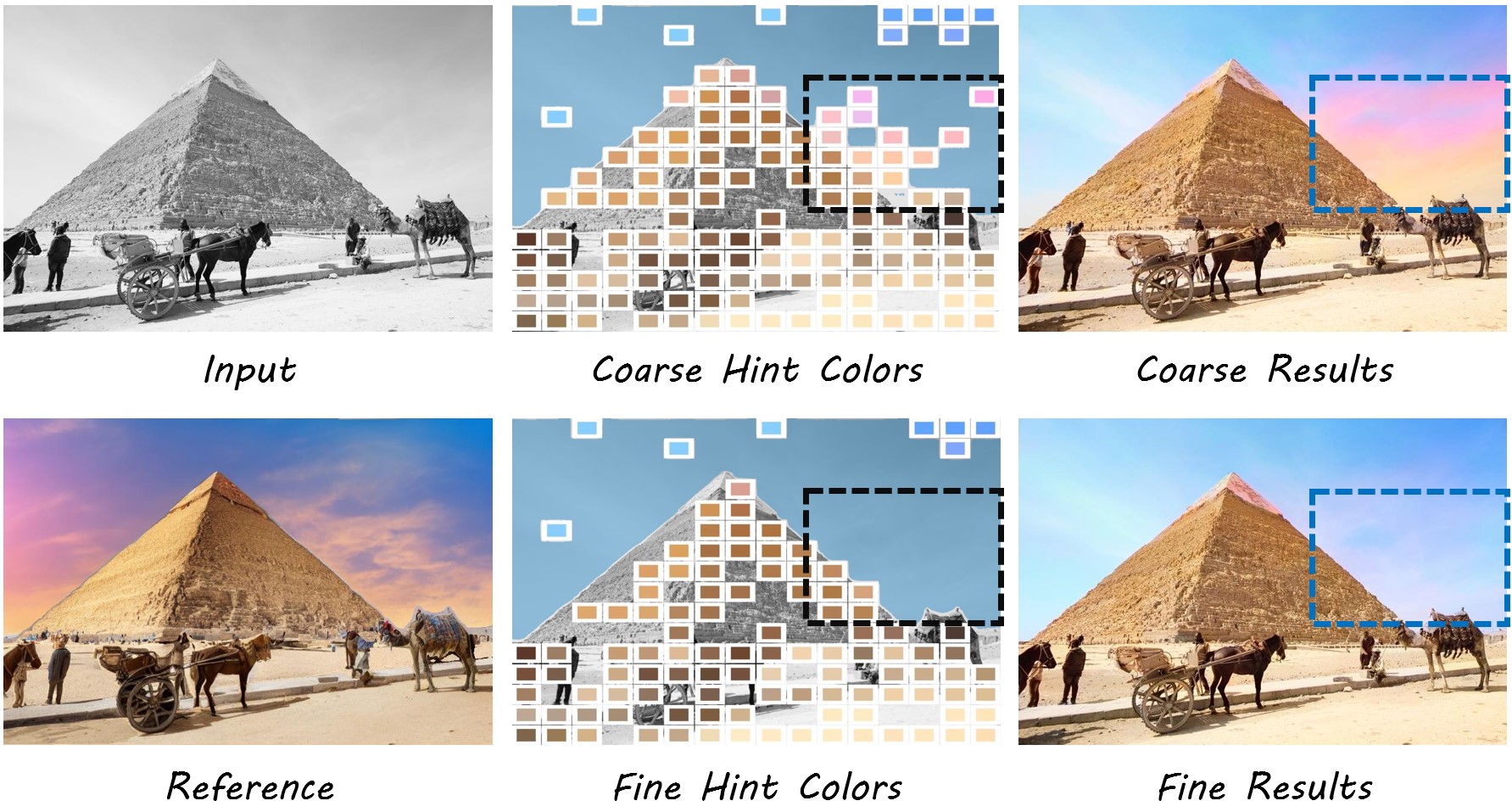}
  \vspace{-0.2in}
  \caption{\textbf{Coarse-to-fine hint colors optimization} can mitigate the color ambiguity within semantic instances by identifying the most representative hint colors, as highlighted by the boxed area in the figure.}
  \label{fig:ambiguity}
\end{figure}


\paragraph{Coarse-to-fine hint colors generation.}
\label{subsec:hint-colors-generation}
Hint color points are generated according to the semantic correspondence between $\mathbf{X}$ and $\mathbf{R}$ as an intermediate representation to instruct the following colorization process. Initially, we employ the correspondence network proposed in ~\cite{zhang2019deepcolor} to compute the pairwise similarity between the features of $\mathbf{X}$ and the features of $\mathbf{R}$ extracted by VGG19~\cite{simonyan2014very}, serving as the semantic correspondence, through which the color of $\mathbf{R}$ can be warped onto $\mathbf{X}$, resulting in a warped image $\mathbf{W}$. Similar to UniColor~\cite{huang2022unicolor}, $\mathbf{X}$ is divided into cells with size $d \times d$. If a cell's average semantic similarity surpasses the threshold value $\mathbf{s}_{\epsilon}$, the mean color of that cell in the warped image is designated as a hint color. This helps in preventing the mismatching area as the hint colors. However, we observe that merely employing the aforementioned method to generate hint colors can lead to semantic color ambiguity, as illustrated in Figure \ref{fig:ambiguity}. This is due to a distinction from the conventional reference-based colorization approach where a semantically similar, high-quality reference is manually provided. In our imagination module, the process of synthesizing $\mathbf{R}$ might introduce color inconsistency within semantic masks. Therefore, we treat the hint colors derived above as a coarse hint colors set $\mathcal{H}_{coarse}$ and obtain the fine hint colors set $\mathcal{H}_{fine}$ through optimization. Utilizing the Semantic-SAM segments $\mathbf{S} = \{ \mathbf{S}^j, j=1,2,\dots, N_S\}$ from the Reference Refinement Module, we associate each segment $\mathbf{S}^j$ with the corresponding coarse hint colors in $\mathcal{H}_{coarse}$. If the number of hint colors in  $\mathbf{S}^j$ exceeds $\mathcal{N}_{\mathcal{H}}$, we cluster the hint colors in $\mathbf{S}^j$ to identify and eliminate any hint colors with anomalous color values from the color distribution of most hint colors. In our implementation, we derive the fine hint colors set by leveraging the DBSCAN algorithm~\cite{ester1996density, jang2019dbscan} for clustering.

\paragraph{Hint colors propagation.}
Similar to UniColor~\cite{huang2022unicolor}, given a black-and-white input and fine hint colors, a Hybrid-Transformer is harnessed to autoregressively predict chrominance features, which are then concatenated with luminance features extracted from the input. The concatenated features are fed into a decoder to produce the final colorization result. We refer readers to UniColor~\cite{huang2022unicolor} for more details.

%% file: 05_experiment.tex
\begin{figure*}
    \centering
    \setlength{\tabcolsep}{0pt} 
    \renewcommand{\arraystretch}{0.5} 
    \begin{tabular}{*{10}{p{0.1\textwidth}}} 

        \includegraphics[width=0.1\textwidth]{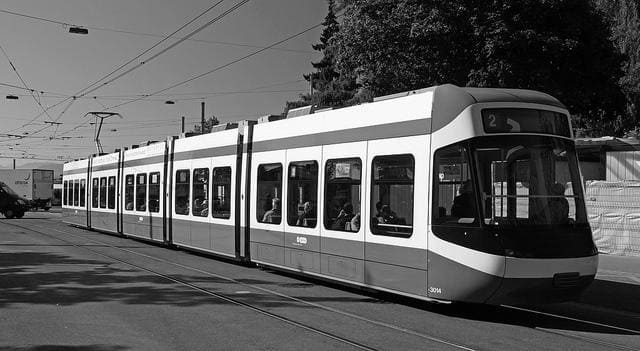} &
        \includegraphics[width=0.1\textwidth]{figs/coco_comparison/000000006040/GT_low_res.jpg} &
        \includegraphics[width=0.1\textwidth]{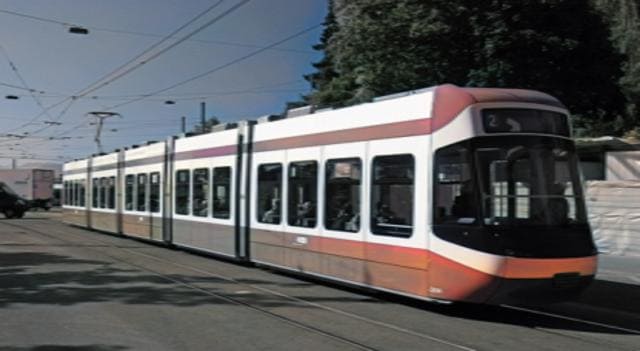} &
        \includegraphics[width=0.1\textwidth]{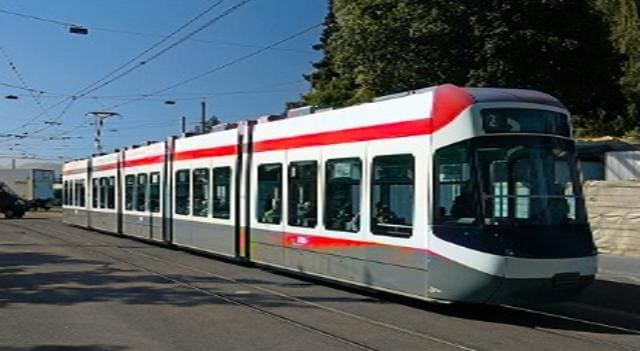} &
        \includegraphics[width=0.1\textwidth]{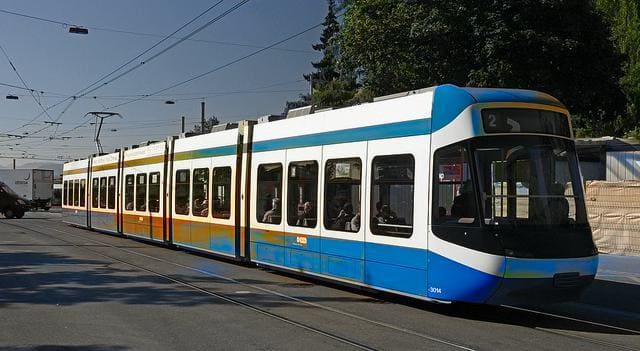} &
        \includegraphics[width=0.1\textwidth]{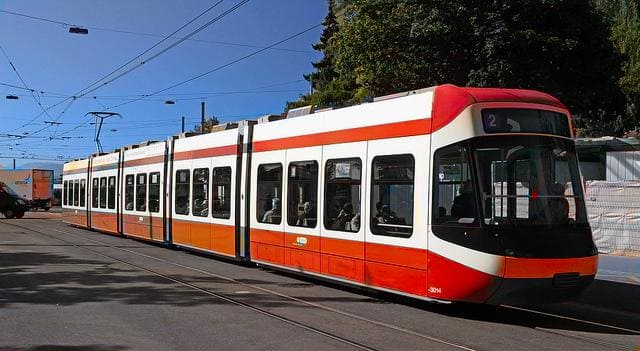} &
        \includegraphics[width=0.1\textwidth]{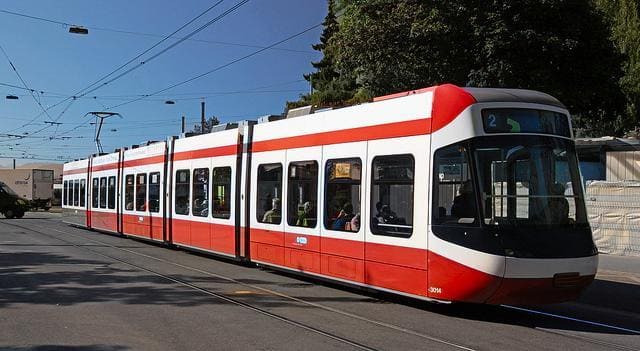} &
        \includegraphics[width=0.1\textwidth]{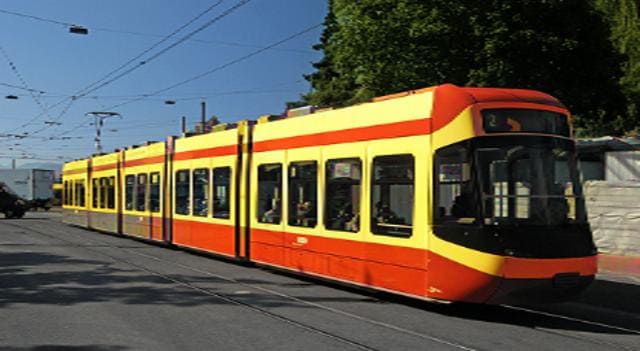} &
        \includegraphics[width=0.1\textwidth]{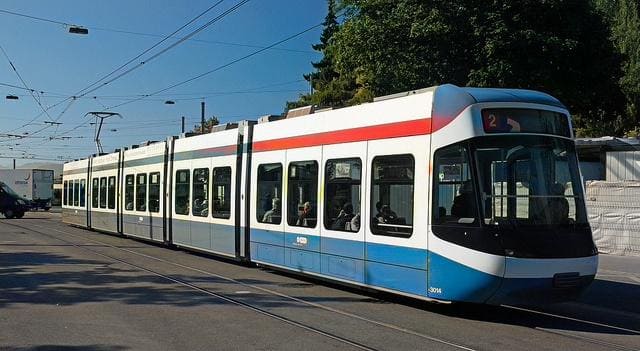} &
        \includegraphics[width=0.1\textwidth]{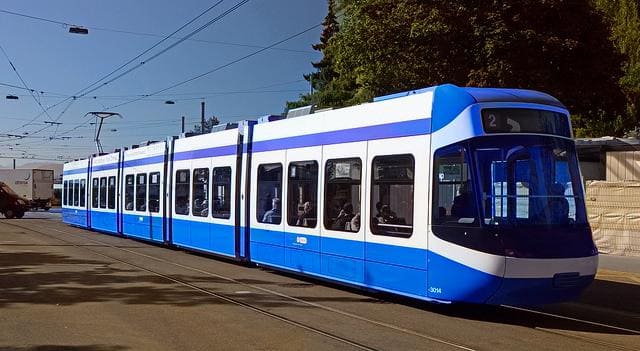} \\

        \includegraphics[width=0.1\textwidth]{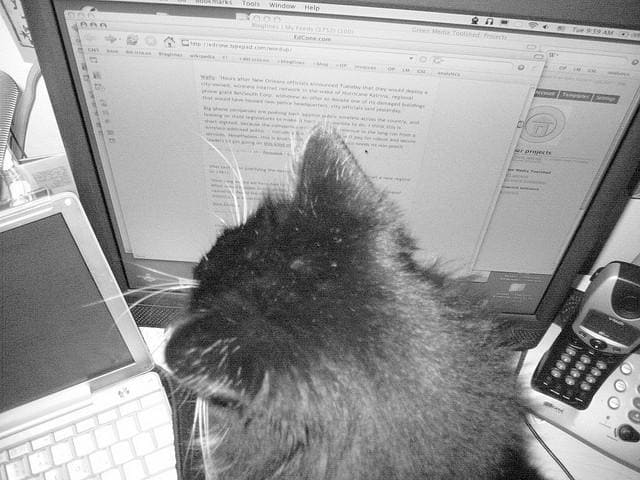} &
        \includegraphics[width=0.1\textwidth]{figs/coco_comparison/000000004795/GT_low_res.jpg} &
        \includegraphics[width=0.1\textwidth]{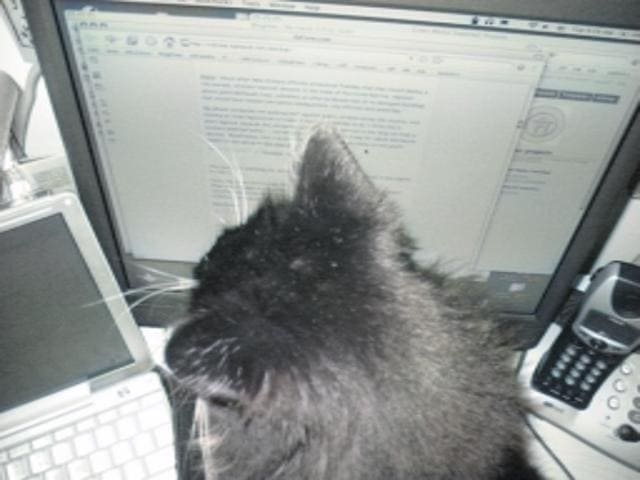} &
        \includegraphics[width=0.1\textwidth]{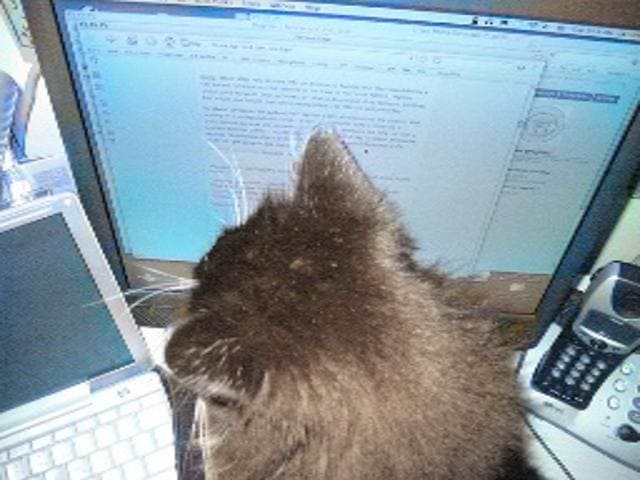} &
        \includegraphics[width=0.1\textwidth]{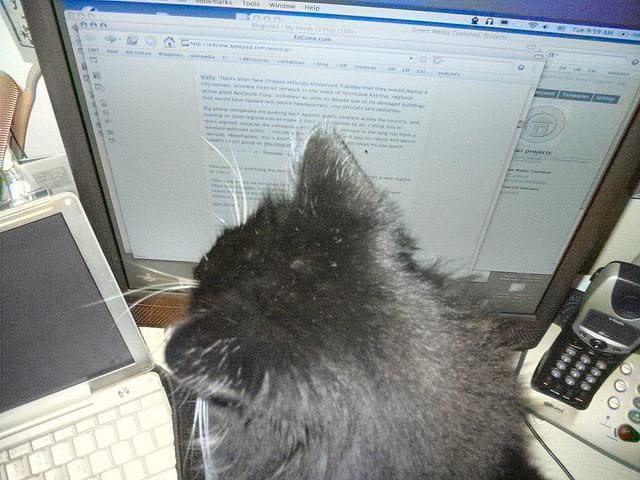} &
        \includegraphics[width=0.1\textwidth]{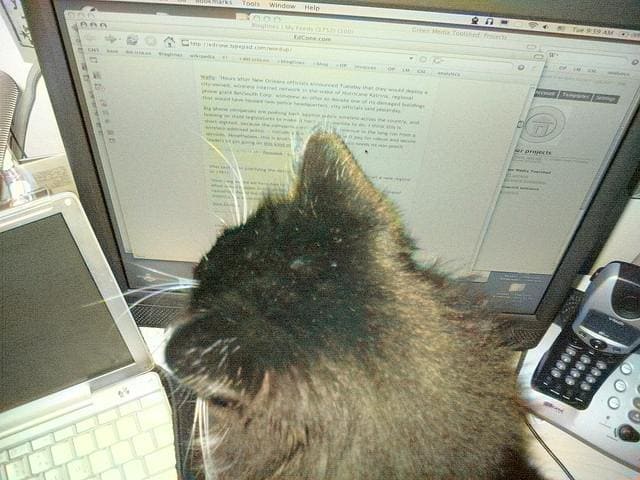} &
        \includegraphics[width=0.1\textwidth]{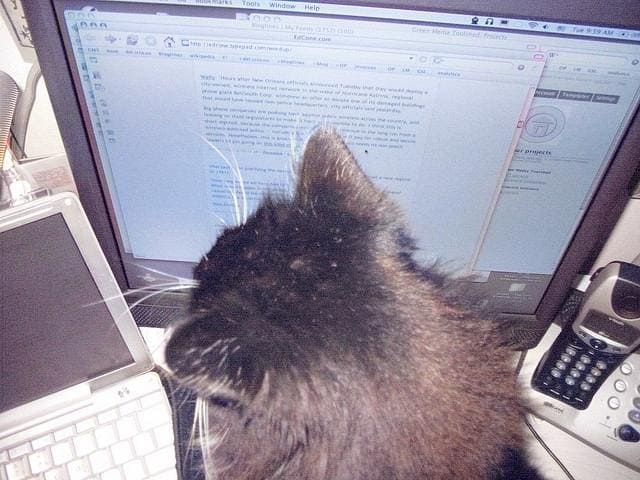} &
        \includegraphics[width=0.1\textwidth]{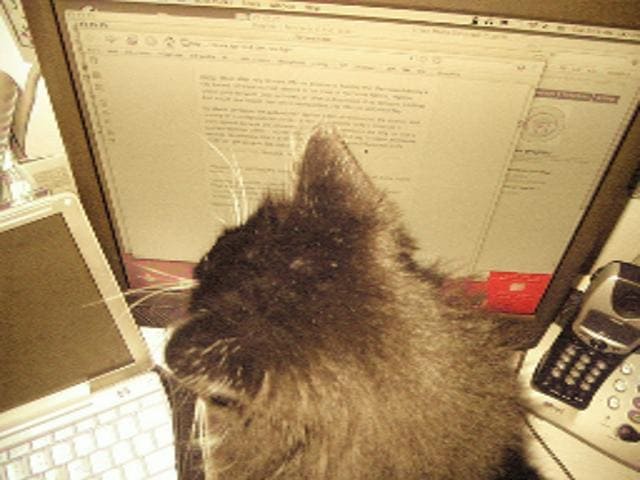} &
        \includegraphics[width=0.1\textwidth]{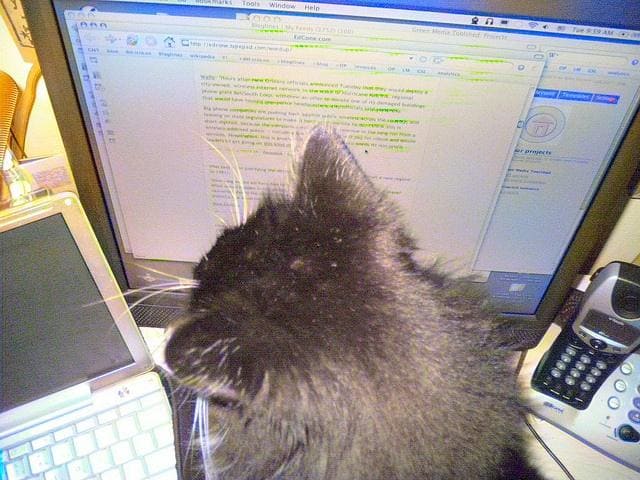} &
        \includegraphics[width=0.1\textwidth]{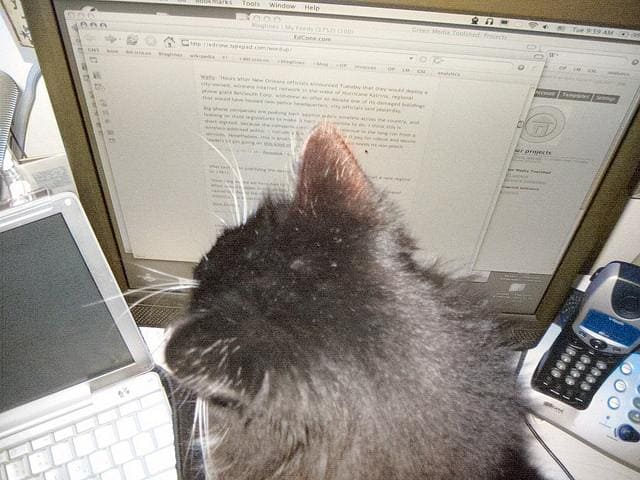} \\
        
        \includegraphics[width=0.1\textwidth]{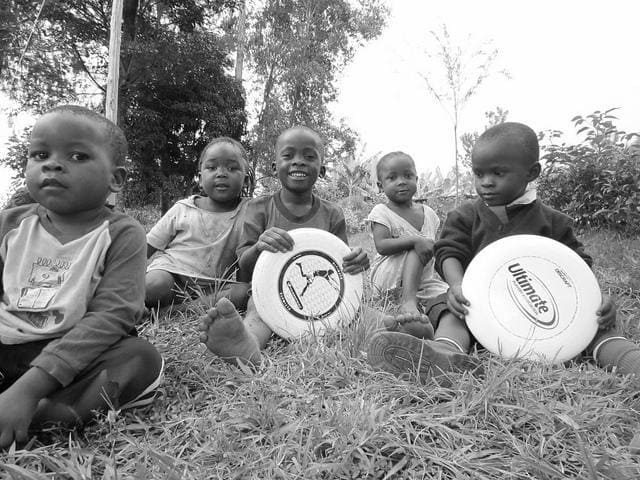} &
        \includegraphics[width=0.1\textwidth]{figs/coco_comparison/000000006954/GT_low_res.jpg} &
        \includegraphics[width=0.1\textwidth]{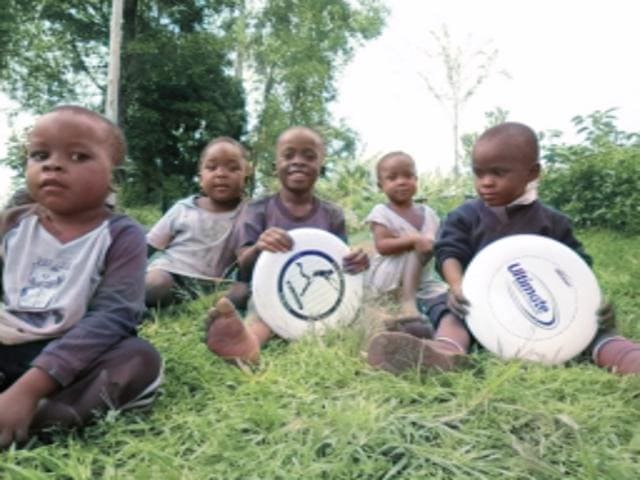} &
        \includegraphics[width=0.1\textwidth]{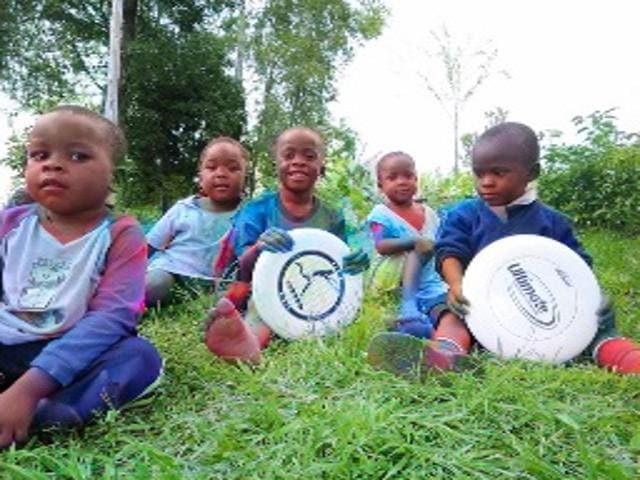} &
        \includegraphics[width=0.1\textwidth]{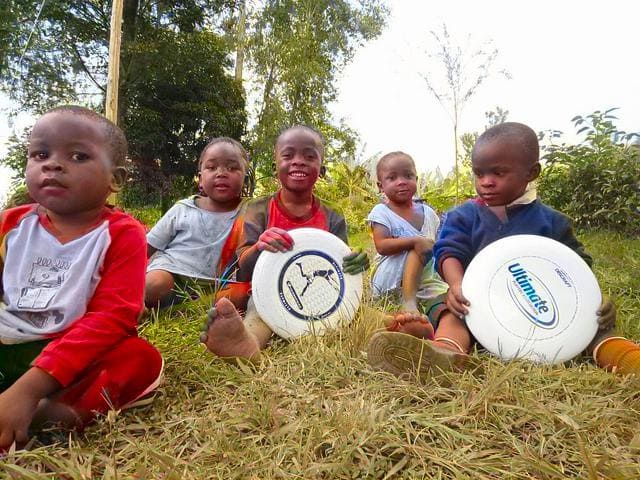} &
        \includegraphics[width=0.1\textwidth]{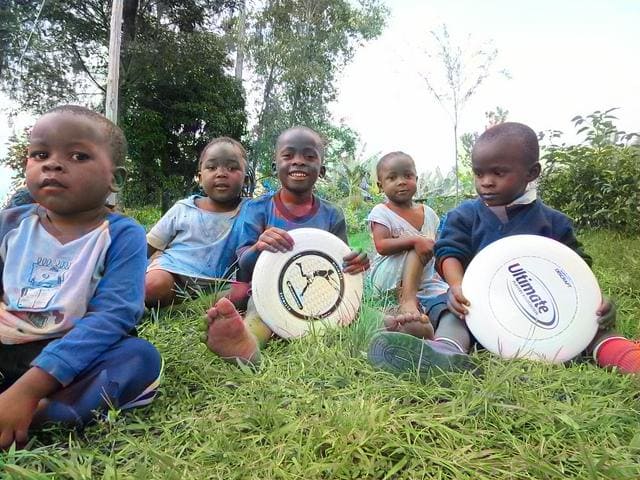} &
        \includegraphics[width=0.1\textwidth]{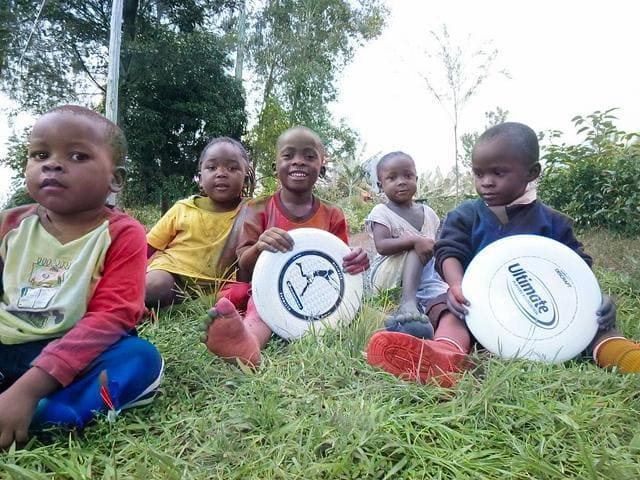} &
        \includegraphics[width=0.1\textwidth]{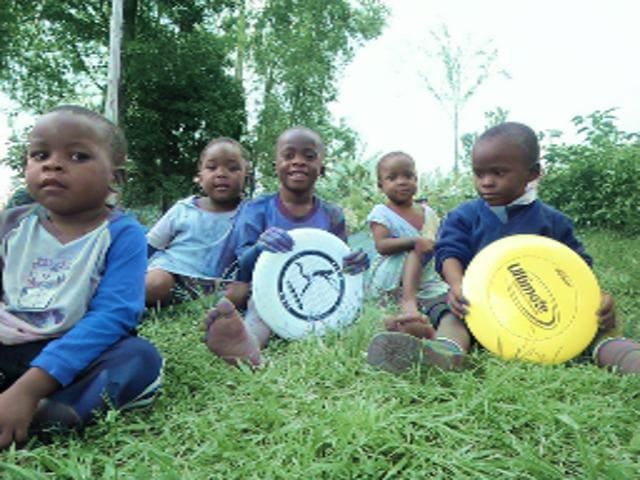} &
        \includegraphics[width=0.1\textwidth]{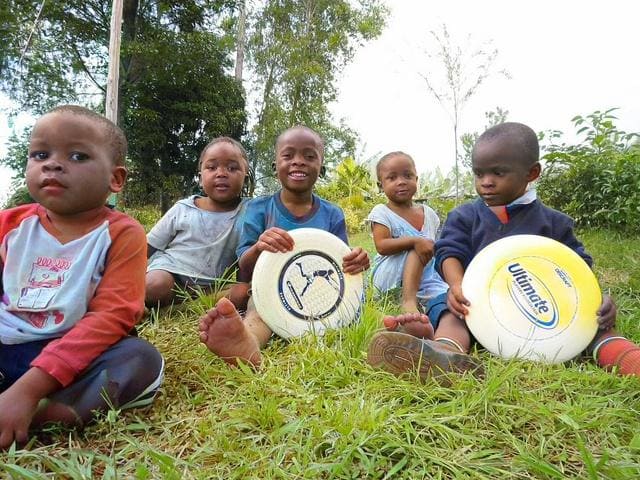} &
        \includegraphics[width=0.1\textwidth]{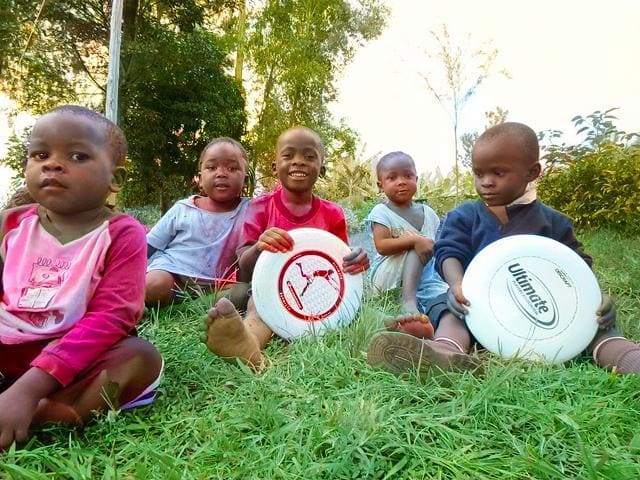} \\

        \includegraphics[width=0.1\textwidth]{figs/coco_comparison/000000058393/gray_low_res.jpg} &
        \includegraphics[width=0.1\textwidth]{figs/coco_comparison/000000058393/GT_low_res.jpg} &
        \includegraphics[width=0.1\textwidth]{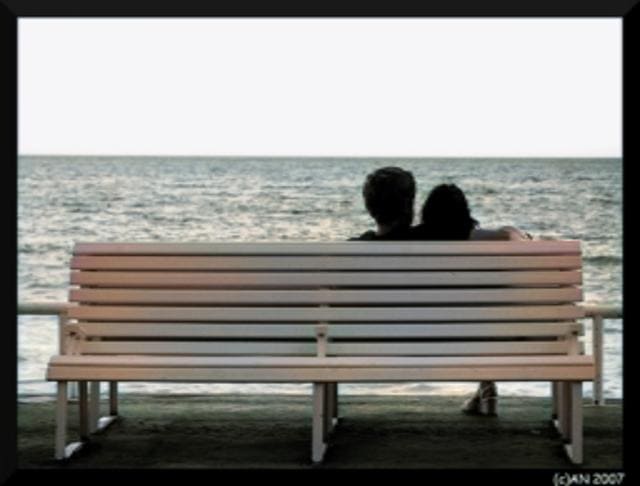} &
        \includegraphics[width=0.1\textwidth]{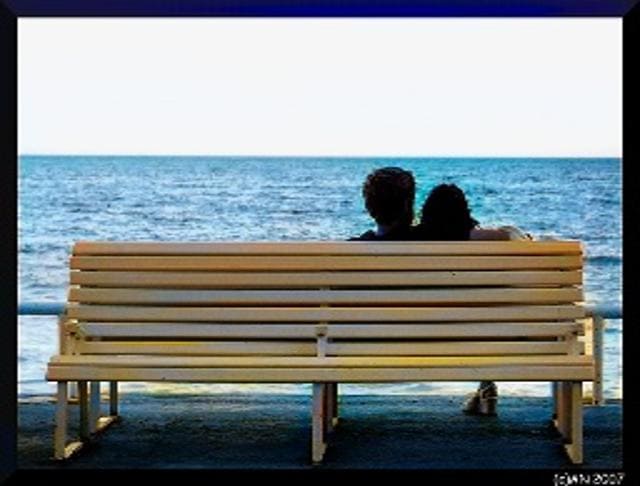} &
        \includegraphics[width=0.1\textwidth]{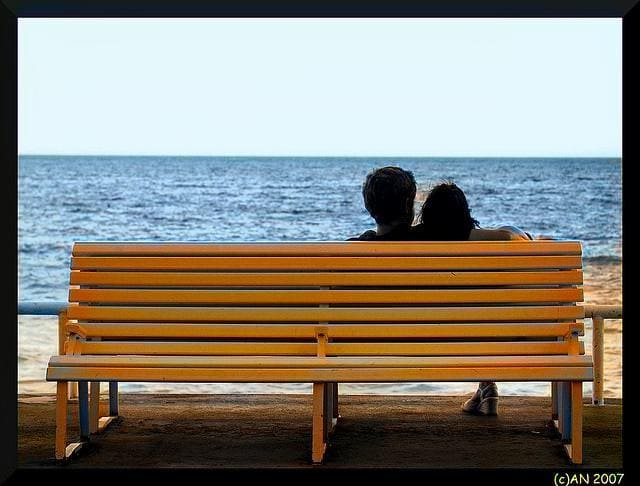} &
        \includegraphics[width=0.1\textwidth]{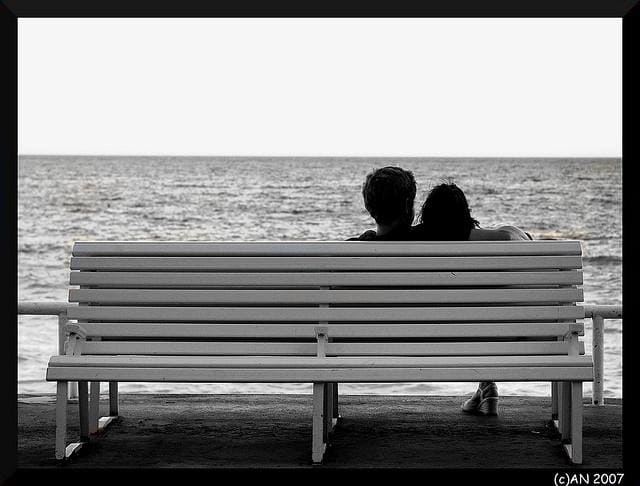} &
        \includegraphics[width=0.1\textwidth]{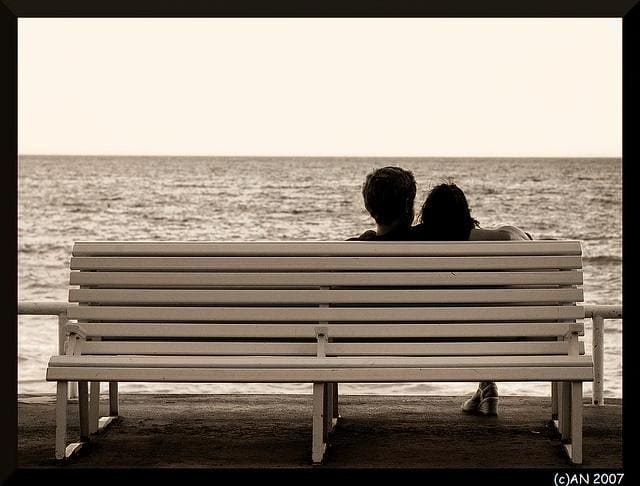} &
        \includegraphics[width=0.1\textwidth]{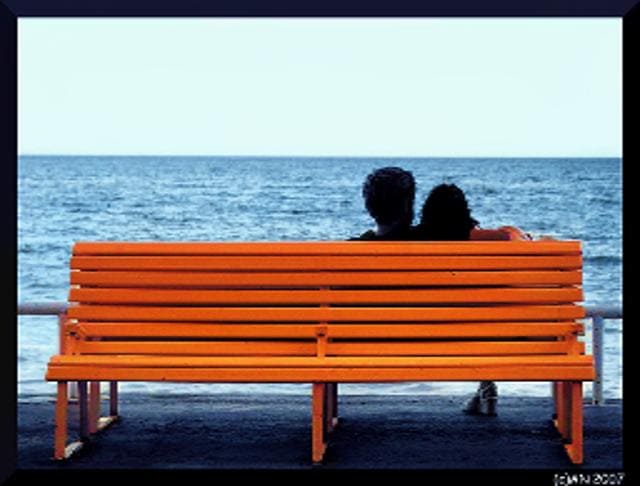} &
        \includegraphics[width=0.1\textwidth]{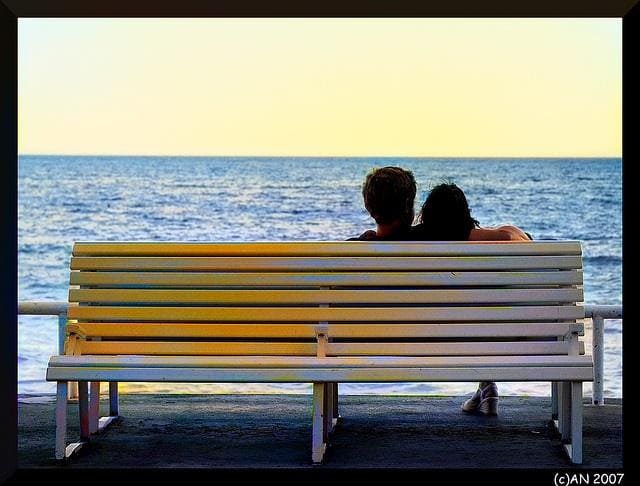} &
        \includegraphics[width=0.1\textwidth]{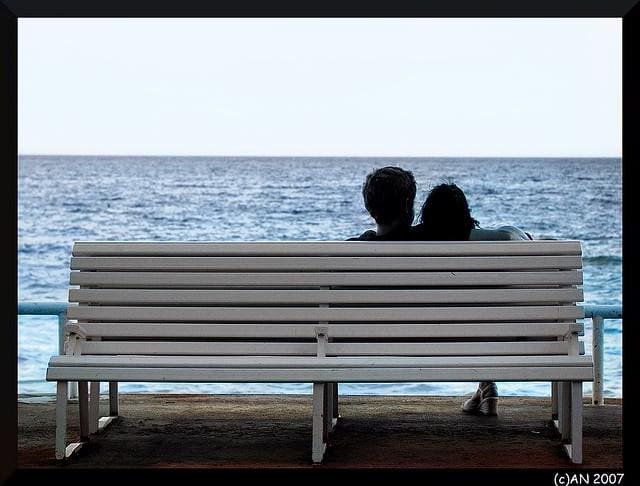} \\

        \includegraphics[width=0.1\textwidth]{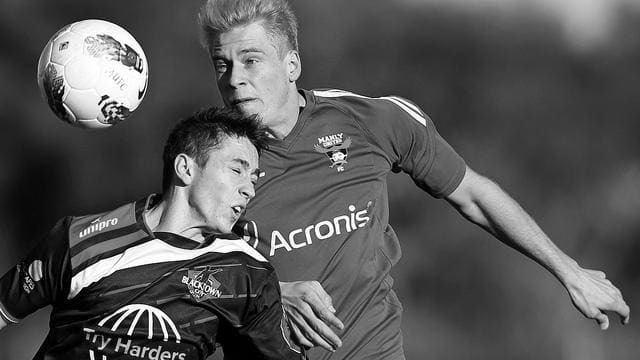} &
        \includegraphics[width=0.1\textwidth]{figs/coco_comparison/000000060102/GT_low_res.jpg} &
        \includegraphics[width=0.1\textwidth]{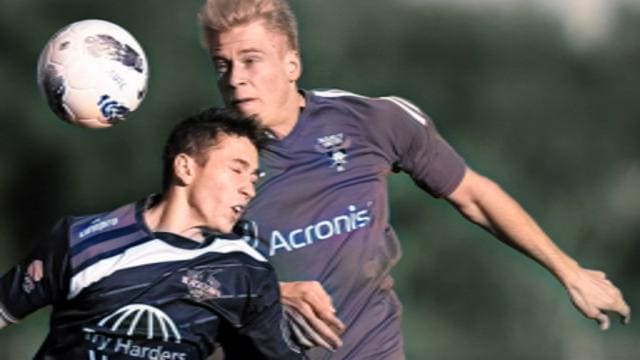} &
        \includegraphics[width=0.1\textwidth]{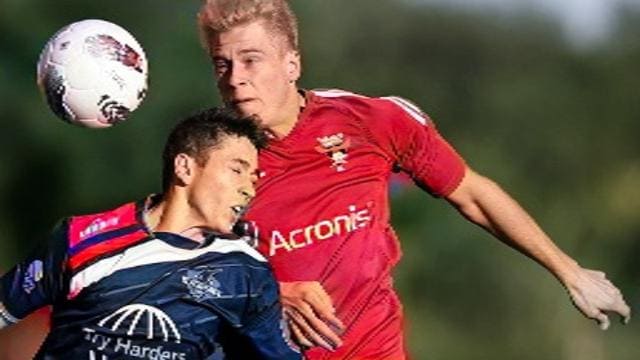} &
        \includegraphics[width=0.1\textwidth]{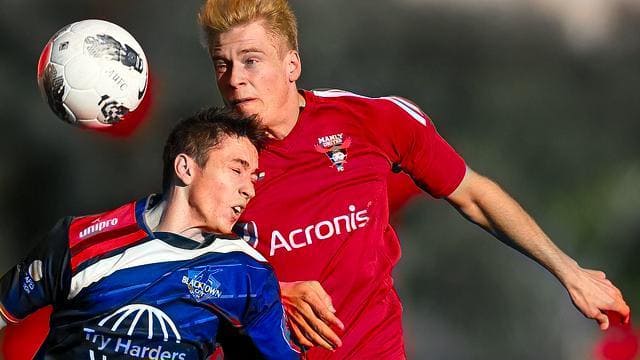} &
        \includegraphics[width=0.1\textwidth]{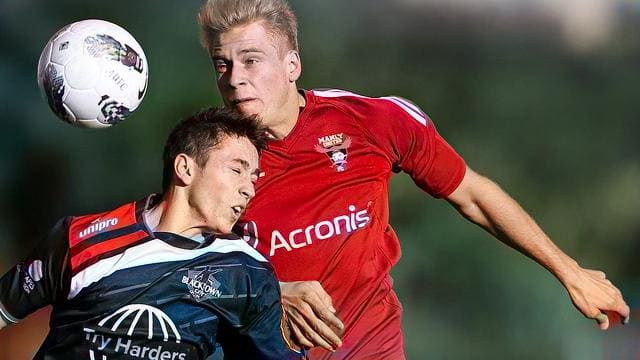} &
        \includegraphics[width=0.1\textwidth]{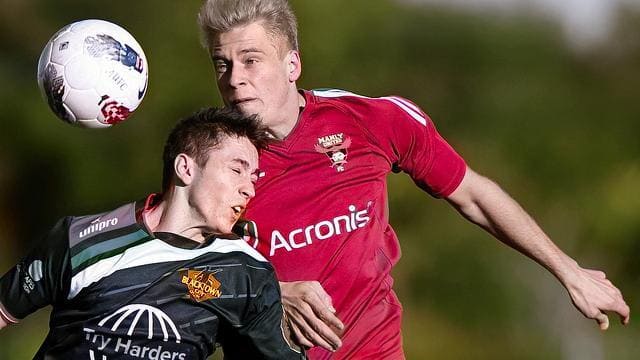} &
        \includegraphics[width=0.1\textwidth]{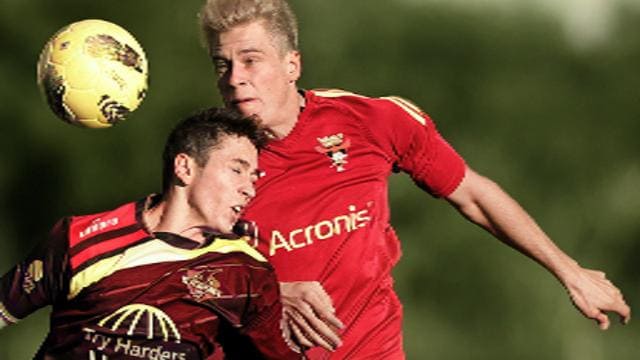} &
        \includegraphics[width=0.1\textwidth]{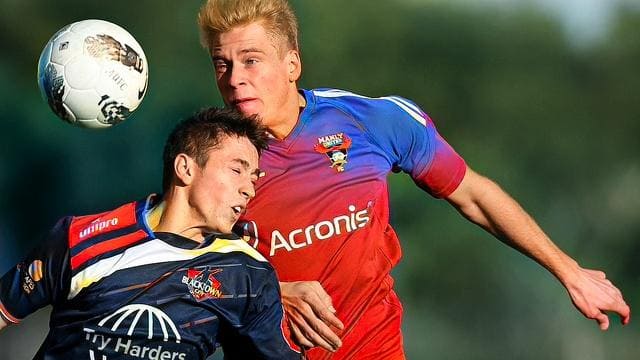} &
        \includegraphics[width=0.1\textwidth]{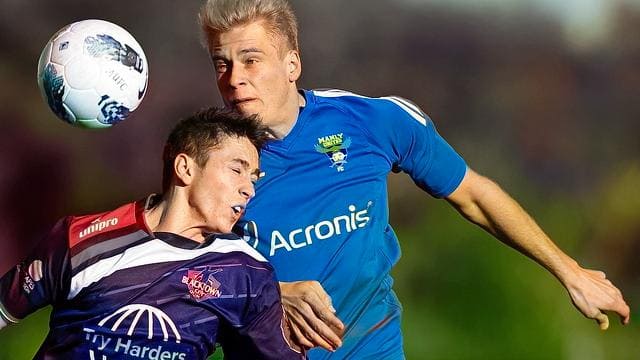} \\

        \multicolumn{1}{c}{\small\scalefont{0.7}Input} &
        \multicolumn{1}{c}{\small\scalefont{0.7}Ground Truth}  &
        \multicolumn{1}{c}{\small\scalefont{0.7}Deoldify~\cite{deoldify2019}} &
        \multicolumn{1}{c}{\small\scalefont{0.7}ColorFormer~\cite{10.1007/978-3-031-19787-1_2}} &
        \multicolumn{1}{c}{\small\scalefont{0.7}BigColor~\cite{Kim2022BigColor}} &
        \multicolumn{1}{c}{\small\scalefont{0.7}GCP~\cite{wu2021towards}} &
        \multicolumn{1}{c}{\small\scalefont{0.7}Unicolor~\cite{huang2022unicolor}} &
        \multicolumn{1}{c}{\small\scalefont{0.7}DISCO~\cite{XiaHWW22}} &
        \multicolumn{1}{c}{\small\scalefont{0.7}DDColor~\cite{kang2022ddcolor}} &
        \multicolumn{1}{c}{\small\scalefont{0.7}Ours}
        
    \end{tabular}
    \caption{\textbf{Qualitative comparisons on the COCO validation set.} Our method can generate more natural and photo-realistic colors than state-of-the-art approaches. Zoom in for the best view.}
    \label{fig:qualitive_compare_coco}
\end{figure*}

\section{Experiments}
\label{sec:Experiments}
\subsection{Experimental setting} 

\noindent \textbf{Dataset.} Our experiments are mainly conducted on standard evaluation benchmark, $i.e.$, COCO-stuff~\cite{caesar2018cocostuff} and ImageNet~\cite{deng2009imagenet}. Both our framework and baselines are evaluated in $(i)$ COCO-stuff validation set (5k images) $(ii)$ ImageNet testing split ctest (10k images), $(iii)$ $In\text{-}the\text{-}wild$ photos collected on the Internet (500 images), qualitatively and quantitatively. Notably, evaluating the colorization performance on arbitrary grayscale images from the Internet, often referred to as $In\text{-}the\text{-}wild$, is incredibly crucial. This is because they might have a completely different color distribution compared to widely used datasets such as ImageNet and COCO-stuff. Such differences pose a significant challenge to the generalization capability of colorization models. In this regard, our framework significantly outperforms the state-of-the-art baselines.

\noindent \textbf{Evaluation Metrics.} In experiments, we mainly care about the framework's performance on perceptual realism and color vividness. For measuring perceptual realism, we adopt Fr\'echet Inception Distance (\textbf{FID})~\cite{heusel2017gans} to compare the distribution similarity between the colorization results and the ground truth. For color vividness, we employ the Colorfulness (\textbf{CF}) metric~\cite{hasler2003measuring} which is close to human vision perception. In addition, following previous works~\cite{Su-CVPR-2020, CT2, zhang2017real}, we provide evaluations on PSNR, SSIM and LIPIS~\cite{zhang2018unreasonable} for reference.

\begin{figure*}[t]
    \centering
    \setlength{\tabcolsep}{0pt} 
    \renewcommand{\arraystretch}{0.5} 
    \begin{tabular}{*{9}{p{0.1111111111\textwidth}}} 

        \includegraphics[width=0.1111111111\textwidth]{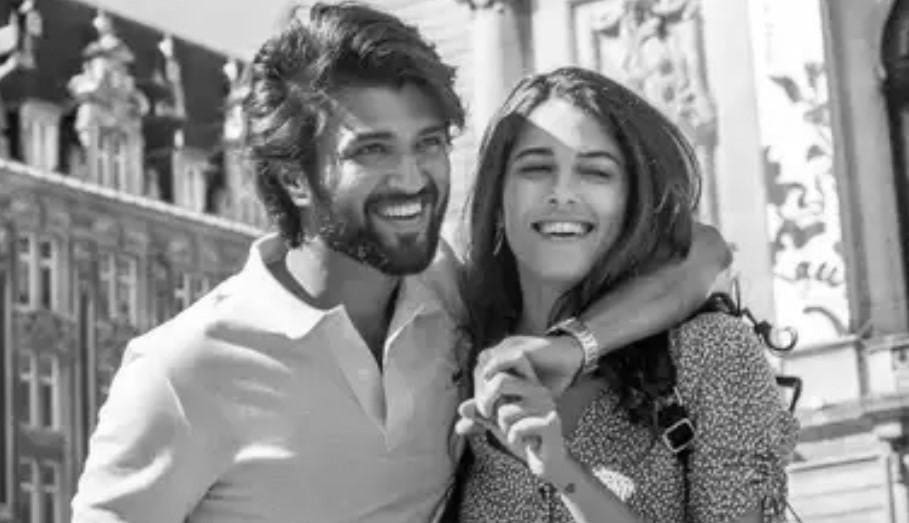} &
        \includegraphics[width=0.1111111111\textwidth]{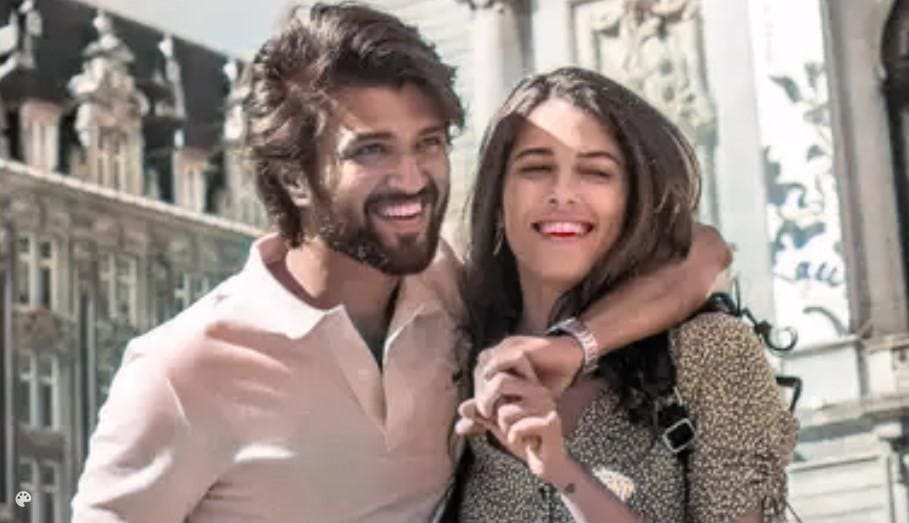} &
        \includegraphics[width=0.1111111111\textwidth]{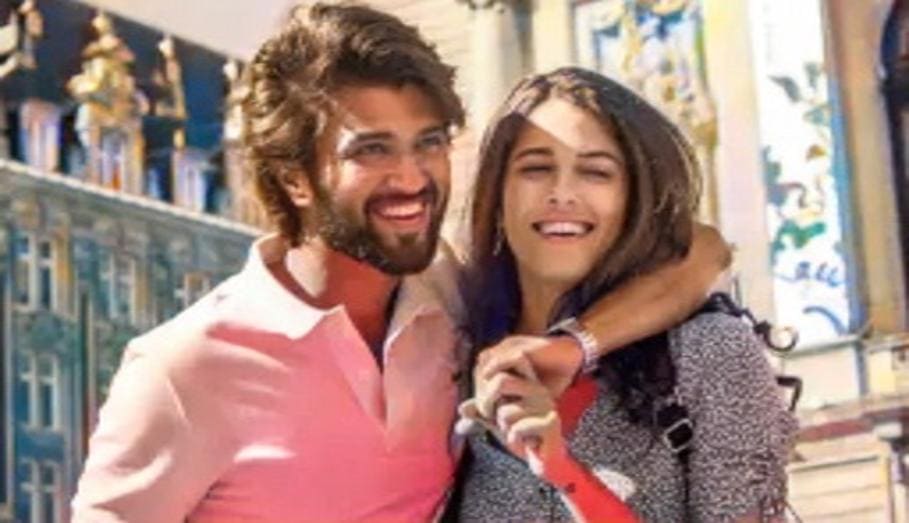} &
        \includegraphics[width=0.1111111111\textwidth]{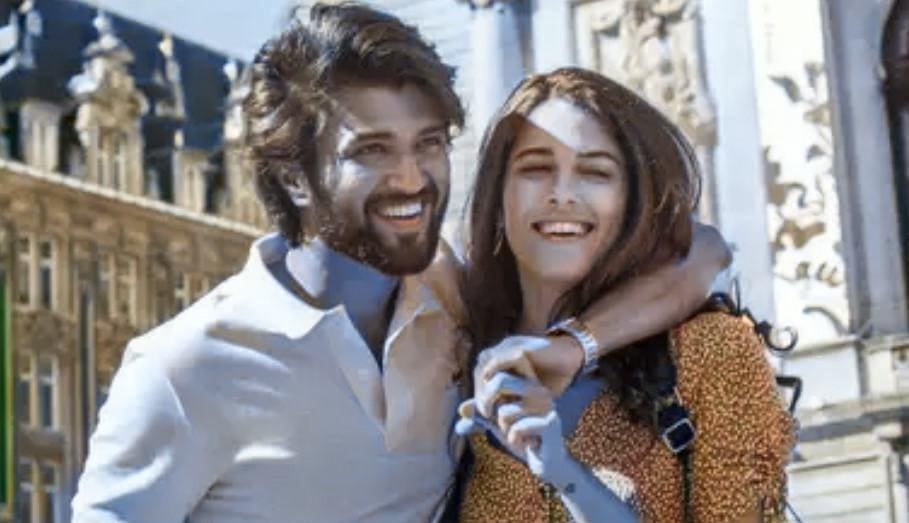} &
        \includegraphics[width=0.1111111111\textwidth]{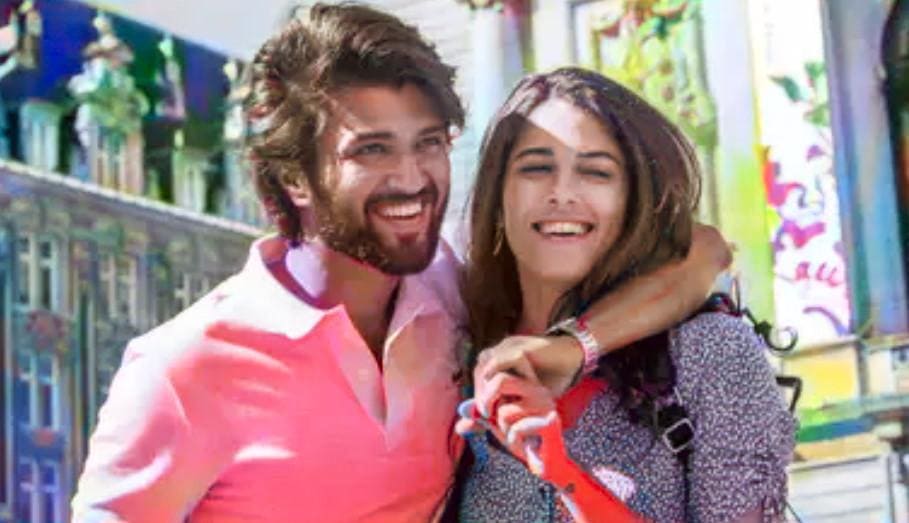} &
        \includegraphics[width=0.1111111111\textwidth]{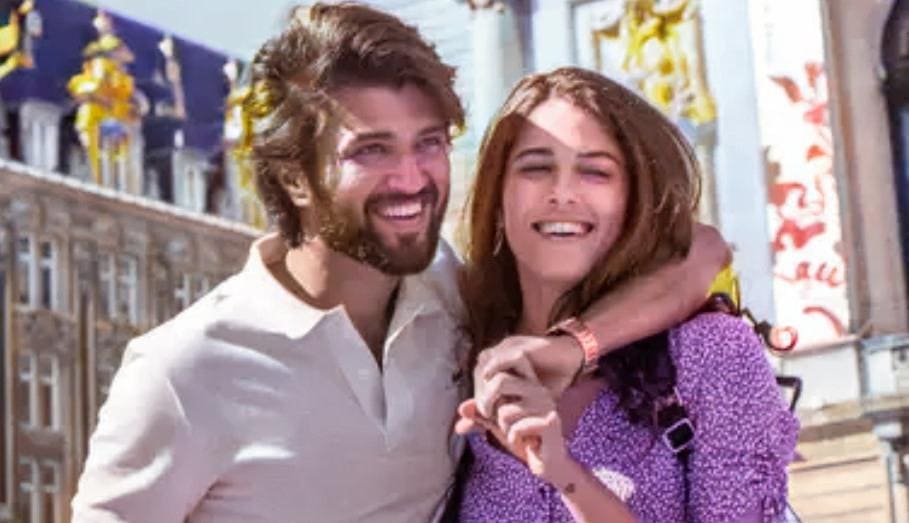} &
        \includegraphics[width=0.1111111111\textwidth]{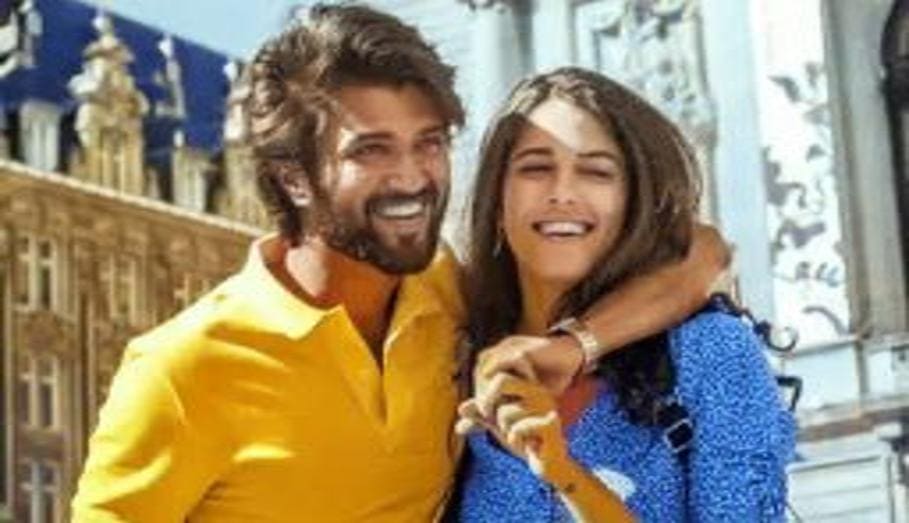} &
        \includegraphics[width=0.1111111111\textwidth]{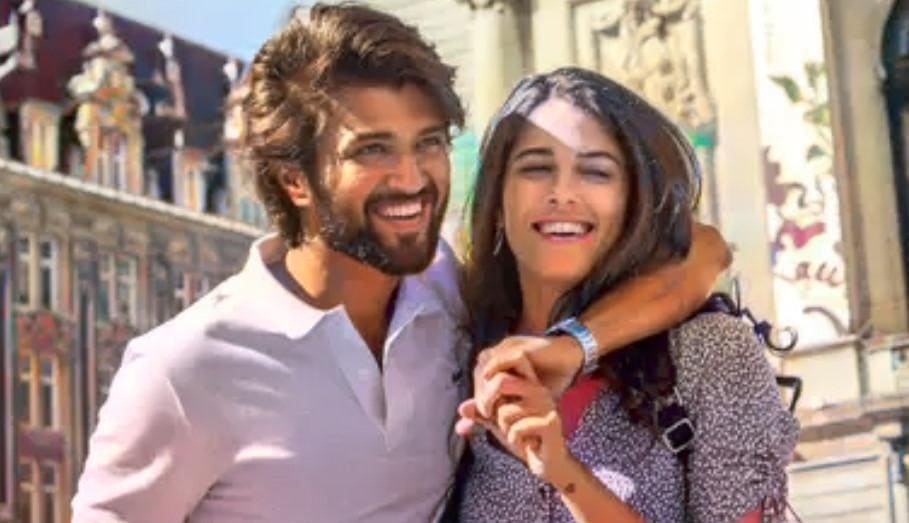} &
        \includegraphics[width=0.1111111111\textwidth]{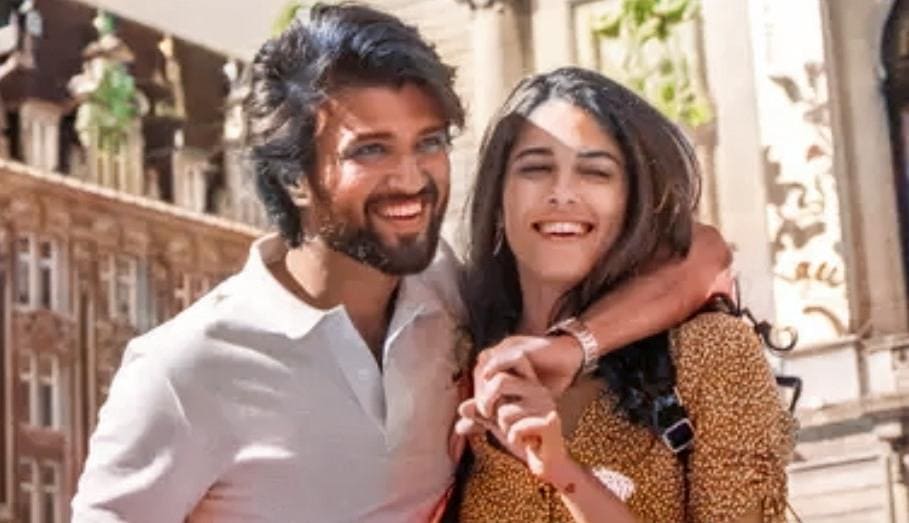}\\

        \includegraphics[width=0.1111111111\textwidth]{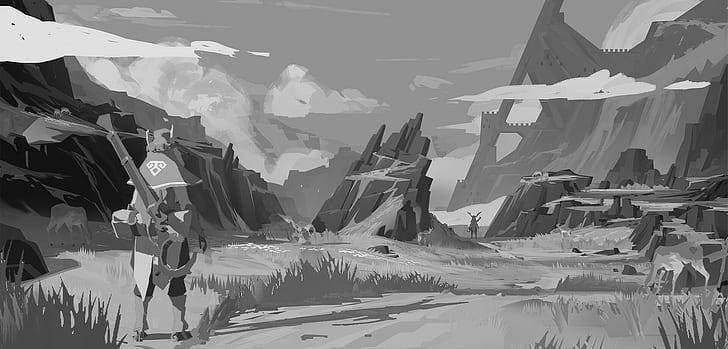} &
        \includegraphics[width=0.1111111111\textwidth]{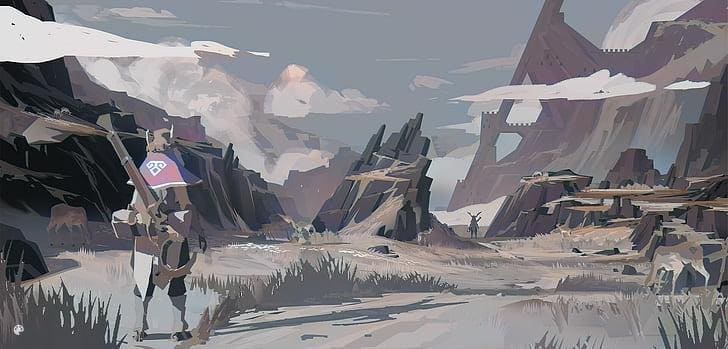} &
        \includegraphics[width=0.1111111111\textwidth]{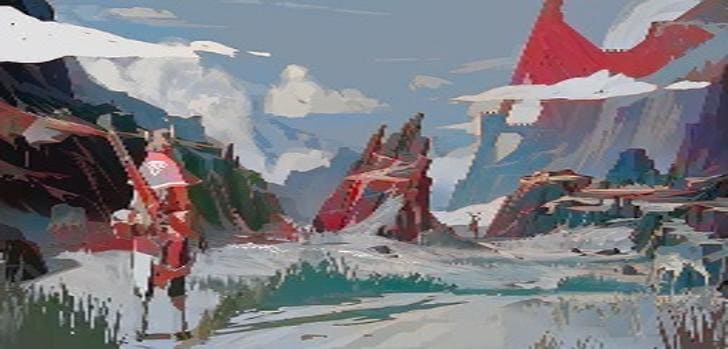} &
        \includegraphics[width=0.1111111111\textwidth]{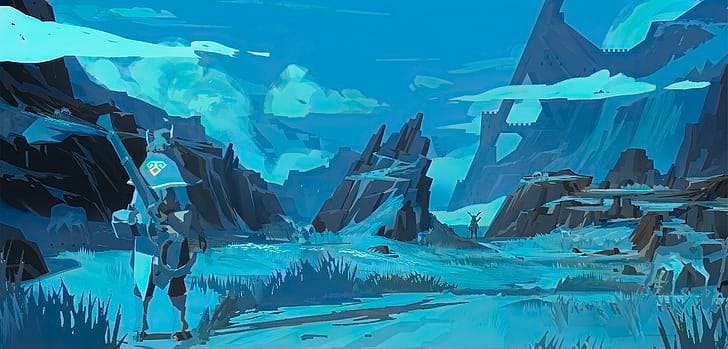} &
        \includegraphics[width=0.1111111111\textwidth]{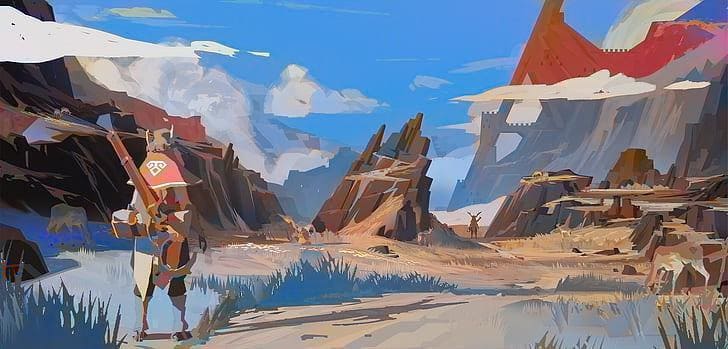} &
        \includegraphics[width=0.1111111111\textwidth]{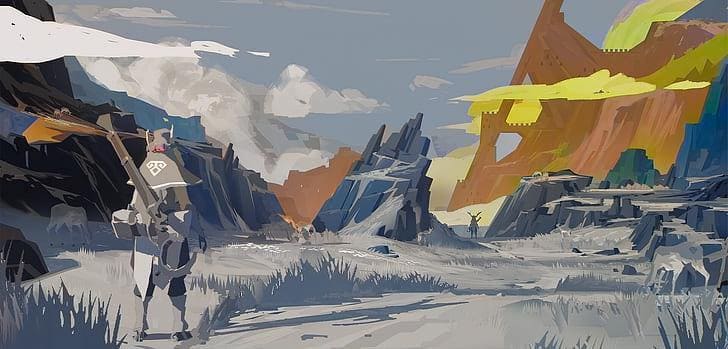} &
        \includegraphics[width=0.1111111111\textwidth]{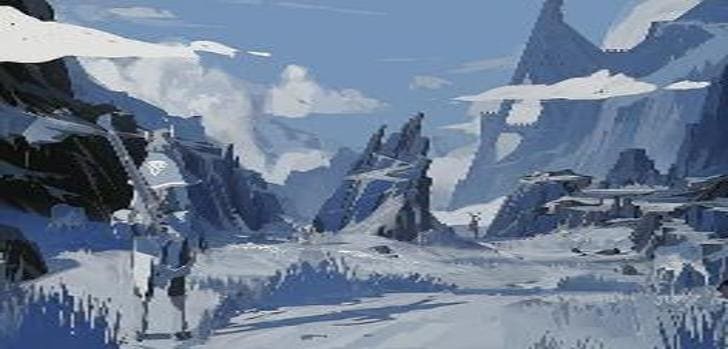} &
        \includegraphics[width=0.1111111111\textwidth]{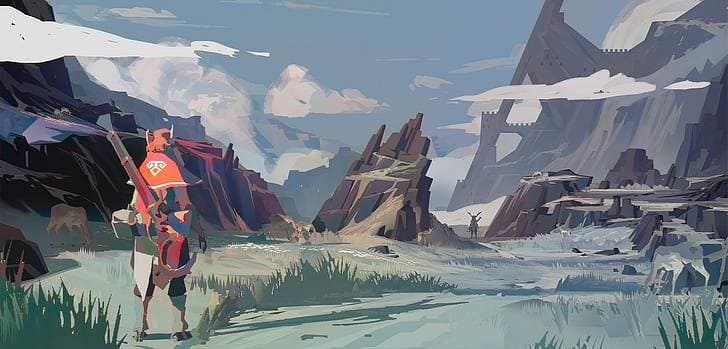} &
        \includegraphics[width=0.1111111111\textwidth]{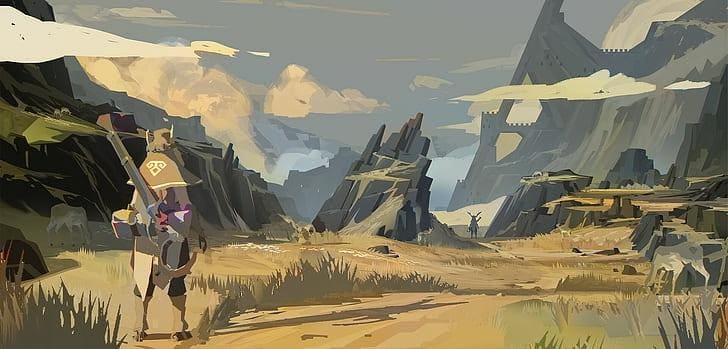} \\

        \includegraphics[width=0.1111111111\textwidth]{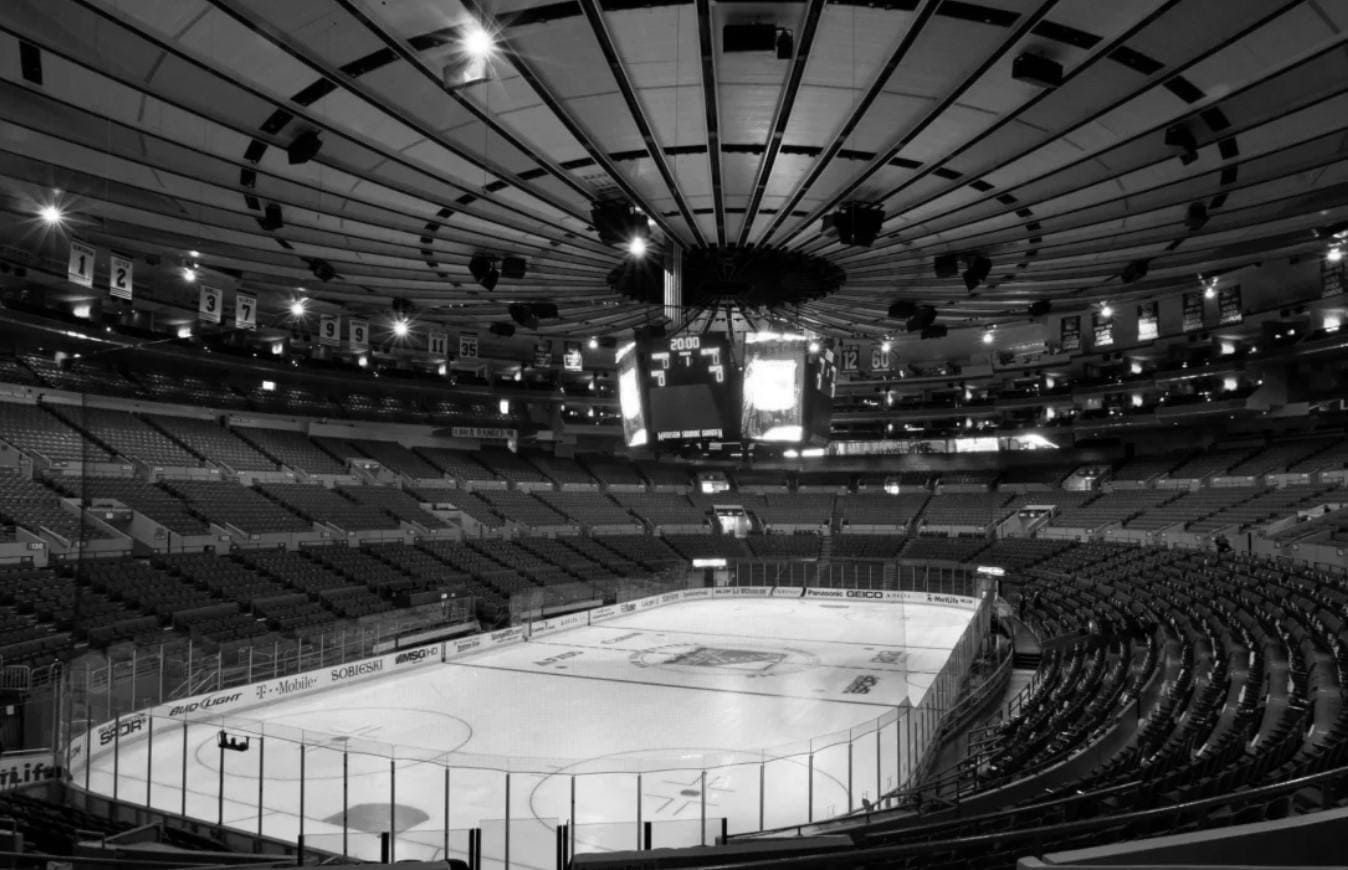} &
        \includegraphics[width=0.1111111111\textwidth]{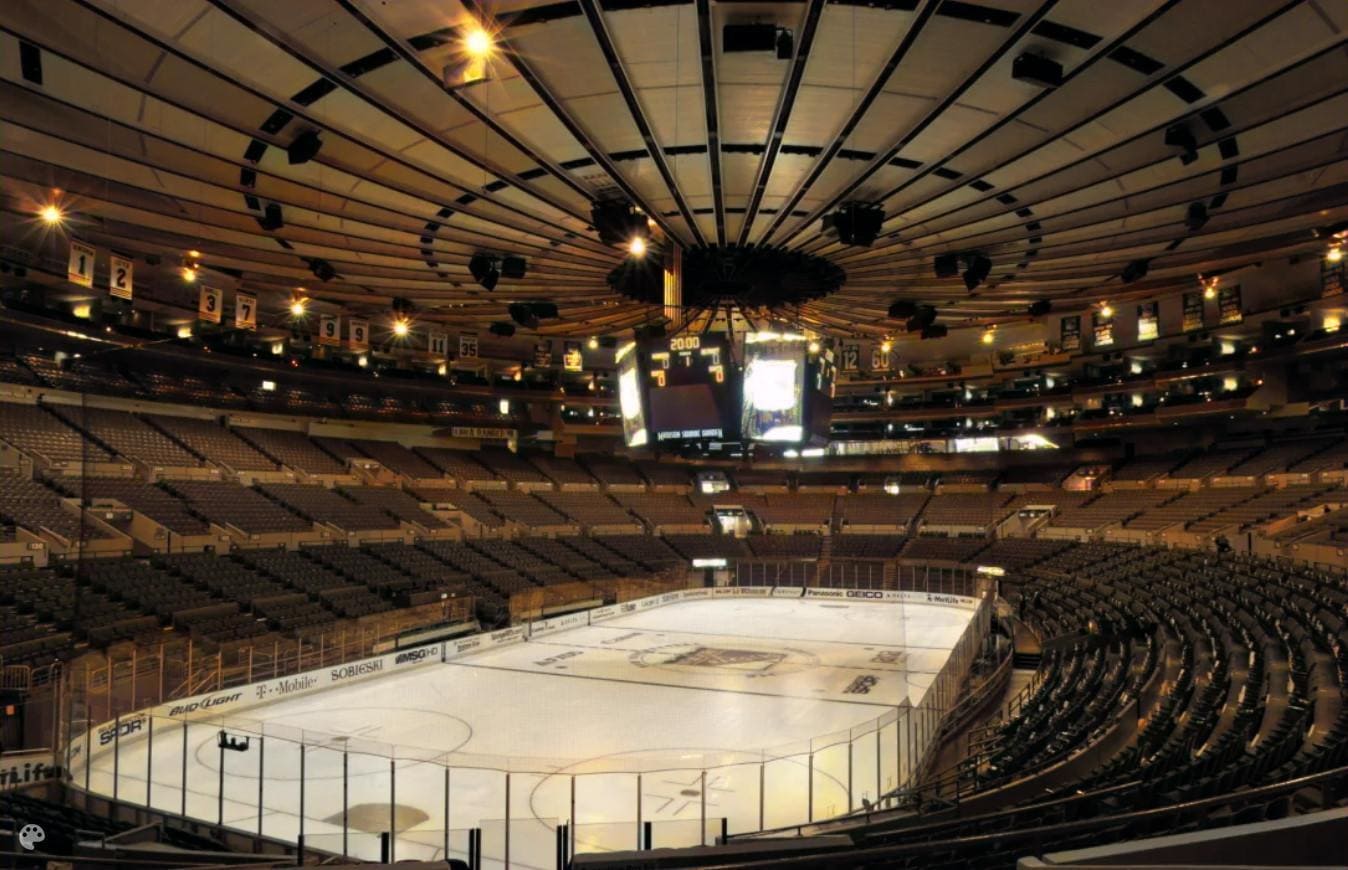} &
        \includegraphics[width=0.1111111111\textwidth]{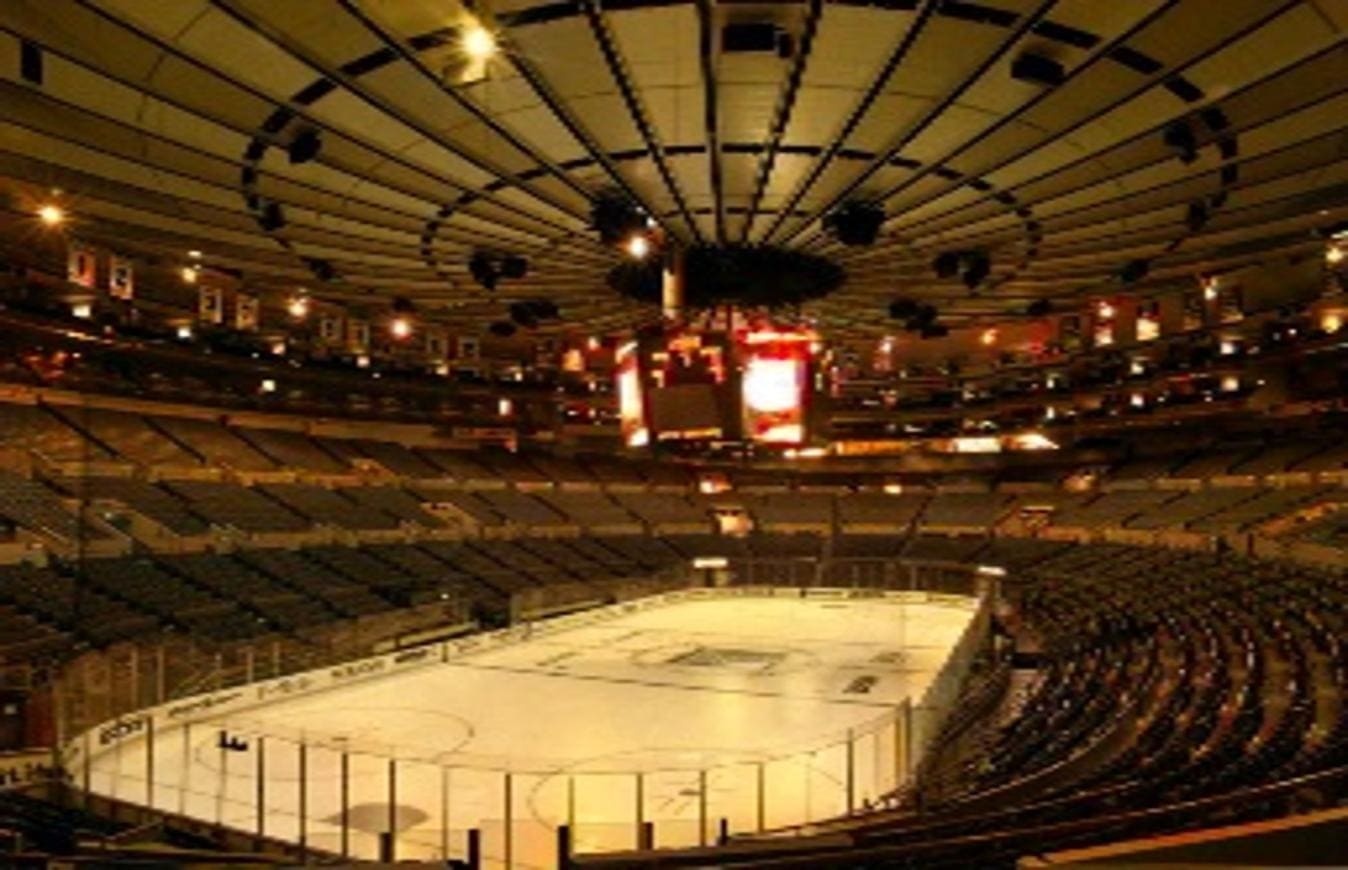} &
        \includegraphics[width=0.1111111111\textwidth]{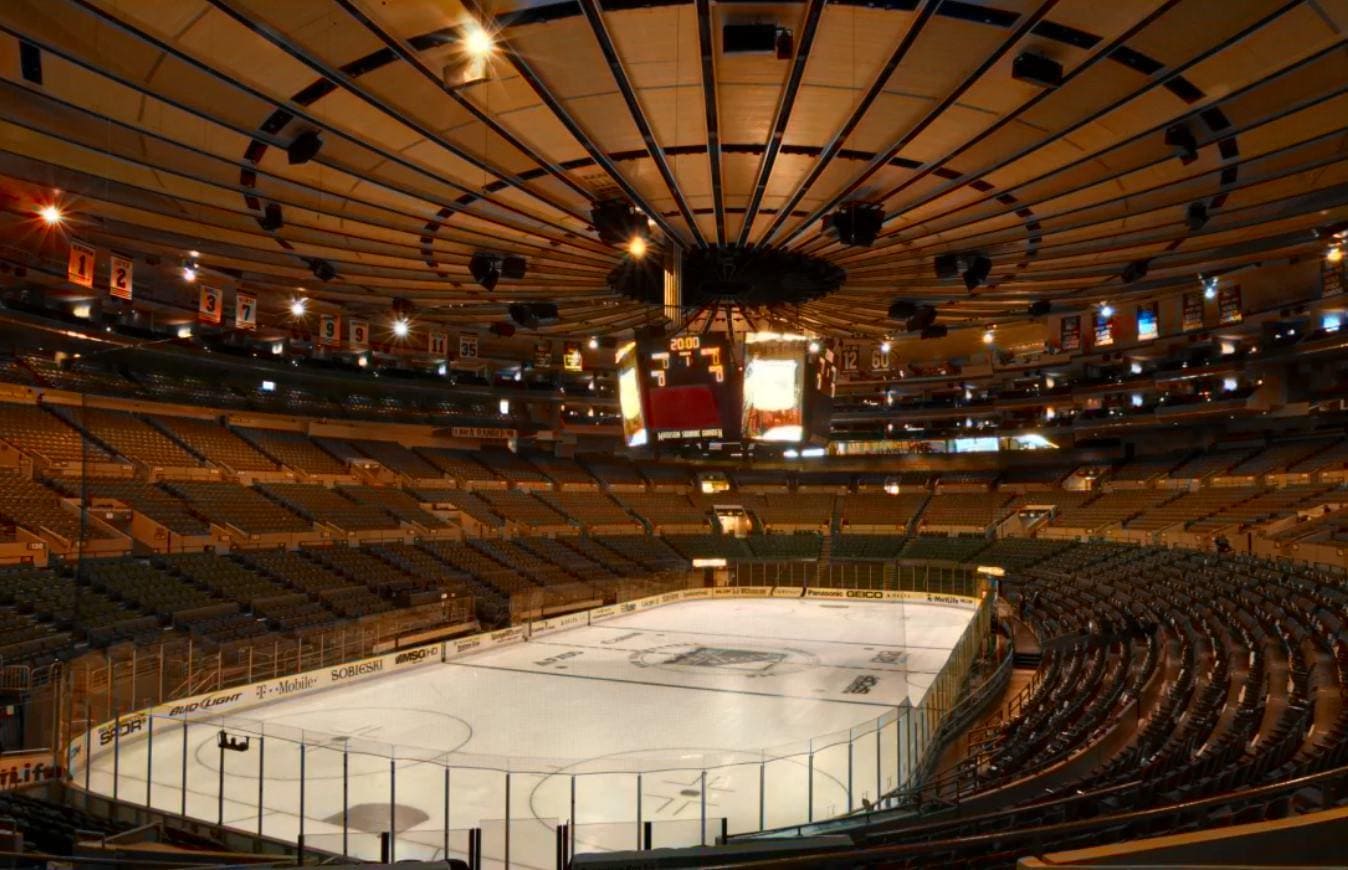} &
        \includegraphics[width=0.1111111111\textwidth]{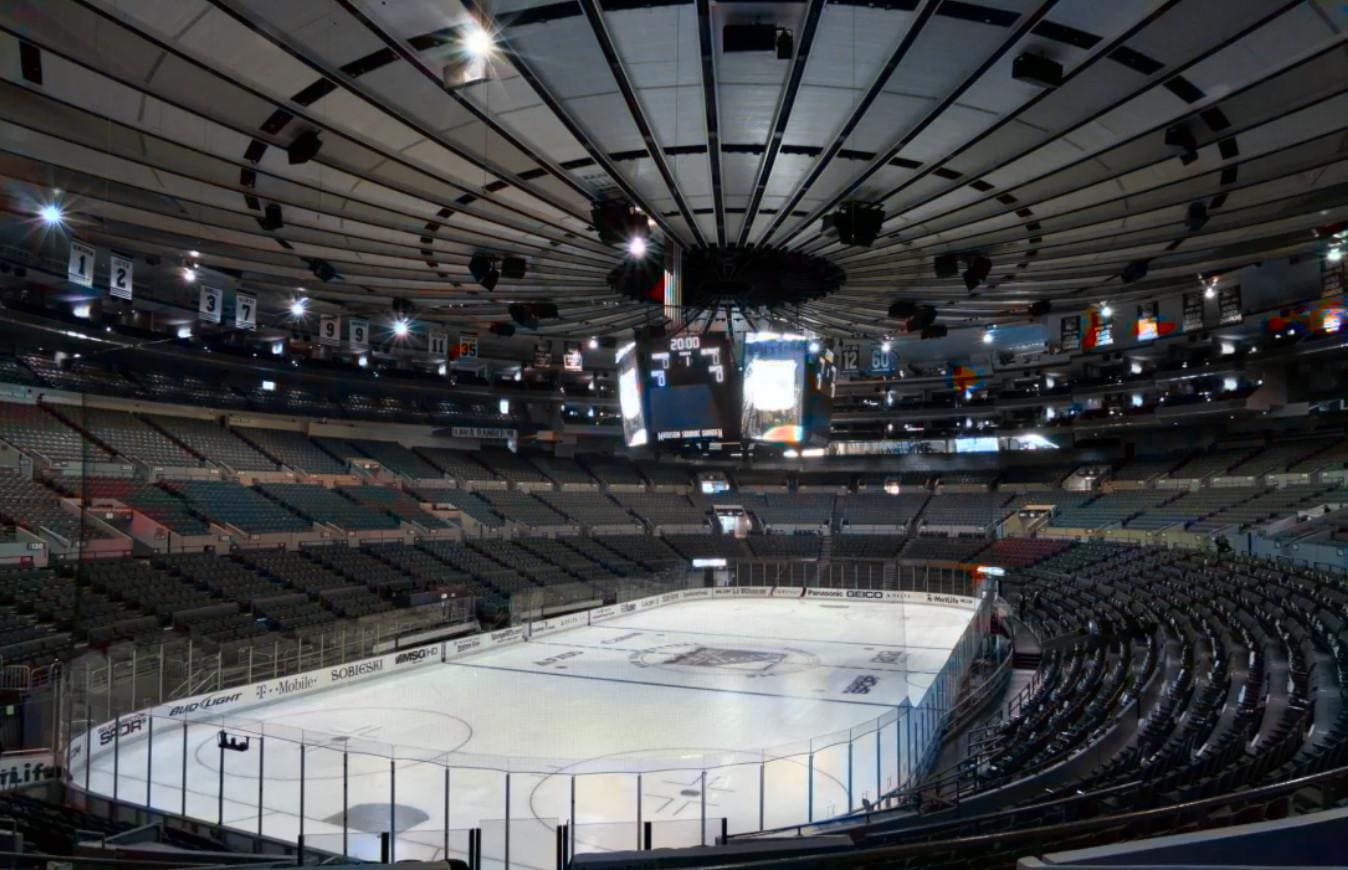} &
        \includegraphics[width=0.1111111111\textwidth]{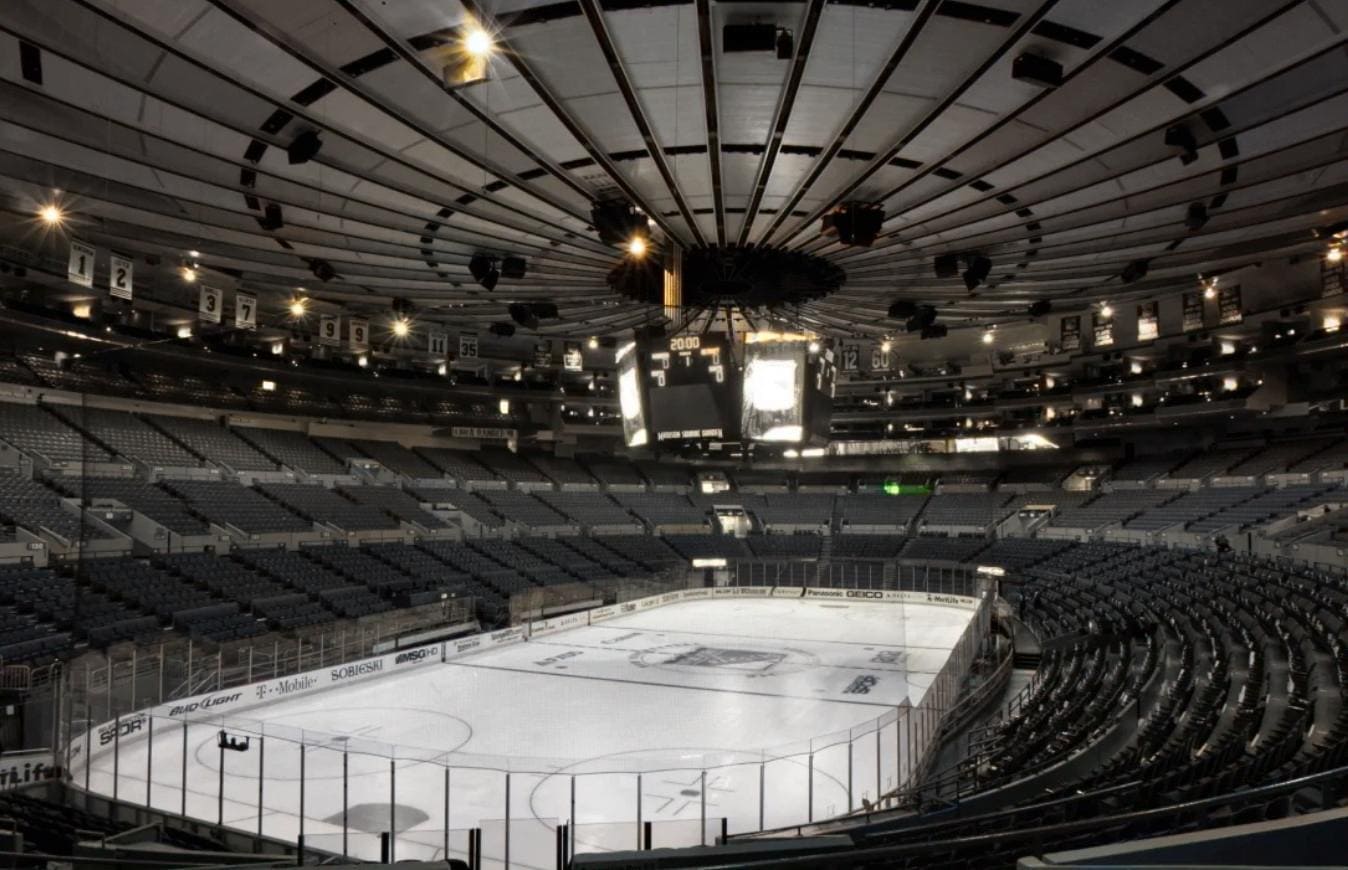} &
        \includegraphics[width=0.1111111111\textwidth]{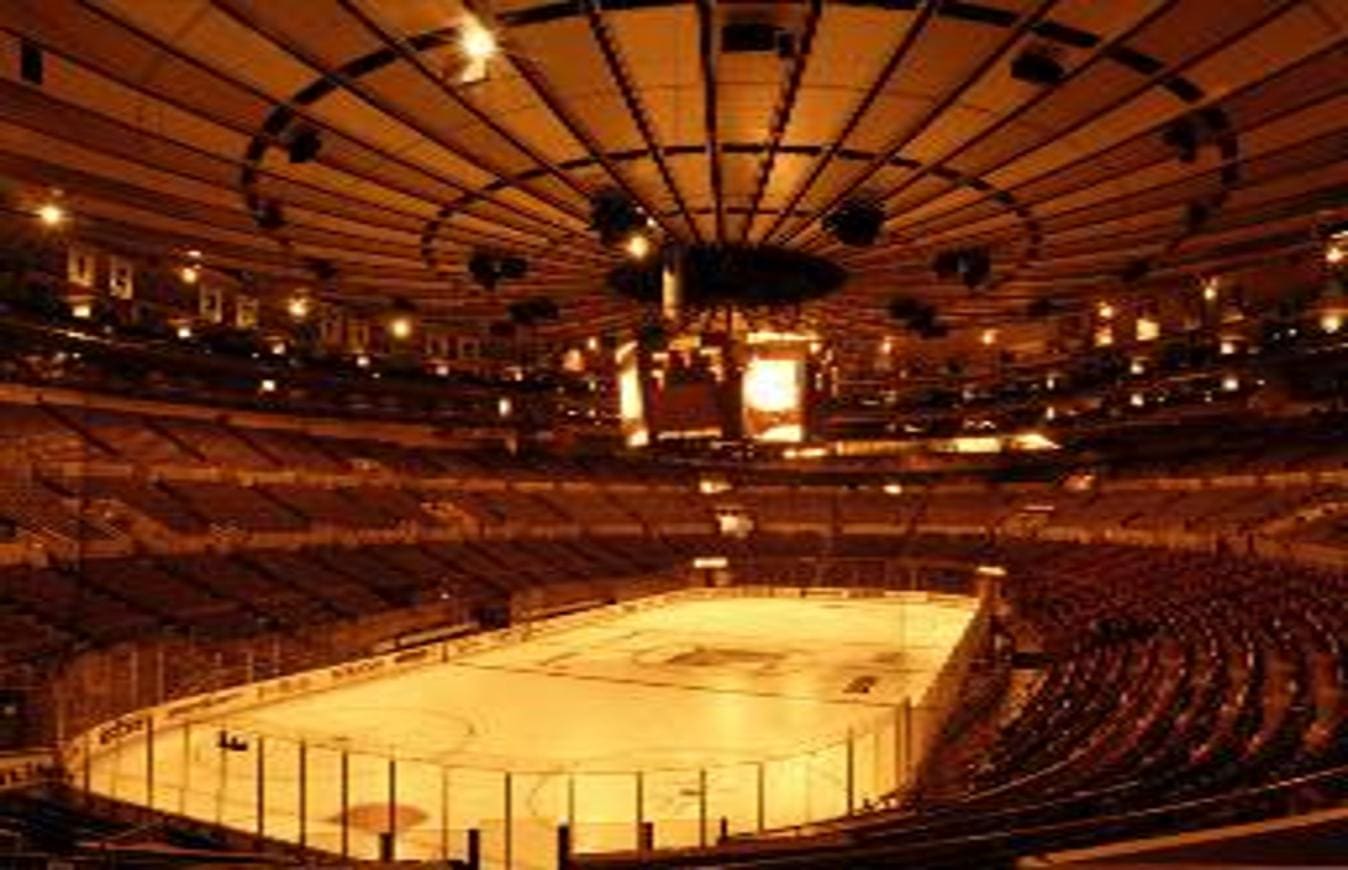} &
        \includegraphics[width=0.1111111111\textwidth]{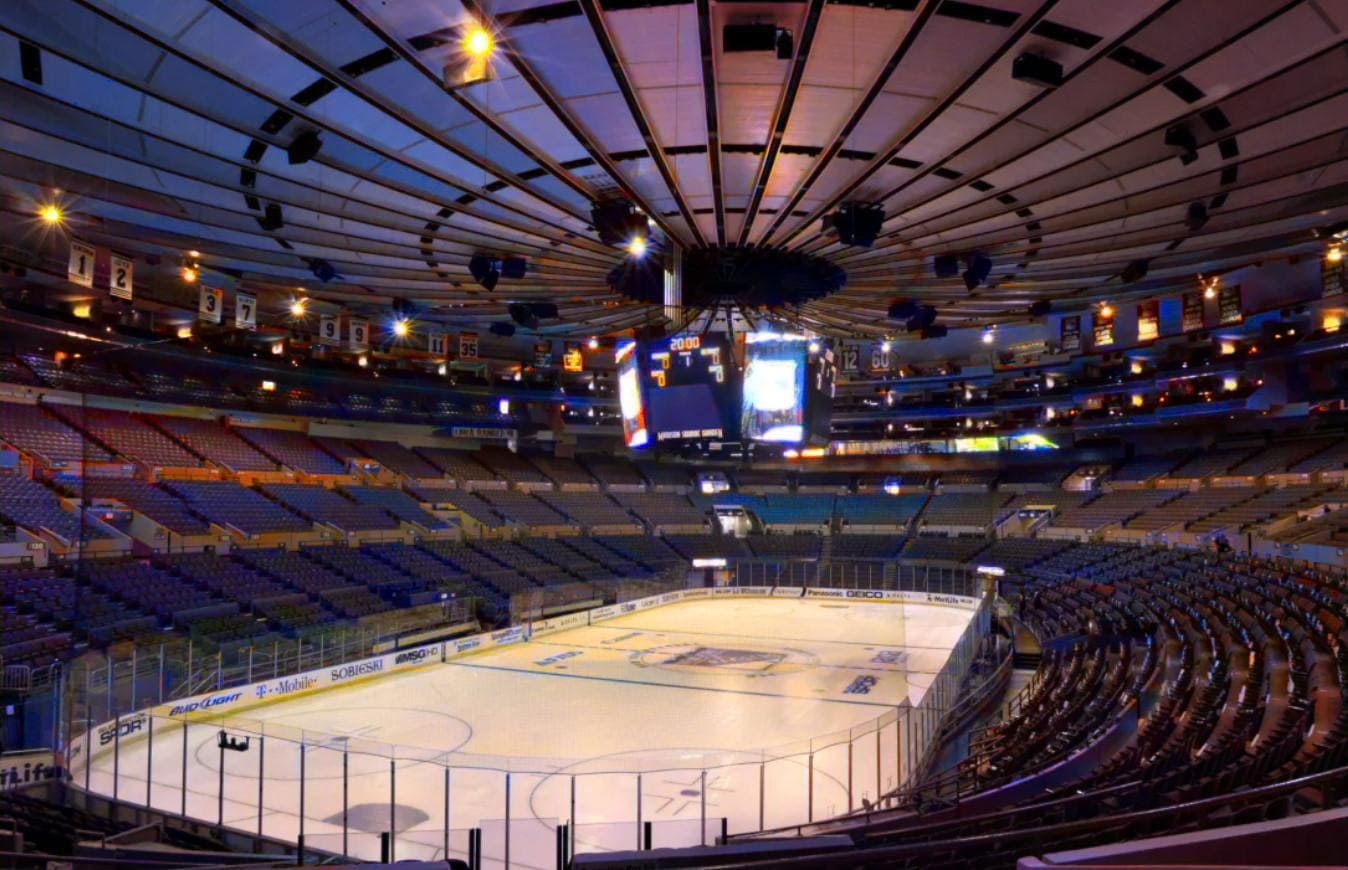} &
        \includegraphics[width=0.1111111111\textwidth]{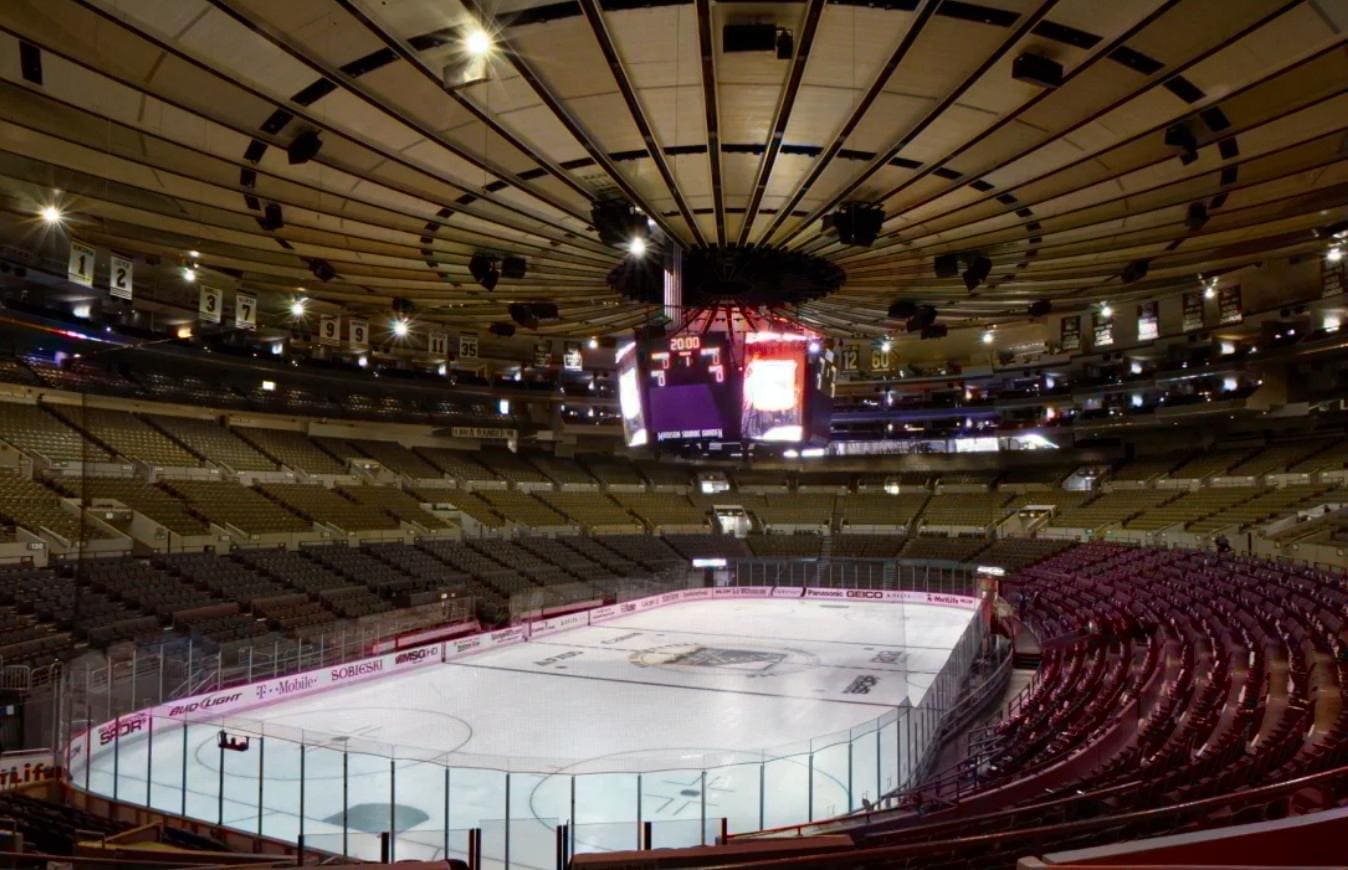}\\

        \includegraphics[width=0.1111111111\textwidth]{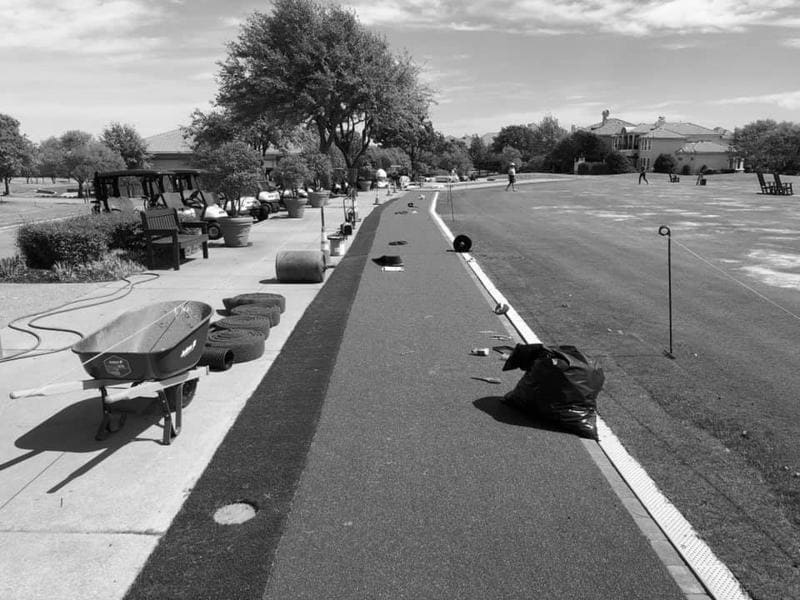} &
        \includegraphics[width=0.1111111111\textwidth]{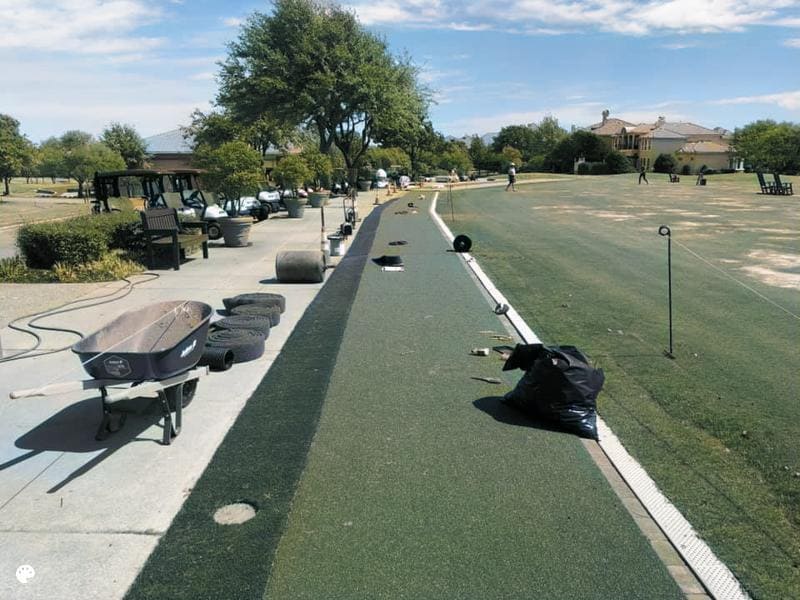} &
        \includegraphics[width=0.1111111111\textwidth]{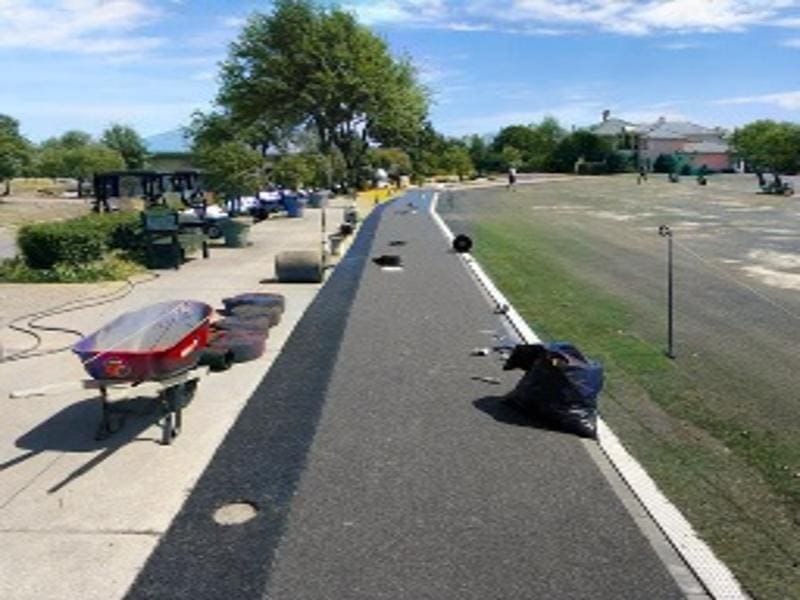} &
        \includegraphics[width=0.1111111111\textwidth]{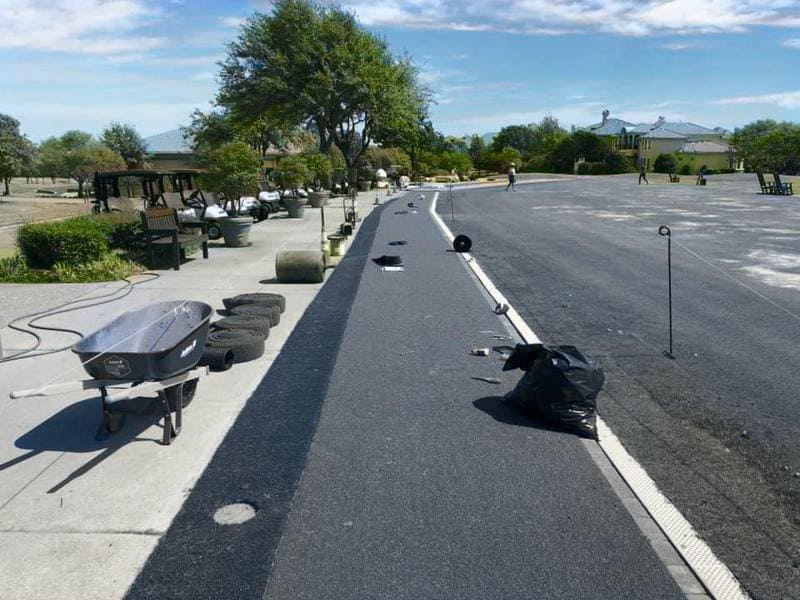} &
        \includegraphics[width=0.1111111111\textwidth]{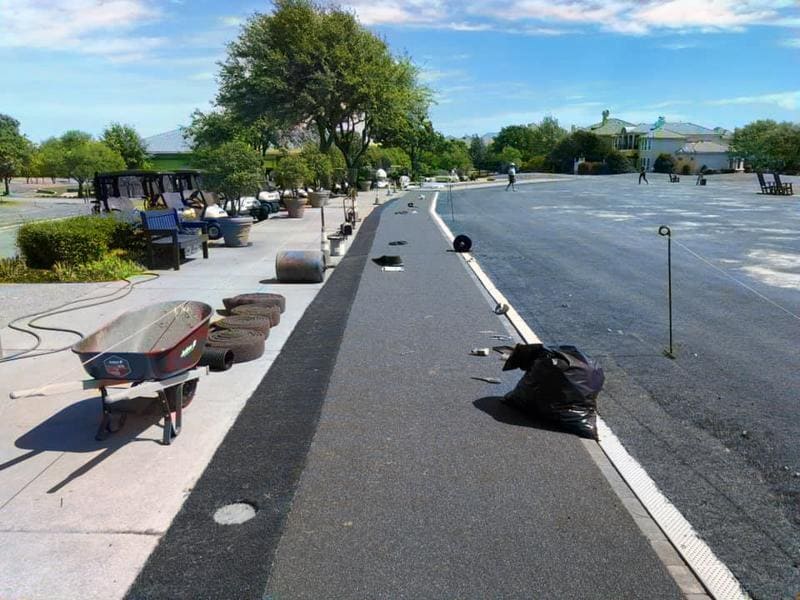} &
        \includegraphics[width=0.1111111111\textwidth]{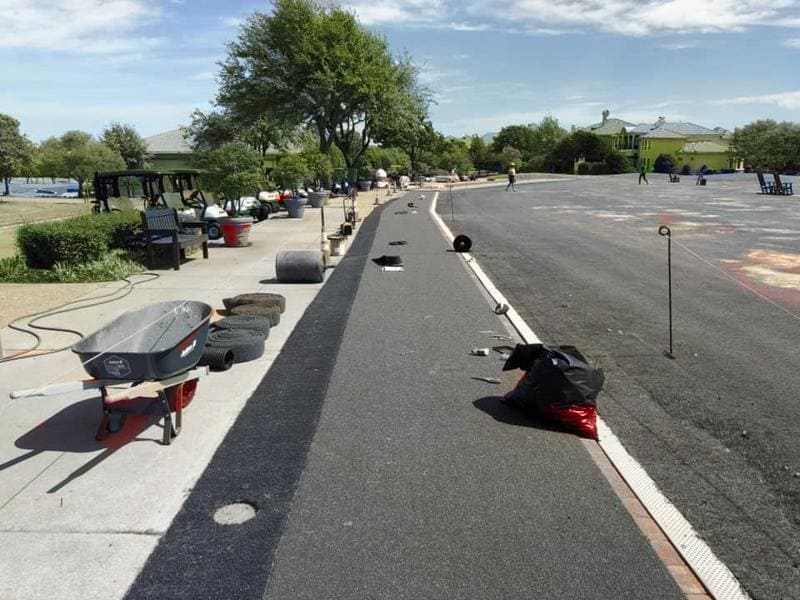} &
        \includegraphics[width=0.1111111111\textwidth]{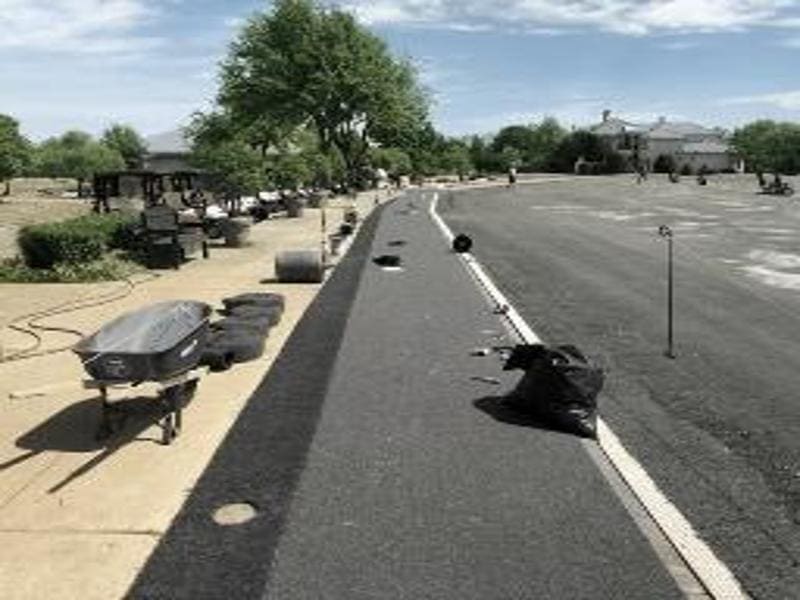} &
        \includegraphics[width=0.1111111111\textwidth]{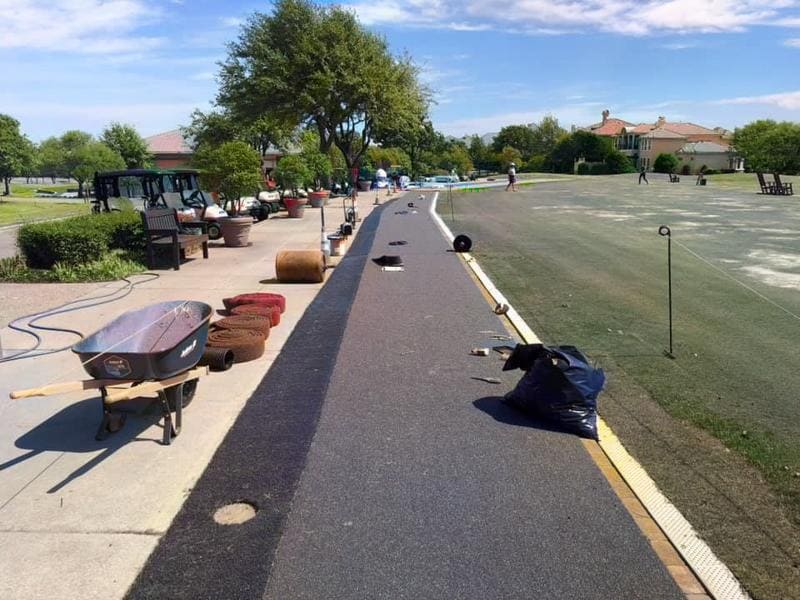} &
        \includegraphics[width=0.1111111111\textwidth]{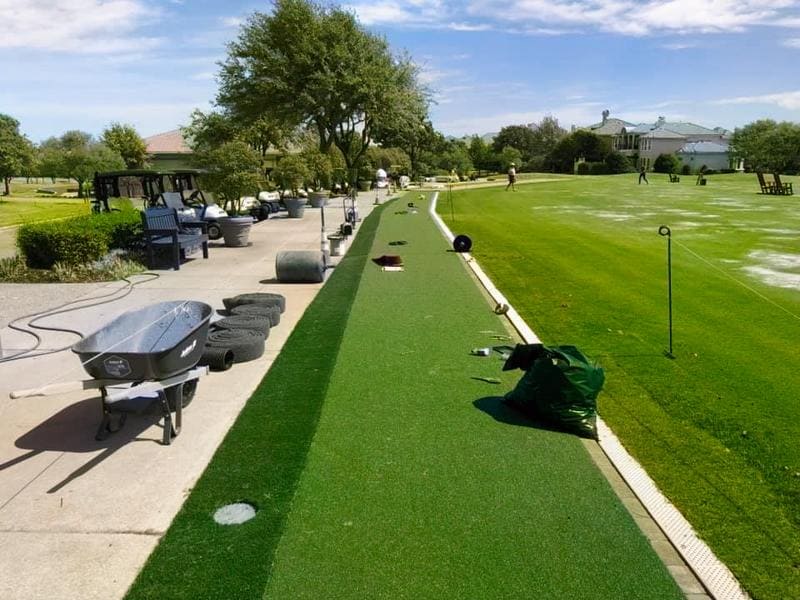}\\

        \multicolumn{1}{c}{\small\scalefont{0.7}Input} &
        \multicolumn{1}{c}{\small\scalefont{0.7}Deoldify~\cite{deoldify2019}} &
        \multicolumn{1}{c}{\small\scalefont{0.7}ColorFormer~\cite{10.1007/978-3-031-19787-1_2}} &
        \multicolumn{1}{c}{\small\scalefont{0.7}BigColor~\cite{Kim2022BigColor}} &
        \multicolumn{1}{c}{\small\scalefont{0.7}GCP~\cite{wu2021towards}} &
        \multicolumn{1}{c}{\small\scalefont{0.7}unicolor~\cite{huang2022unicolor}} &
        \multicolumn{1}{c}{\small\scalefont{0.7}DISCO~\cite{XiaHWW22}} &
        \multicolumn{1}{c}{\small\scalefont{0.7}DDColor~\cite{kang2022ddcolor}}&
        \multicolumn{1}{c}{\small\scalefont{0.7}Ours}
    \end{tabular}
    \caption{\textbf{Qualitative comparisons on the in-the-wild validation set.} Our method can generate significantly more natural and photo-realistic colors than state-of-the-art approaches. Zoom in for the best view.}
    \label{fig:qualitive_compare_in_the_wild}
\end{figure*}

\noindent \textbf{Baselines.} We adopt the most recent and related automatic colorization methods as baselines, including Deoldify~\cite{deoldify2019}, Su et al.~\cite{Su-CVPR-2020}, Coltran~\cite{kumar2021colorization}, GCP~\cite{wu2021towards}, CT2~\cite{CT2}, ColorFormer~\cite{10.1007/978-3-031-19787-1_2}, BigColor~\cite{Kim2022BigColor}, UniColor (Automatic)~\cite{huang2022unicolor}, DISCO~\cite{XiaHWW22} and DDColor~\cite{kang2022ddcolor}. In our experiments, we directly use the released evaluation results by Su et al~\cite{Su-CVPR-2020} and DISCO~\cite{XiaHWW22}, and we use automatic colorization results by Unicolor~\cite{huang2022unicolor} without user hints. For the other baselines, we compare using their official codes and weights.

\noindent \textbf{Condition Setting.}
We provide an ablation study in Table~\ref{tab:conditions} on the condition setup for ControlNet~\cite{zhang2023adding}. Various feature maps can be used as conditions for ControlNet, with Canny and HED maps demonstrating superior performance. Notably, the combination of Canny and HED map conditions yields optimal results, prompting our choice of this configuration.

\begin{table}[h!]
\centering
\fontsize{6pt}{7pt}\selectfont
\begin{tblr}{
  colspec = {Q[c] Q[c] Q[c] Q[c] Q[c] Q[c]},
  hlines,
}
Condition setting & Canny & HED & Depth & Semantic & Canny\&HED \\
CF $\uparrow$ & \underline{46.99} & \underline{45.28} & 42.81 & 45.02 & \textbf{47.18}
\end{tblr}
\vspace{-0.05in}
\caption{\textbf{Ablation study on the condition setup.} We generate $N=8$ reference candidates for each condition setting. The test images are from the COCO-stuff validation dataset.}
\label{tab:conditions}
\end{table}

\noindent \textbf{Hyper-parameters Setup.}
We provide an ablation study in Table~\ref{tab:hyper_ablation} on the hyper-parameter choice: the number of reference candidates $N$, the number of segments $N_{\mathbf{S}}$, and the threshold of hint colors $\mathcal{N}_{\mathcal{H}}$. We empirically select the hyper-parameters as $N=8$, $N_{\mathbf{S}}=10$, and $\mathcal{N}_{\mathcal{H}} = 10$ based on the balance of performance and running efficiency.

\begin{table}[h!]
\centering
\fontsize{6pt}{7pt}\selectfont
\begin{tblr}{
  colspec = {Q[c] Q[c] Q[c] Q[c] Q[c] Q[c] Q[c] Q[c]},
  hlines,
}

N & 2 & 4 & 8 & 16 & \textbf{32} & 8 & 8 \\
$N_S$ & 10 & 10 & 10 & 10 & \textbf{10} & 40 & 10 \\
$\mathcal{N}_{\mathcal{H}}$ & 10 & 10 & 10  & 10 & \textbf{10} & 10 & 50 \\
CF $\uparrow$ & 38.81 & 43.13 & 47.18 & 47.82 & \textbf{48.69} & 42.09 & 45.53 \\

\end{tblr}
\vspace{-0.05in}
\caption{\textbf{Ablation study on the hyper-parameters choice: N, $N_S$ and $\mathcal{N}_{\mathcal{H}}$.} The test images are from the COCO-stuff validation dataset.}
\label{tab:hyper_ablation}
\end{table}

\subsection{Comparisons to Baselines}

\noindent \textbf{Qualitative Comparisons.} We show the qualitative comparisons in Figure \ref{fig:qualitive_compare_coco} and Figure~\ref{fig:qualitive_compare_in_the_wild}. Note that the test images are from COCO-stuff validation set and $In\text{-}the\text{-}wild$ set respectively. Our colorization results are more natural, photorealistic, vivid, and visually pleasing than baselines in both scenarios, especially on the $In\text{-}the\text{-}wild$ set, significantly alleviating artifacts and preserving color consistency. Since there might be quite a lot of $unseen$ color distributions in $In\text{-}the\text{-}wild$ set, state-of-the-art baselines suffer from color bleeding, grayish colorization, color inconsistency, and visual artifacts. In contrast, our imagination module $\mathcal{I}$ can provide a wide range of reference candidates, exhibiting sufficient color diversity and strong colorization robustness. Our framework has taken a significant leap forward, to some extent, achieving a universal off-the-shelf colorization application. More comparisons are available in the supplementary. 

\noindent \textbf{Quantitative Comparisons.} We show quantitative comparisons in Table~\ref{tab:Quantitative_Comparison}. Considering the color multimodality in the colorization task, the ground truth based evaluation metrics PSNR, SSIM, and LPIPS can not accurately reflect the actual performance, for there might exist huge differences between the colorful and plausible colorization results and the ground truth. Our framework shows the best quantitative results in \textbf{CF}, indicating that we can achieve the most colorful and perceptually photorealistic colorization results.

\begin{figure}[t]
  \centering
  \vspace{0in}
  \includegraphics[width=0.9\linewidth]{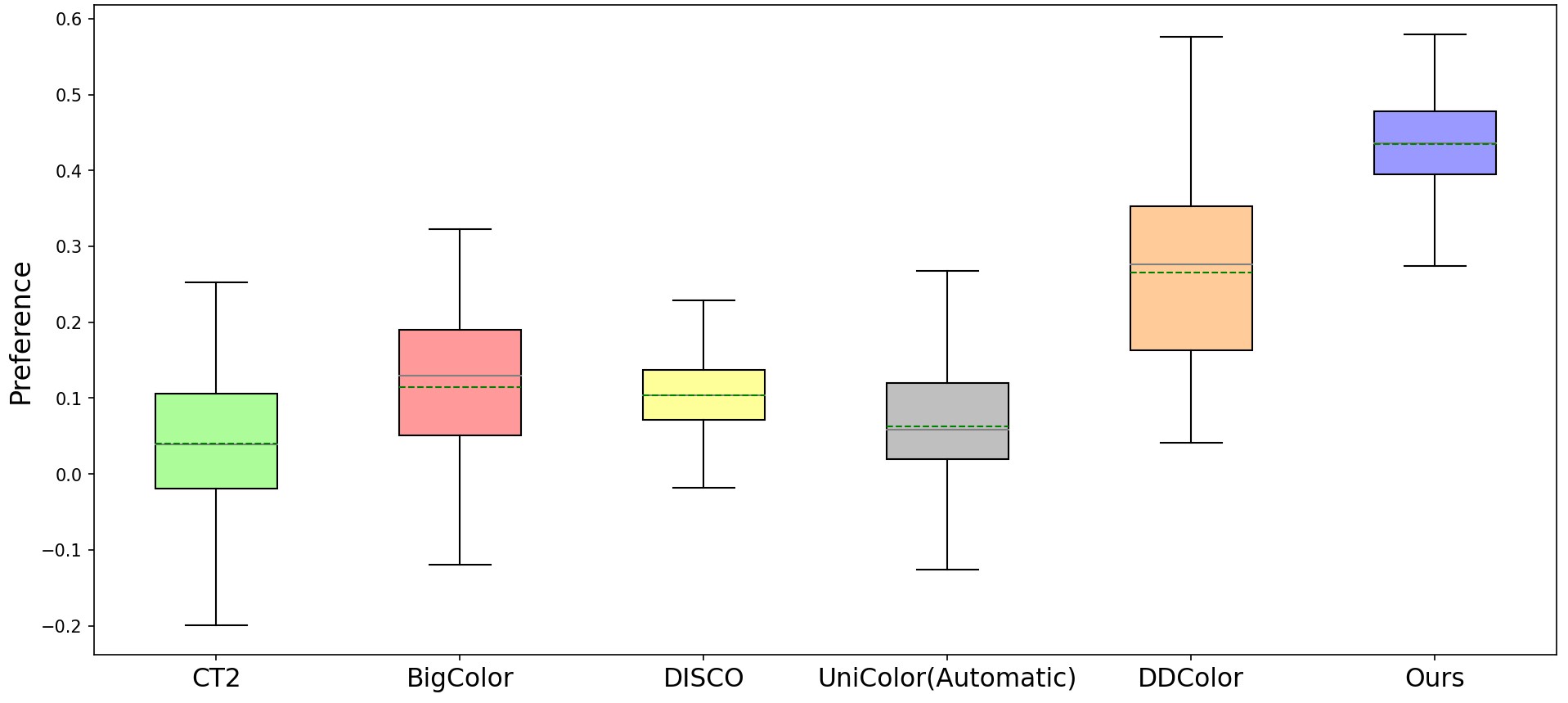}
  \caption{\textbf{User study.} The dashed green lines are the mean preference percentage.}
  \label{fig:user_study}
\end{figure}

\vspace{-0.2in}
\paragraph{User study.}
We conduct a user study to evaluate which approach is more preferred by human observers. Specifically, we compare our approach with five strong baselines: DDColor~\cite{kang2022ddcolor}, DISCO~\cite{XiaHWW22}, UniColor (Automatic)~\cite{huang2022unicolor}, Bigcolor~\cite{Kim2022BigColor}, CT2~\cite{CT2}. 30 Users are presented with five colorized images (our approach and one from each of the four baselines) simultaneously in a single view, with the positions randomized for each comparison. They are then tasked with selecting the most realistic and colorful one among them. We randomly select 30 images from the COCO-stuff validation set, 30 images from the ImageNet validation set, and 20 images sourced from the internet. As shown in Figure \ref{fig:user_study}, Our approach outperforms DDColor~\cite{kang2022ddcolor}, DISCO~\cite{XiaHWW22}, UniColor (Automatic)~\cite{huang2022unicolor}, Bigcolor~\cite{Kim2022BigColor}, CT2~\cite{CT2} with a preferred rate of $43.3\%, 26.7\%, 10.0\%, 6.3\%, 10.0\%, 3.3\%$.

\begin{table*}
\centering
\fontsize{8pt}{9pt}\selectfont
\begin{tblr}{
  colspec = {lccccccccccc},
  cell{1}{1} = {r=2}{},
  cell{1}{2} = {c=5}{},
  cell{1}{7} = {c=5}{},
  vline{1-2,13} = {1-2}{},
  vline{7,12-13} = {1-2}{},
  vline{1-2,7,12-13} = {3-11}{},
  hline{1,3,11-12} = {-}{},
  hline{2} = {2-12}{},
}
Method      & COCO-Stuff &      &      &       &    & ImageNet &      &         &    &    & In-the-wild \\
            & FID $\downarrow$  & SSIM $\uparrow$ &  PSNR $\uparrow$ & LPIPS $\downarrow$ &  \underline{\textbf{CF}} $\uparrow$ & FID $\downarrow$  & SSIM $\uparrow$ &  PSNR $\uparrow$ & LPIPS $\downarrow$ &  \underline{\textbf{CF}} $\uparrow$  & \underline{\textbf{CF}} $\uparrow$          \\
Coltran~\cite{kumar2021colorization}      &    13.1         &  0.359    &   13.3   &   0.513    &  38.5  &   15.5   &   0.264   &   8.65    & 0.723   &  \textbf{55.9}  &         38.0    \\
GCP~\cite{wu2021towards}         &      7.35      &   0.900   &  22.4    &   0.194    &  37.9  &  3.36  &     0.931     &    23.5   &  0.238  &  33.9  &     37.0        \\
CT2~\cite{CT2}          &      22.9      &   0.358   &   13.5   &  0.518     &  46.2  &  8.44   &    0.354      &   14.5     & 0.480   &  41.6  &        44.5     \\
ColorFormer~\cite{10.1007/978-3-031-19787-1_2} &      8.69      & 0.756     &   21.3   &  0.216     & 41.0   &   3.83    &   0.834   &   22.5    &  0.196  &  36.0  &        39.9     \\
BigColor~\cite{Kim2022BigColor}    &      8.53      &  0.832   &  20.8    &    0.217   &  43.4  &  3.59   &  0.893    &    21.5   &  0.212  &  40.4  &     37.8        \\
UniColor~\cite{huang2022unicolor}     &   7.90         &  0.855    &    22.4  &   0.195    &  36.4  &   6.22    &  0.909    &    22.0   &  0.238  &  35.7  &   33.9          \\
DISCO~\cite{XiaHWW22}        &     11.2       &  0.738    &   19.4   &     0.236  &  46.2  &    9.21   &   0.783   &    21.0   &   0.265 &  44.5  &    43.5    \\
DDColor~\cite{kang2022ddcolor}      &      6.19      &   0.904  &  23.1    &   0.169    &  44.7  &     3.16   &  0.885    &  23.4     &  0.186  &  42.2  &    41.9             \\
Ours        &       7.21     &   0.859  &   23.3   &    0.180   &  \textbf{47.2}  &  3.62  &      0.884    &     23.8     &  0.207  &  \underline{48.8}  &      \textbf{48.9}
\end{tblr}
\caption{\textbf{Quantitative Comparison with baselines on three datasets.} We provide evaluations on PSNR, SSIM, and LPIPS just for reference because those metrics can not accurately reflect the actual performance. Our framework shows the best quantitative results in \textbf{CF}, indicating we can generate the most colorful and perceptually photorealistic results.}
\vspace{-0.1in}
\label{tab:Quantitative_Comparison}
\end{table*}

\subsection{Ablation Study}

\begin{figure}[t]
  \centering
  \includegraphics[width=1\linewidth]{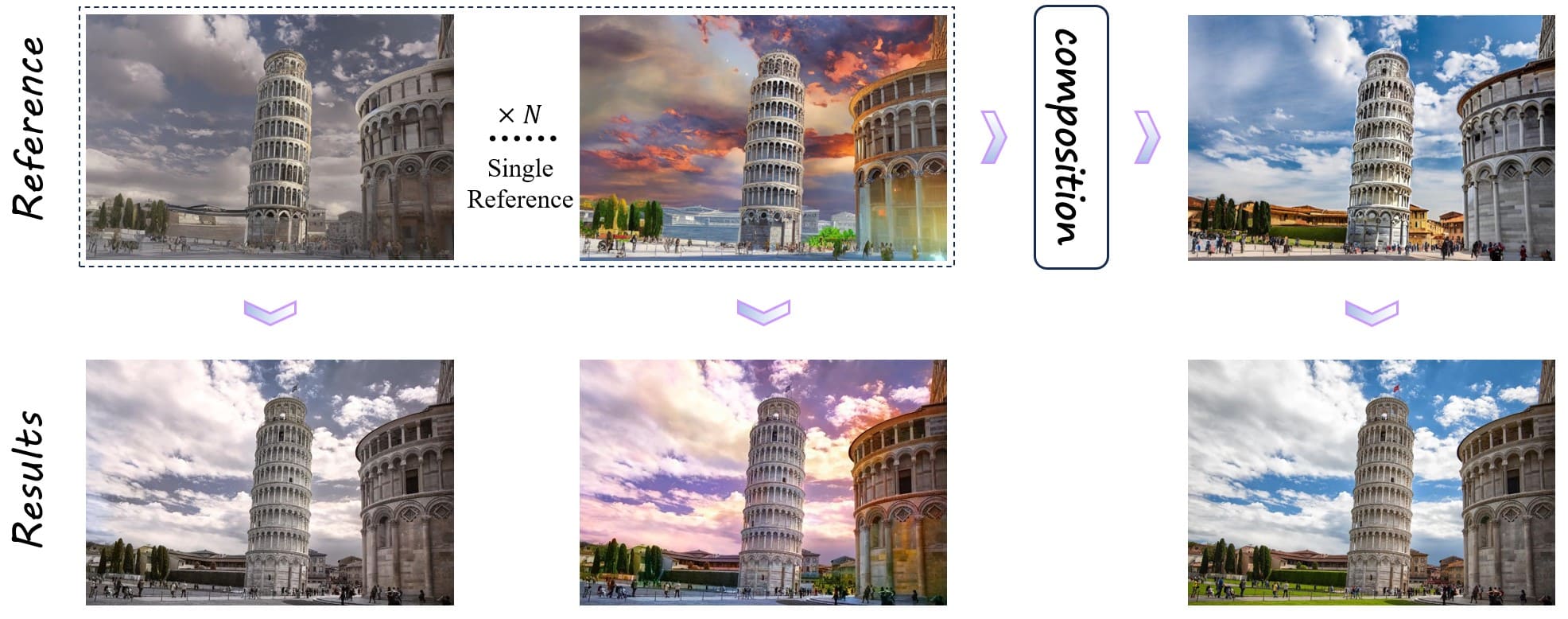}
  \vspace{-0.2in}
  \caption{\textbf{Ablation Study: Reference Refinement Module \ref{subsec:Reference-Refinement-Module}.} Without our composition strategy, the diversity and completeness of the color space are compromised. Consequently, the reference exhibits inconsistency, visual artifacts, and color bleeding, leading to unsatisfactory colorization results. }
  \label{fig:refinement_module}
\end{figure}

\paragraph{Reference refinement module.}
Figure \ref{fig:refinement_module} illustrates the effectiveness of our Reference Refinement Module (see Section \ref{subsec:Reference-Refinement-Module}), which can significantly improve the colorfulness and photorealism of colorization results. In the top row, reference candidates are composed to an optimal reference, with ‘×N’ indicating there are N candidates in all. In the bottom row, we showcase colorization results guided by two randomly selected reference candidates and the optimal reference. Upon comparison, Without our composition strategy, the diversity and completeness of the color space are compromised. ControlNet~\cite{zhang2023adding} struggles to generate a comprehensive color space from a single reference. Consequently, eliminating the composition strategy compromises the completeness of this color space, leading to inconsistencies, visual artifacts, and color bleeding. Quantitatively, Table \ref{tab:abalation_composition} further highlights a marked performance difference when using our Reference Refinement Module \ref{subsec:Reference-Refinement-Module} compared to not using it.

\noindent \textbf{Diverse Colorization.}
Thanks to our imagination module \ref{subsec:Imagination-Module}, we can synthesize multiple reasonable and colorful references, yielding diverse colorization results, as shown in Figure \ref{fig:Diverse_Colorization}.

\noindent \textbf{Hint Colors Optimization.}
shows that the hint colors ambiguity issue can arise in the absence of a coarse-to-fine optimization strategy. As depicted in Figure \ref{fig:ambiguity}, the reference's sky exhibits a substantial variation in hint color values, resulting in color bleeding and incorrect colorization. By employing a clustering algorithm, our coarse-to-fine optimization identifies the most representative set of hint colors, which reduces color bleeding and ensures a pleasing colorization result.

\subsection{Limitation and Discussion}
As depicted in Figure \ref{fig:failure_cases}, our approach encounters challenges when dealing with images populated with numerous identical instances. We attribute such cases to the inherent limitations of existing generative models like ControlNet which struggle with generating diverse and photo-realistic images with numerous distinct humans. Thus, we believe our method will address such cases with more advanced generative models emerging. While our paper is designed for image colorization, it can be combined with other post-processing methods~\cite{lai2018learning,lei2020blind,lei2022deep,lei2023blind} to achieve temporally consistent video colorization. Besides, further enhancement may require the colorization framework to achieve an equilibrium: ensure a consistent color prior for a whole semantic area, while still recognizing and respecting the unique color cues of each instance within.

The inference cost is another concern. After accelerating ControlNet with LCM-LoRA~\cite{luo2023latent}, our model takes about eight seconds to colorize a $512\times 512$ image on a single Nvidia GeForce RTX 3090. We believe using a more efficient generative model can reduce the inference time in the future.


%% file: 10_conclusion.tex
\begin{figure}[t]
\centering
\begin{tabular}{@{}c@{\hspace{0.5mm}}c@{\hspace{0.5mm}}c@{}}

\includegraphics[width=0.33\columnwidth]{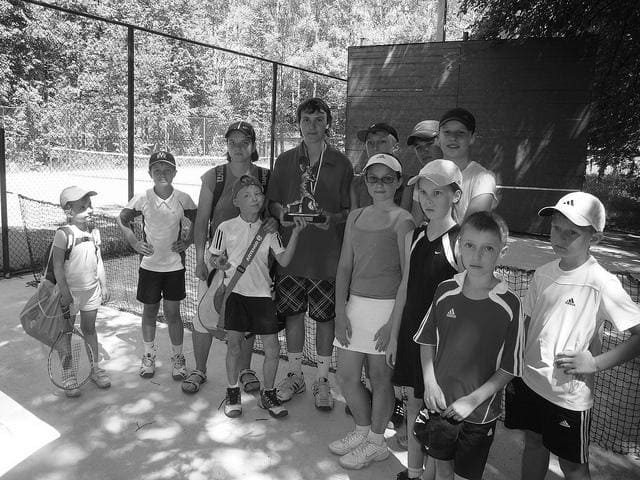}&
\includegraphics[width=0.33\columnwidth]{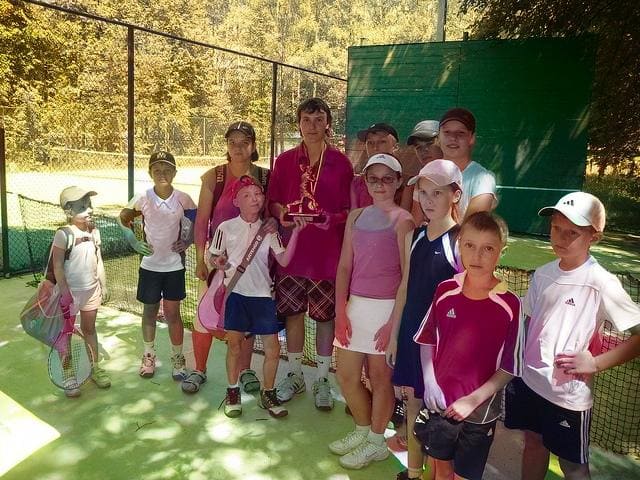}&
\includegraphics[width=0.33\linewidth]{figs/failure_cases/gt_low_res.jpg}\\
\begin{minipage}{0.33\columnwidth}
\centering
\footnotesize Input
\end{minipage} &
\begin{minipage}{0.33\columnwidth}
\centering
\footnotesize Ours
\end{minipage} &
\begin{minipage}{0.33\columnwidth}
\centering
\footnotesize Ground truth
\end{minipage} \\
\end{tabular}
 \vspace{-0.1in}
\caption{\textbf{Failure Cases.} Our approach might produce artifacts and inconsistencies when colorizing images with too many identical instances.}
 \label{fig:failure_cases}
\end{figure}

\begin{table}[t]
\begin{center}
\resizebox{1.0\linewidth}{!}{
\begin{tabular}{l@{\hspace{3em}}cccccc}
\toprule
\multicolumn{1}{l}{reference} & FID $\downarrow$ &  CF $\uparrow$  & SSIM $\uparrow$ &  PSNR $\uparrow$ & LPIPS $\downarrow$ \\
\midrule
composition & \textbf{7.21} &  \textbf{47.18} &  \textbf{0.859} &  \textbf{23.3} & \textbf{0.180}\\
single & 8.28 & 43.87 & 0.848 & 21.2 & 0.230\\
\bottomrule
\end{tabular}
}
\end{center}
\vspace{-0.1in}
\caption{\textbf{Ablation on without and with composition strategy.} All the tested images are from COCO-Stuff validation set. Our composition strategy in the Reference Refinement Module plays a pivotal role in improving colorization results from quantitative perspectives.}
\label{tab:abalation_composition}
\end{table}

\section{Conclusion}
\label{sec:Conclusion}

In this paper, we present a novel framework for automatic image colorization that emulates the imagination process of human experts. Our primary contribution is the Imagination Module and the Reference Refinement Module, which together generate reference images that are semantically similar, structurally aligned, and instance-aware. The Colorization Module then colorizes the black-and-white inputs with the guidance of these reference images. Our framework surpasses state-of-the-art baselines both qualitatively and quantitatively, achieving controllable, editable, and diverse colorization results, which is non-trivial in the automatic colorization community. We posit that the concept of imagination has the potential to enhance a variety of computer vision tasks.

\paragraph{Acknowledgements} 
This work was supported by the InnoHK program.

%% file: 12_appendix.tex

\section{Visualization of reference synthsis}
\begin{figure*}[t]
    \centering
    \setlength{\tabcolsep}{0pt} 
    \renewcommand{\arraystretch}{0.5} 
    \begin{tabular}{*{6}{p{0.16\textwidth}}} 
        \includegraphics[width=0.166\textwidth]{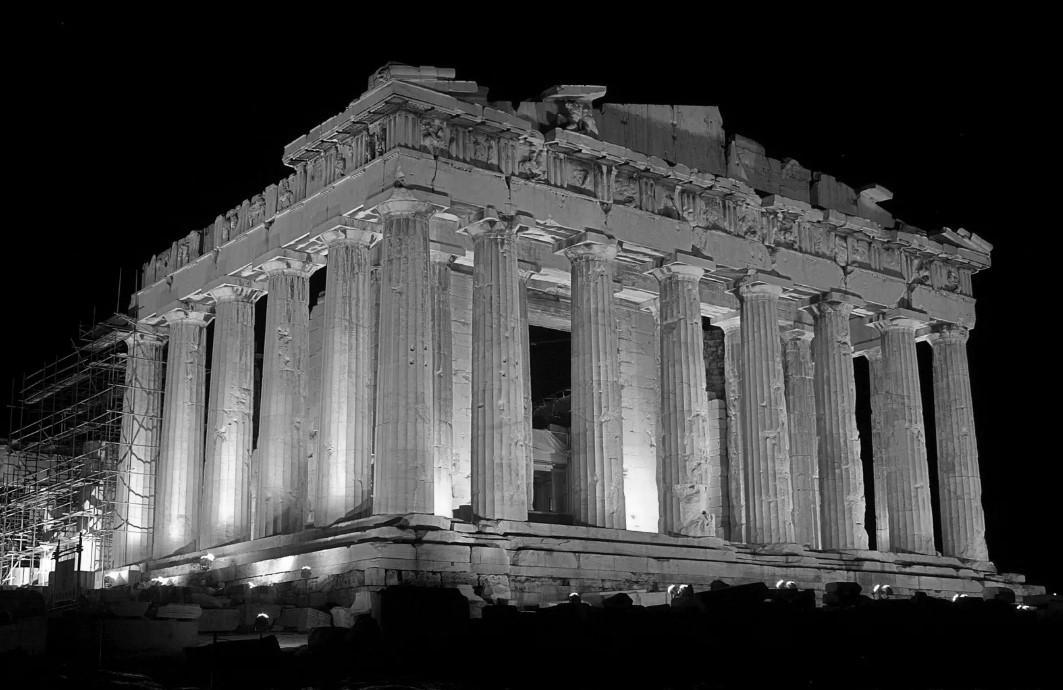} &
        \includegraphics[width=0.166\textwidth]{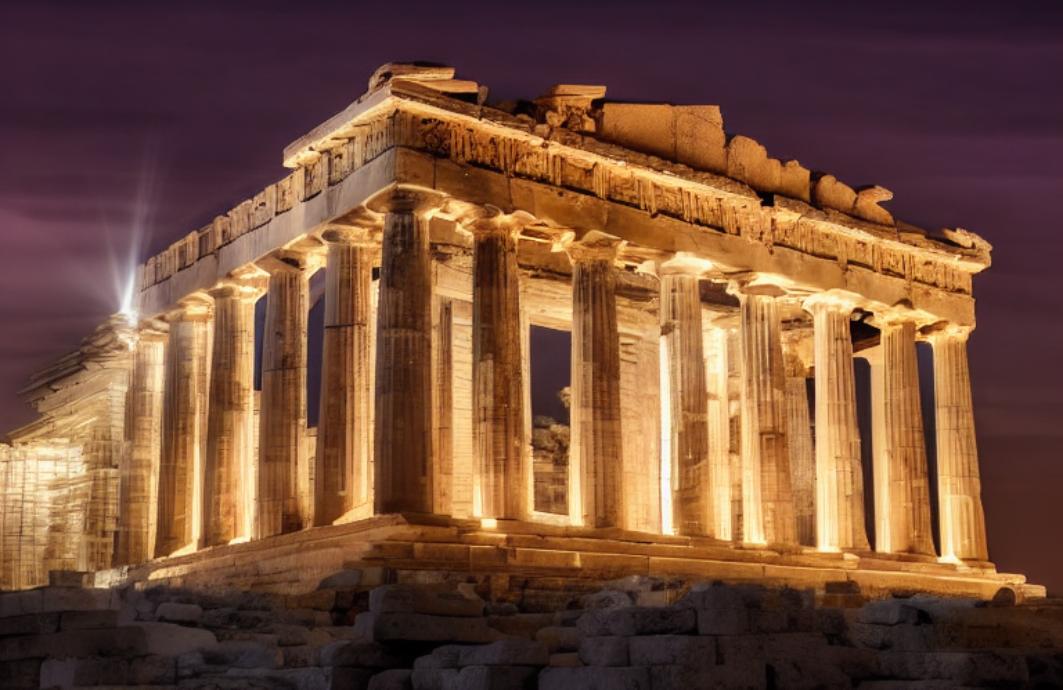} &
        \includegraphics[width=0.166\textwidth]{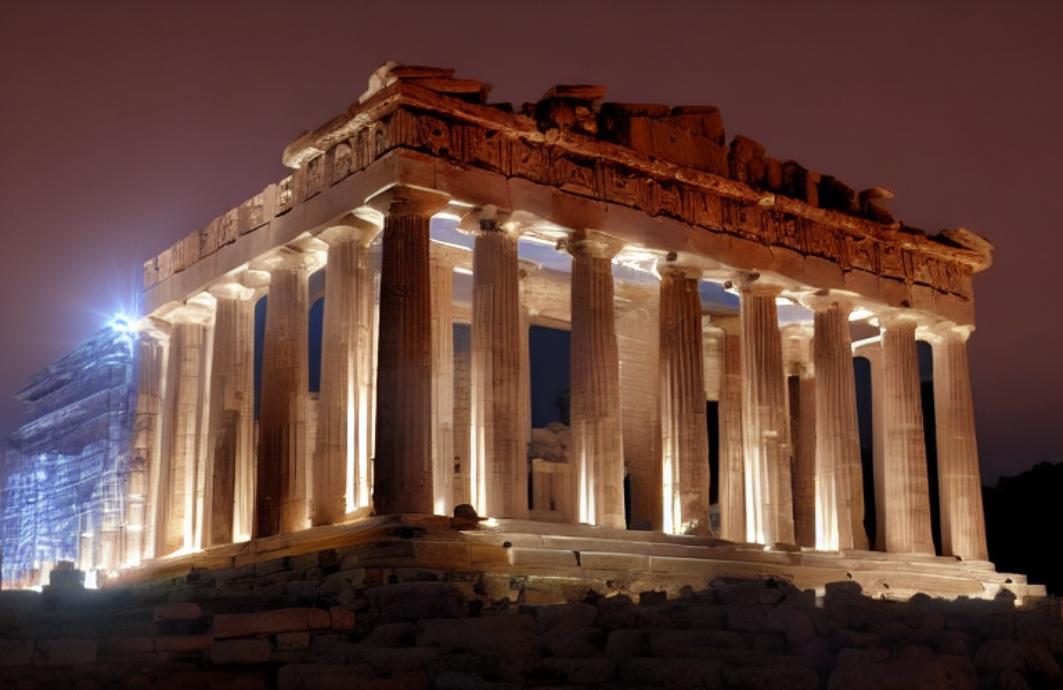} &
        \includegraphics[width=0.166\textwidth]{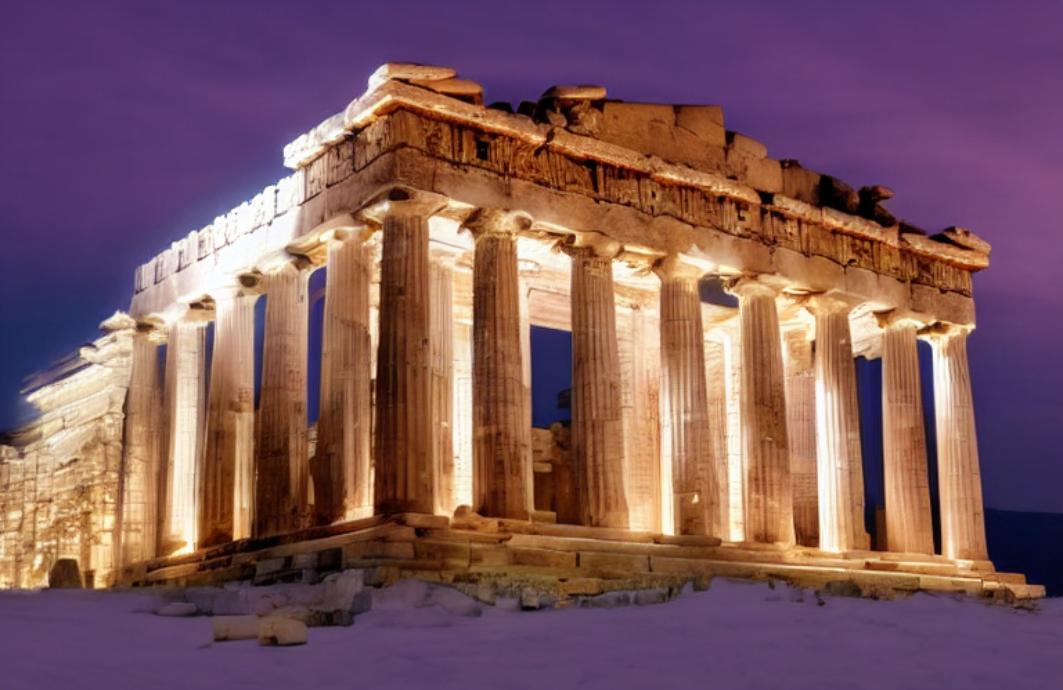} &
        \includegraphics[width=0.166\textwidth]{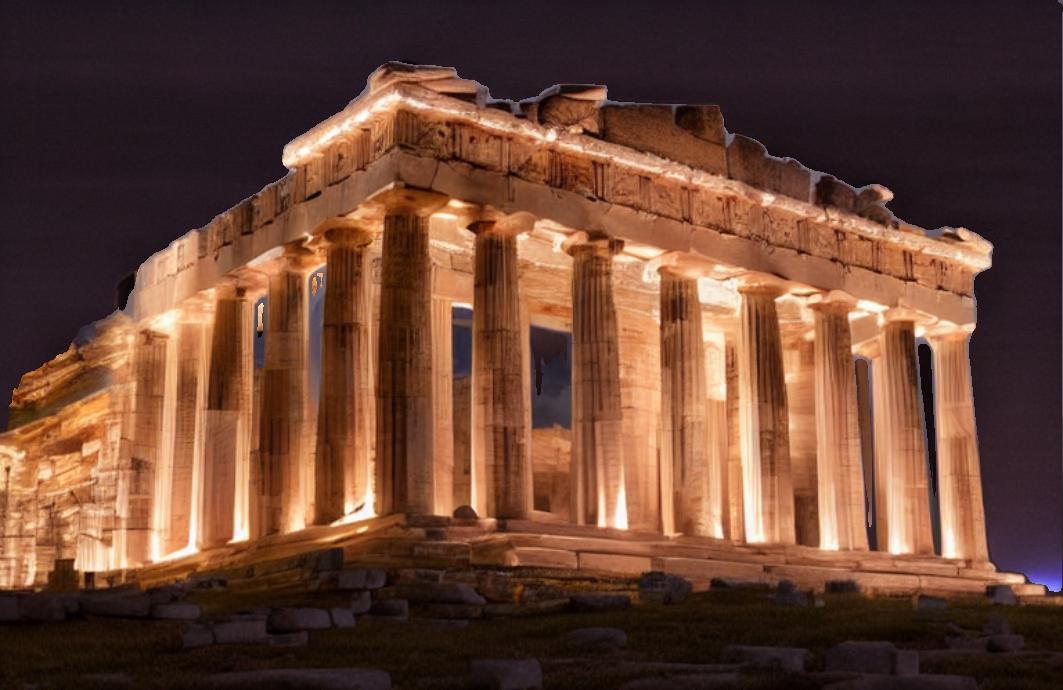} &
        \includegraphics[width=0.166\textwidth]{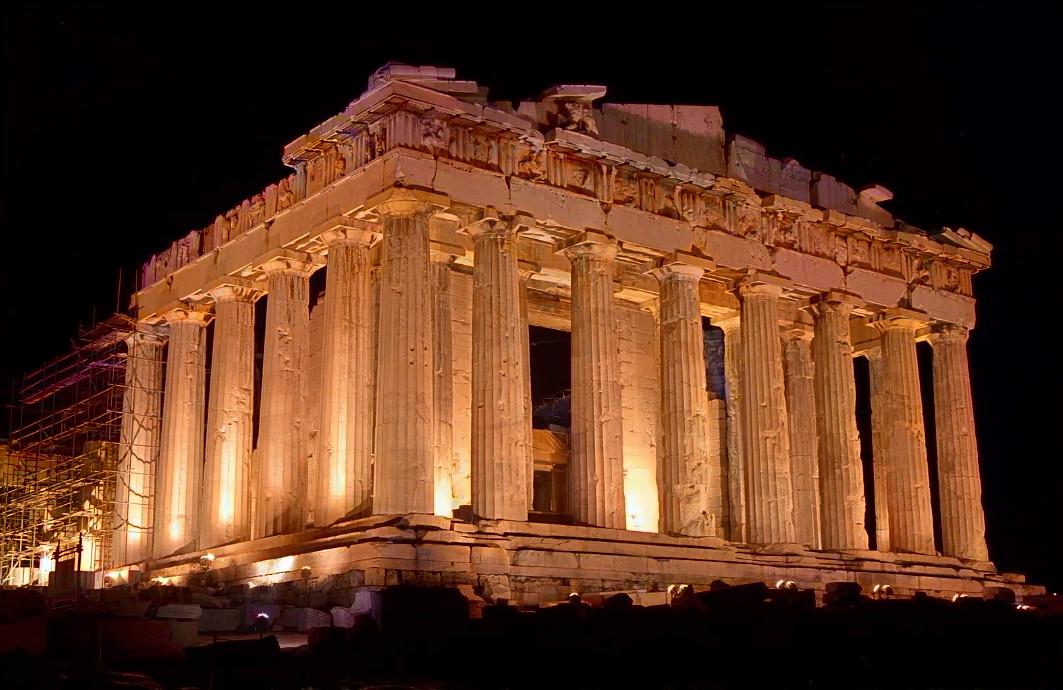} \\
        
        \includegraphics[width=0.166\textwidth]{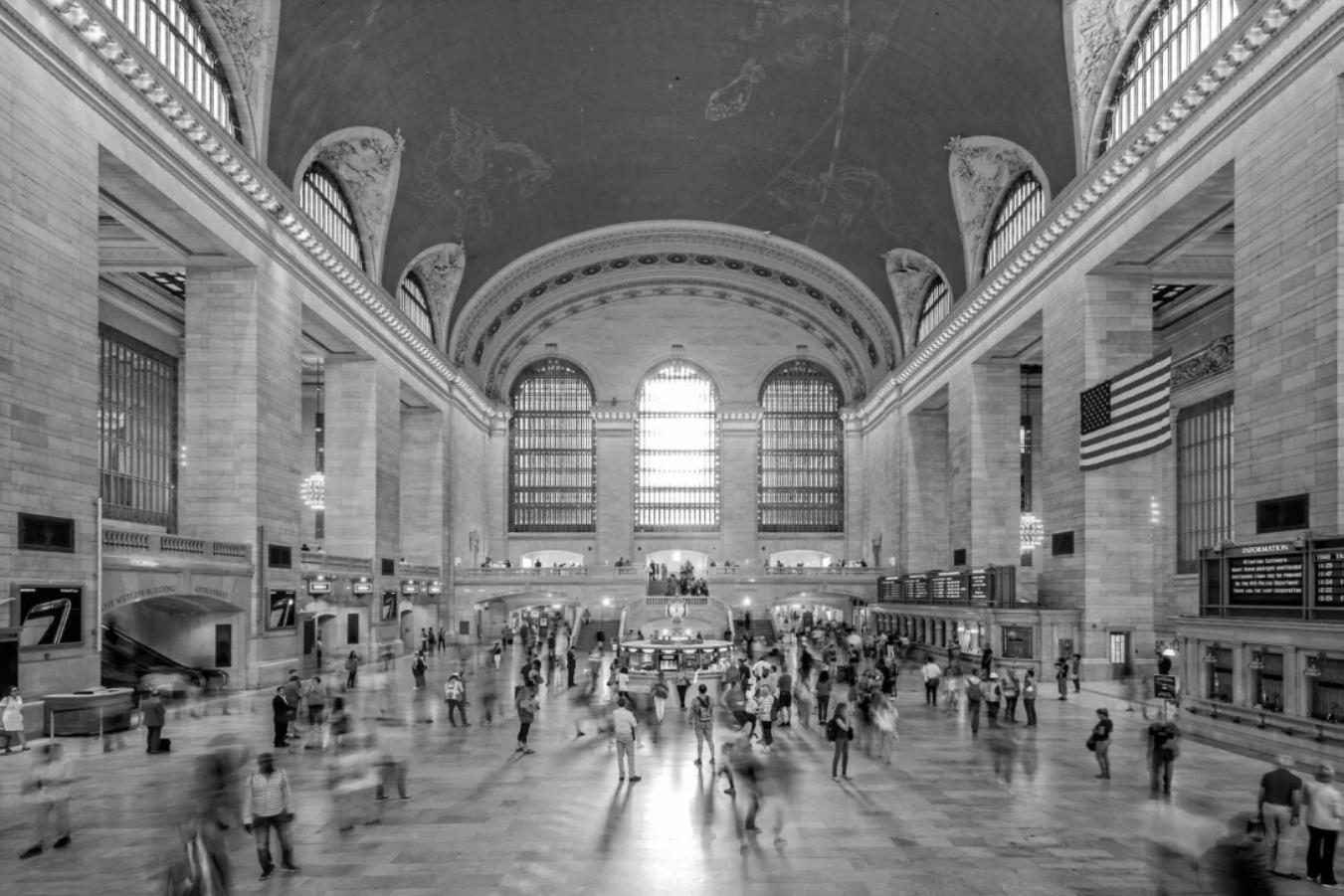} &
        \includegraphics[width=0.166\textwidth]{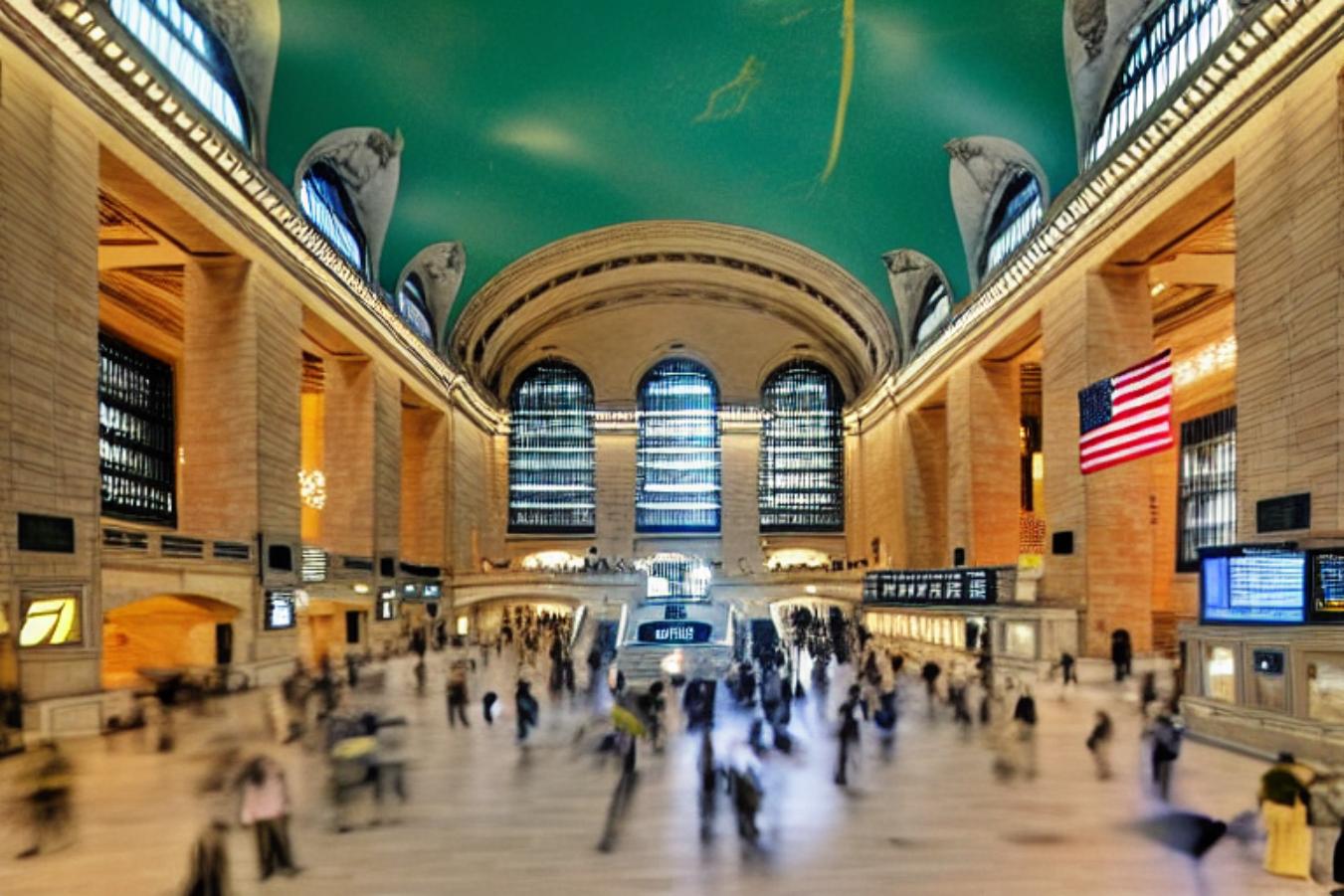} &
        \includegraphics[width=0.166\textwidth]{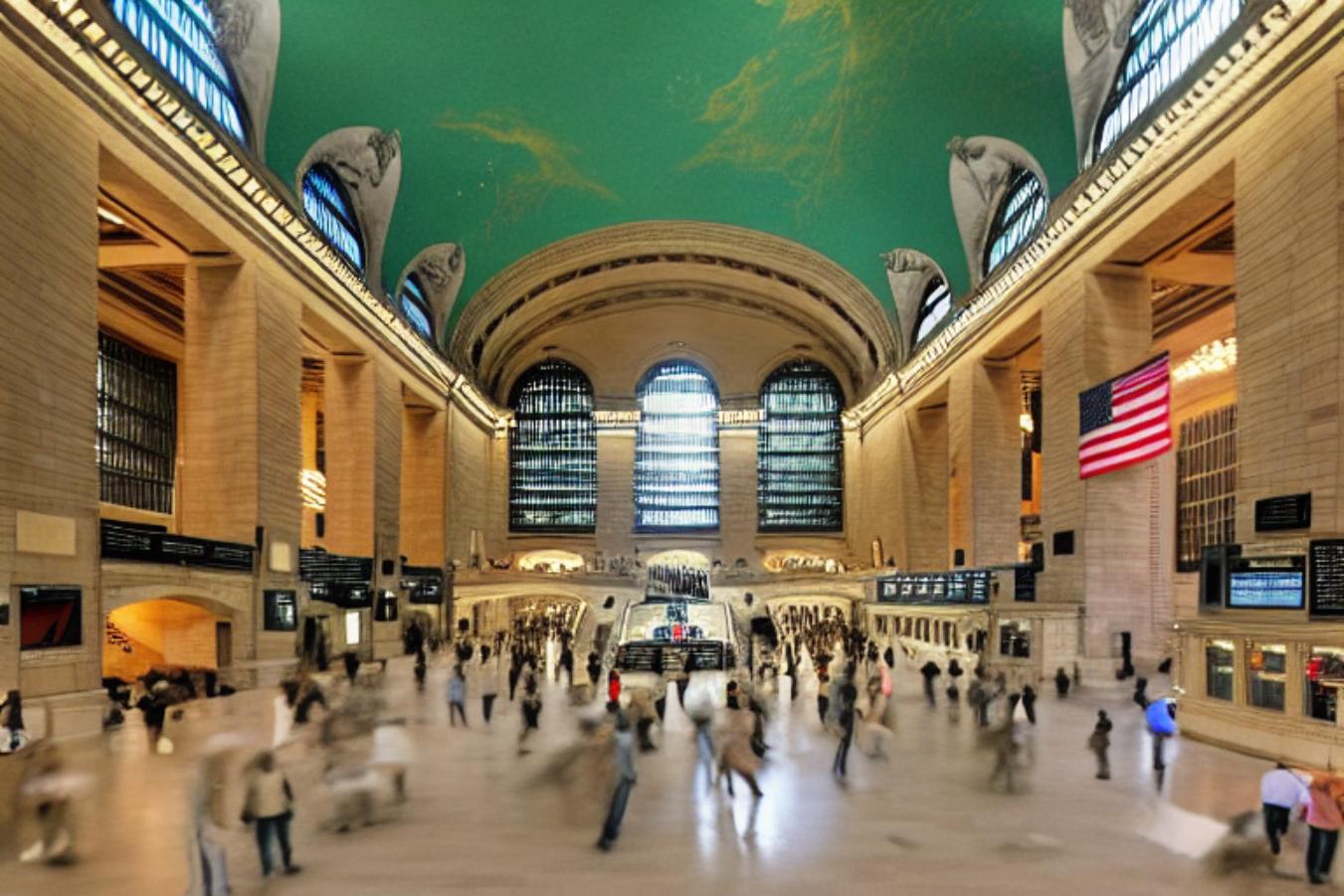} &
        \includegraphics[width=0.166\textwidth]{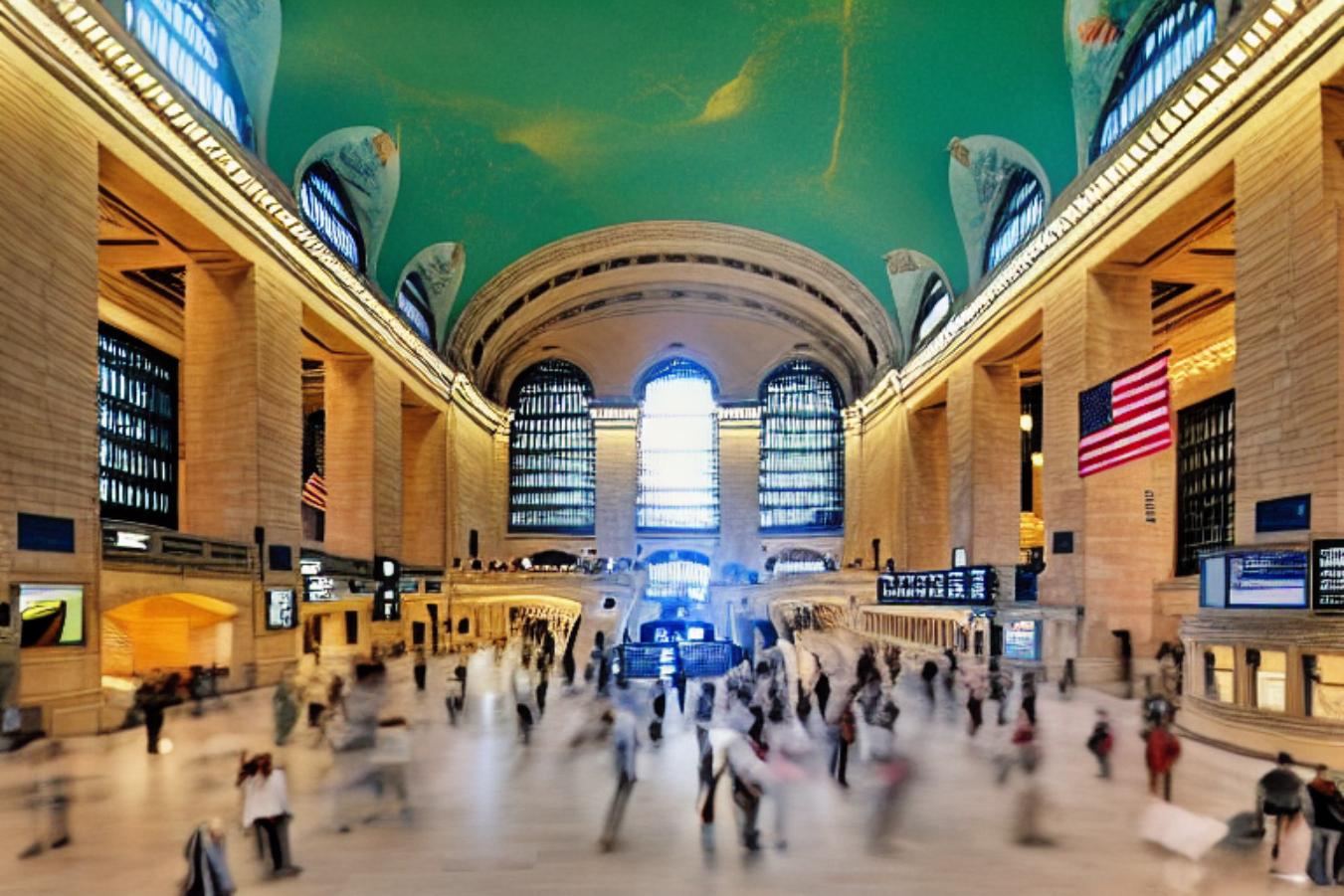} &
        \includegraphics[width=0.166\textwidth]{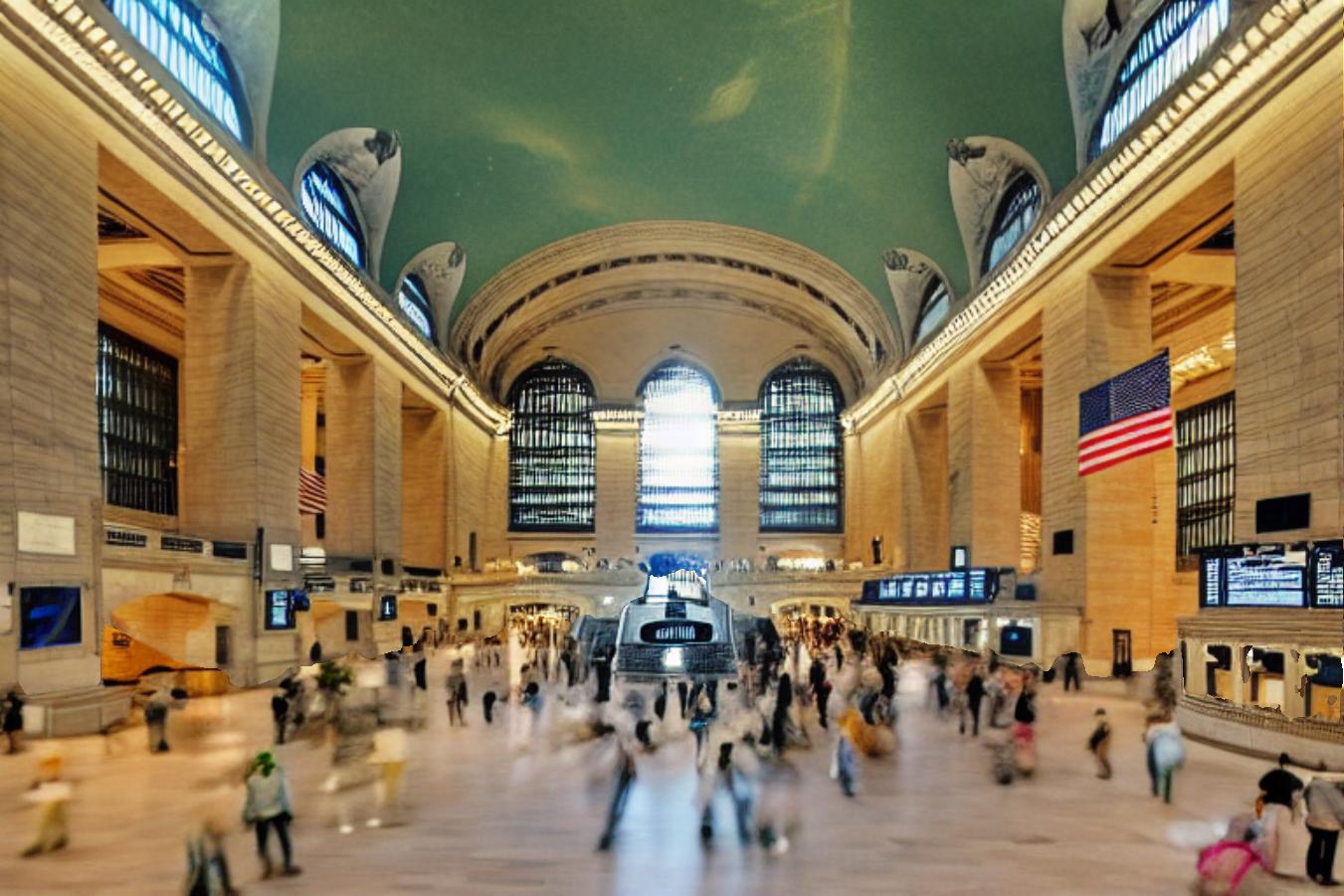} &
        \includegraphics[width=0.166\textwidth]{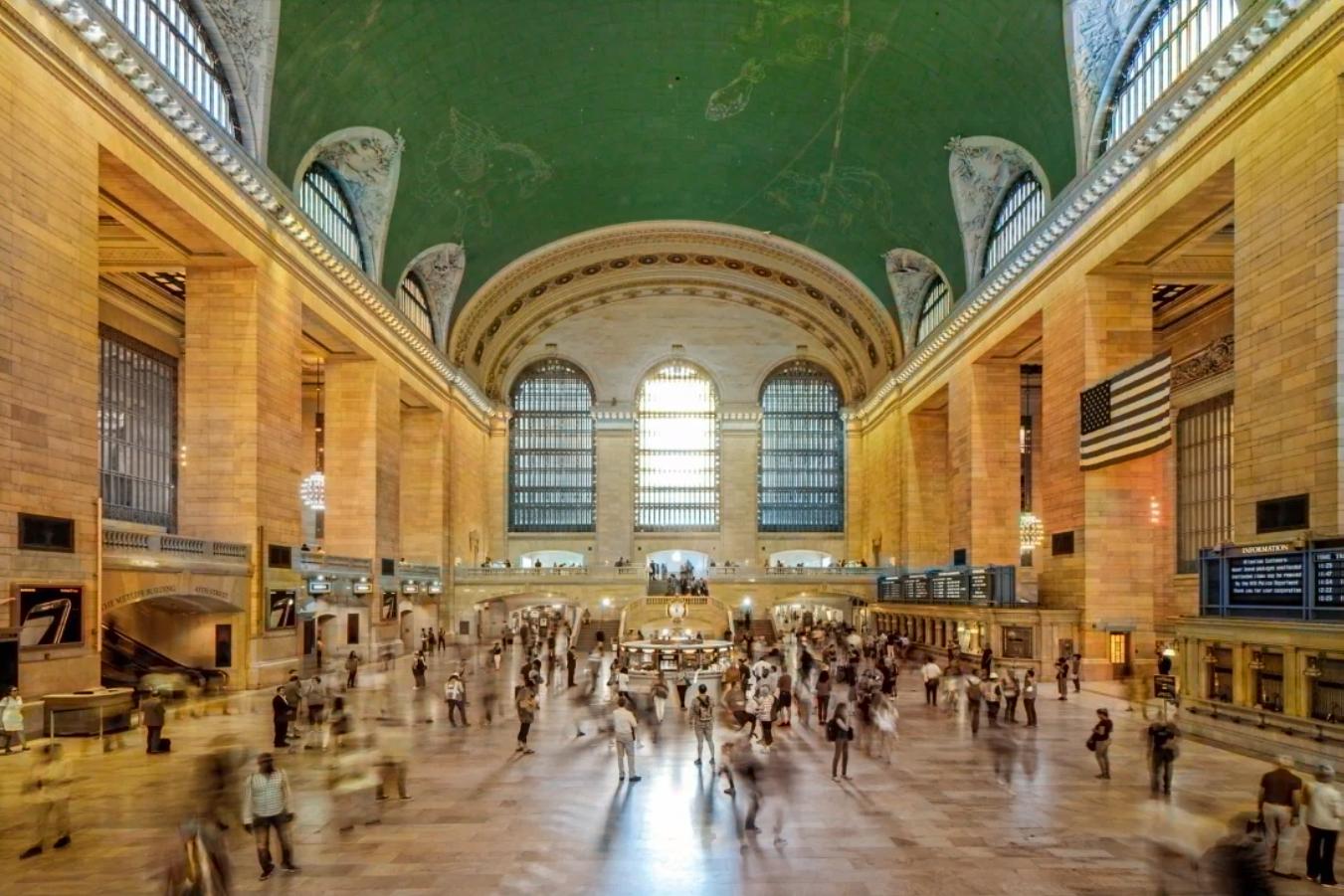} \\


        \includegraphics[width=0.166\textwidth]{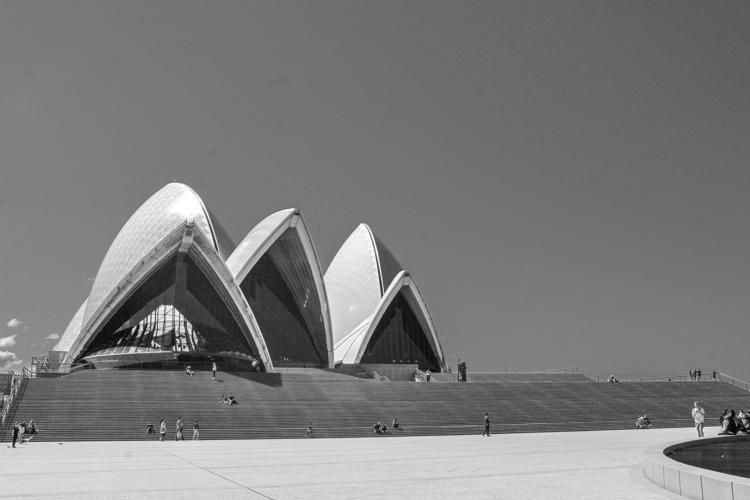} &
        \includegraphics[width=0.166\textwidth]{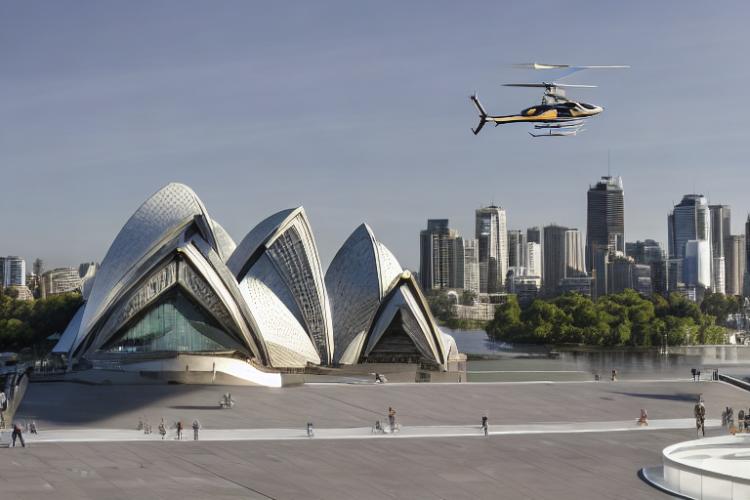} &
        \includegraphics[width=0.166\textwidth]{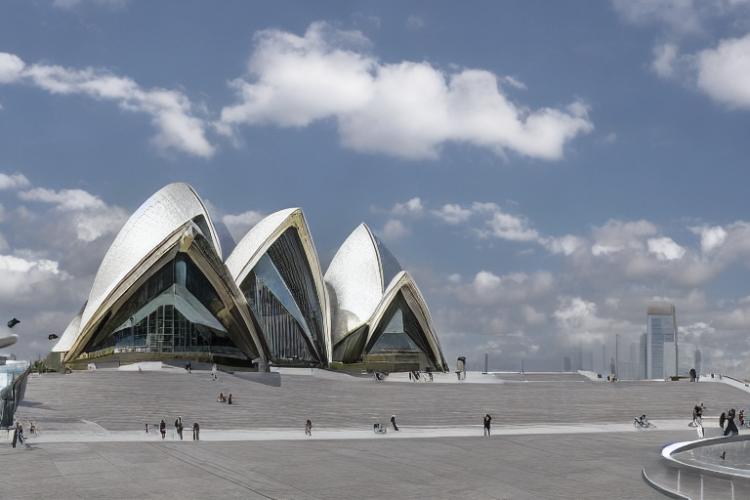} &
        \includegraphics[width=0.166\textwidth]{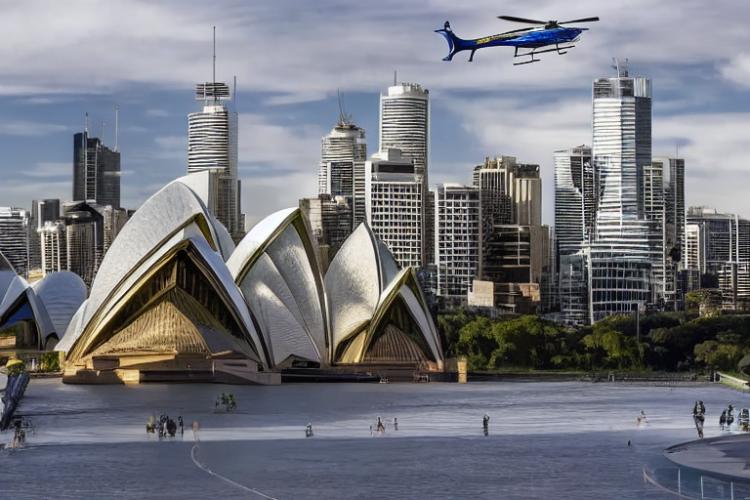} &
        \includegraphics[width=0.166\textwidth]{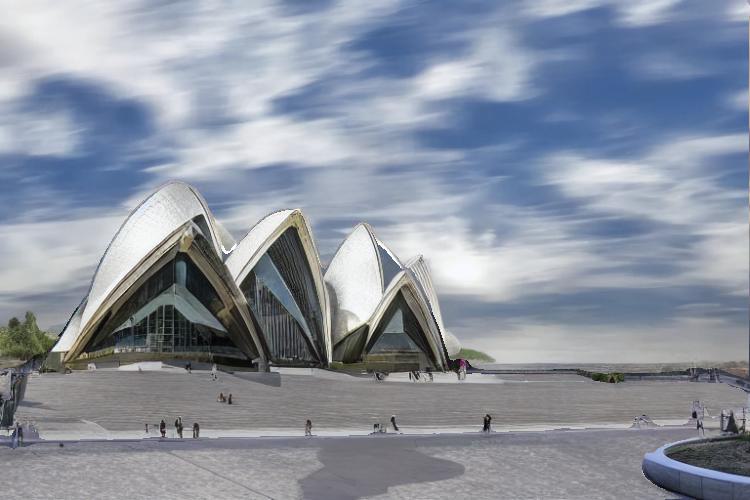} &
        \includegraphics[width=0.166\textwidth]{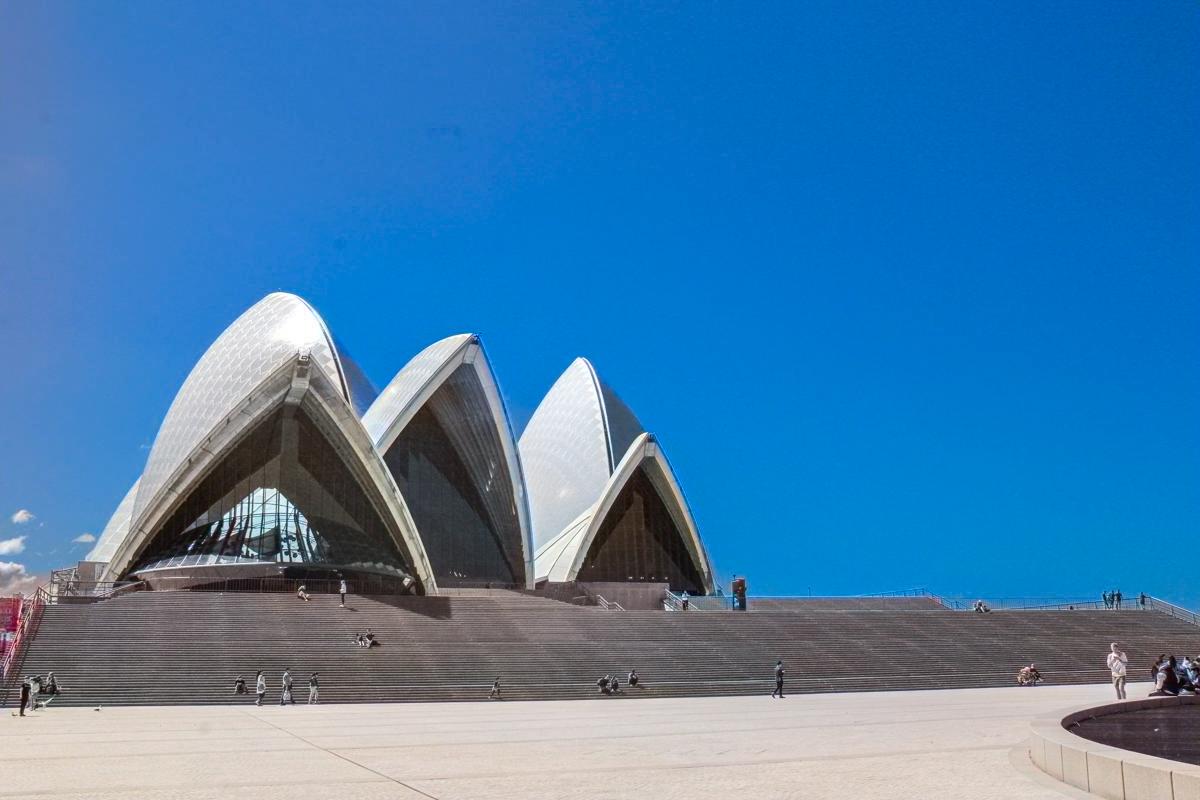}\\

        \includegraphics[width=0.166\textwidth]{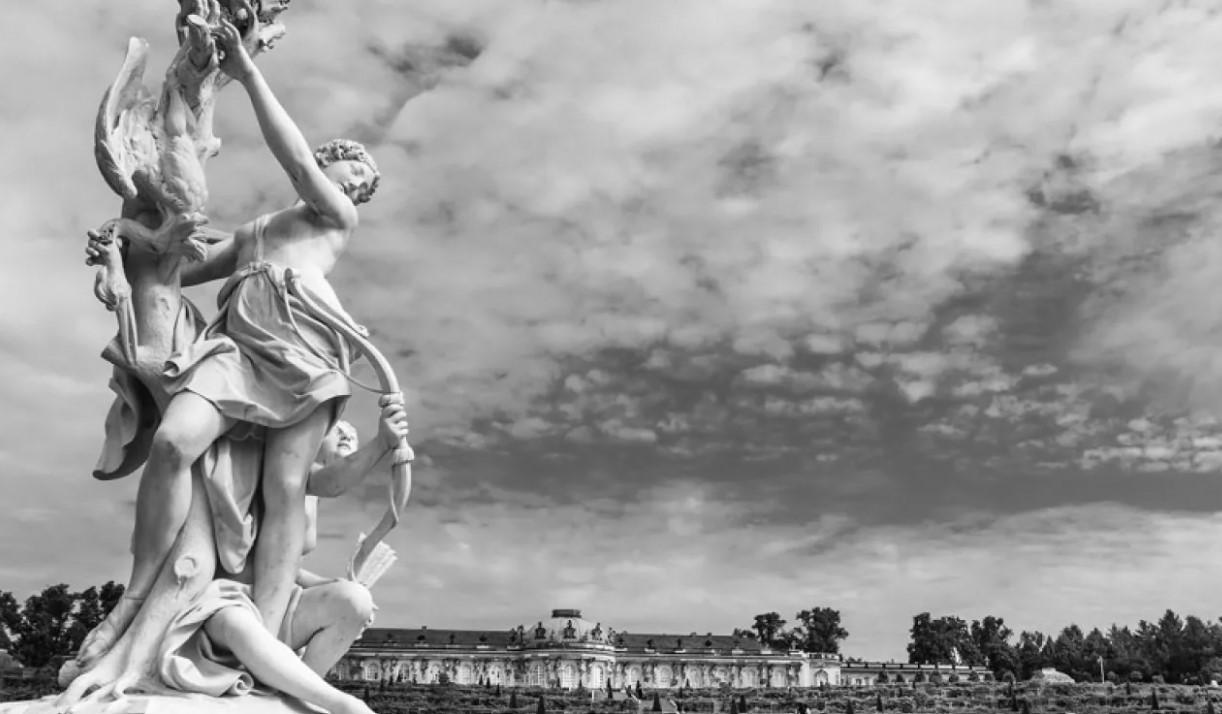} &
        \includegraphics[width=0.166\textwidth]{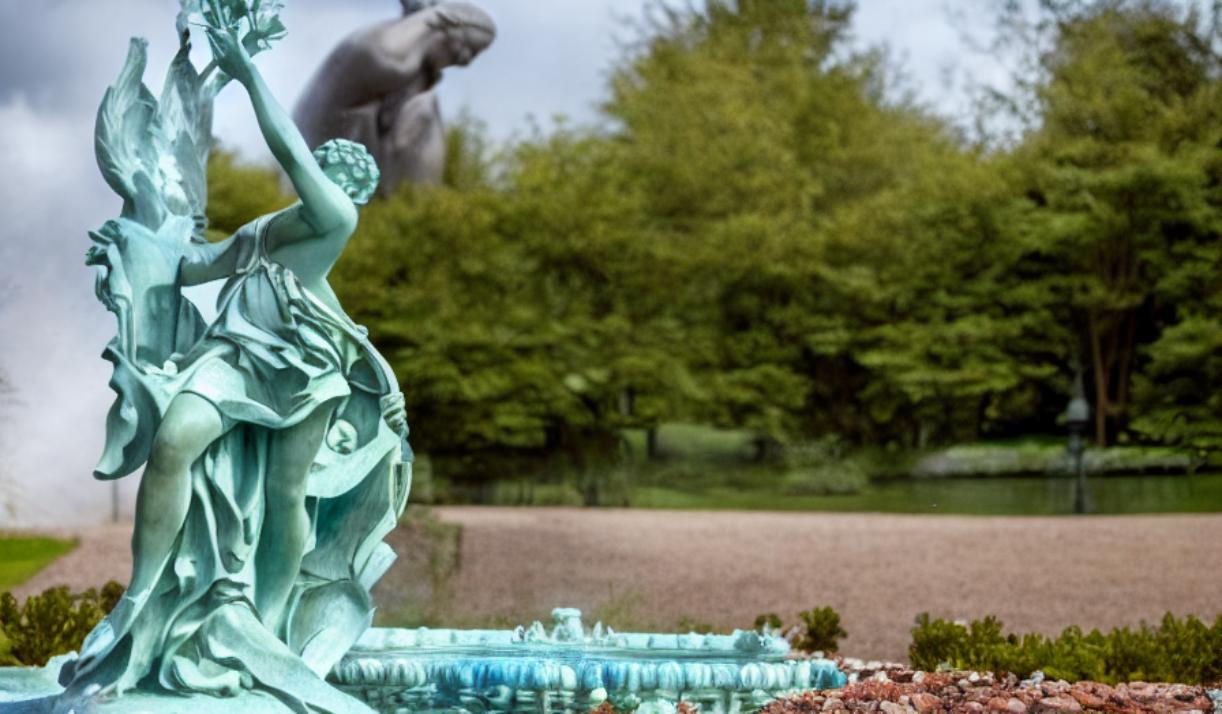} &
        \includegraphics[width=0.166\textwidth]{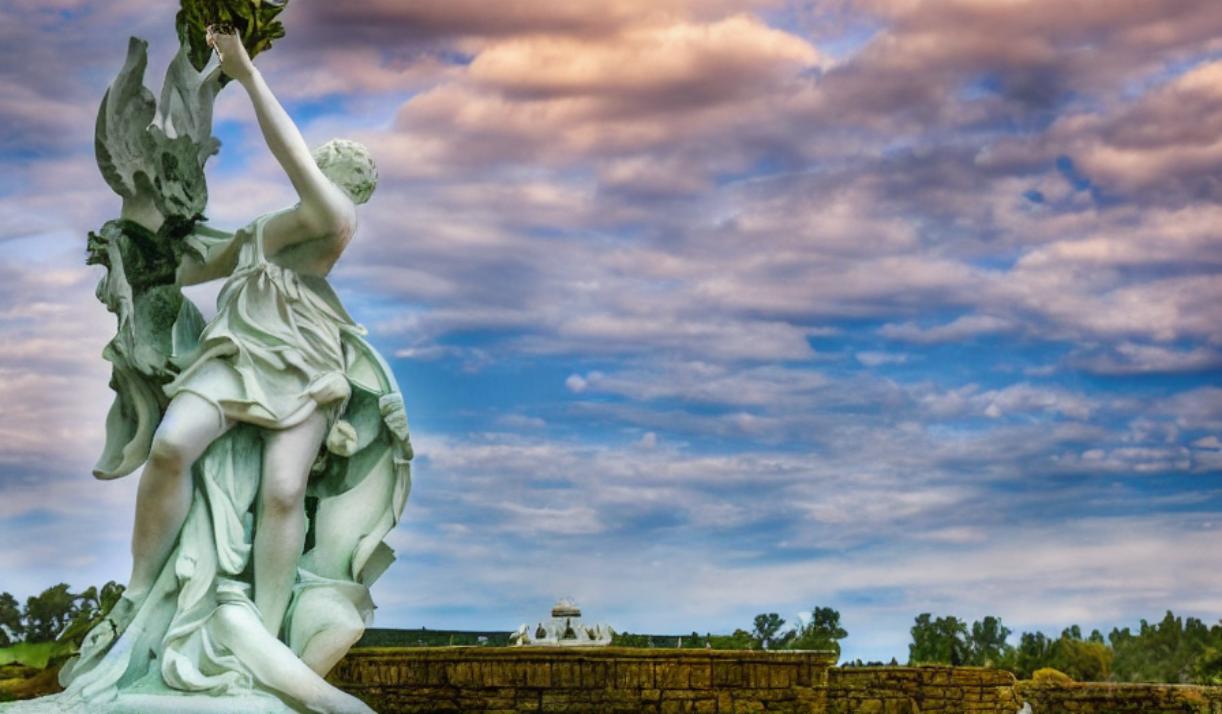} &
        \includegraphics[width=0.166\textwidth]{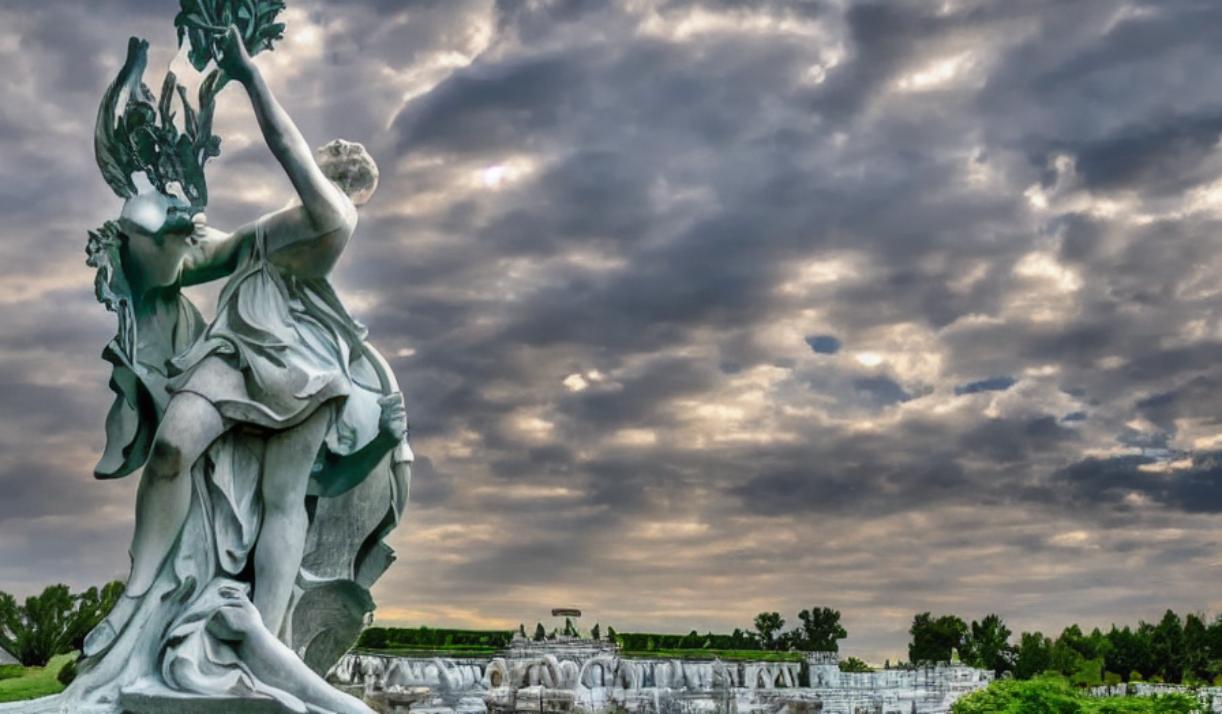} &
        \includegraphics[width=0.166\textwidth]{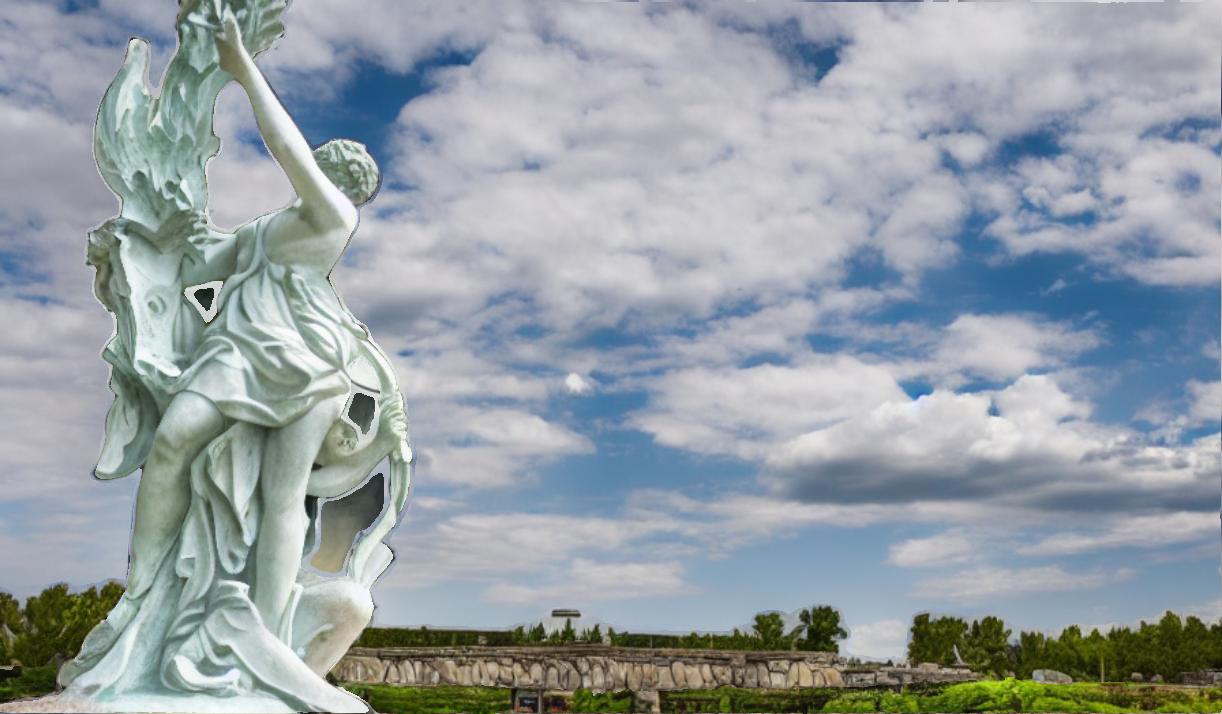} &
        \includegraphics[width=0.166\textwidth]{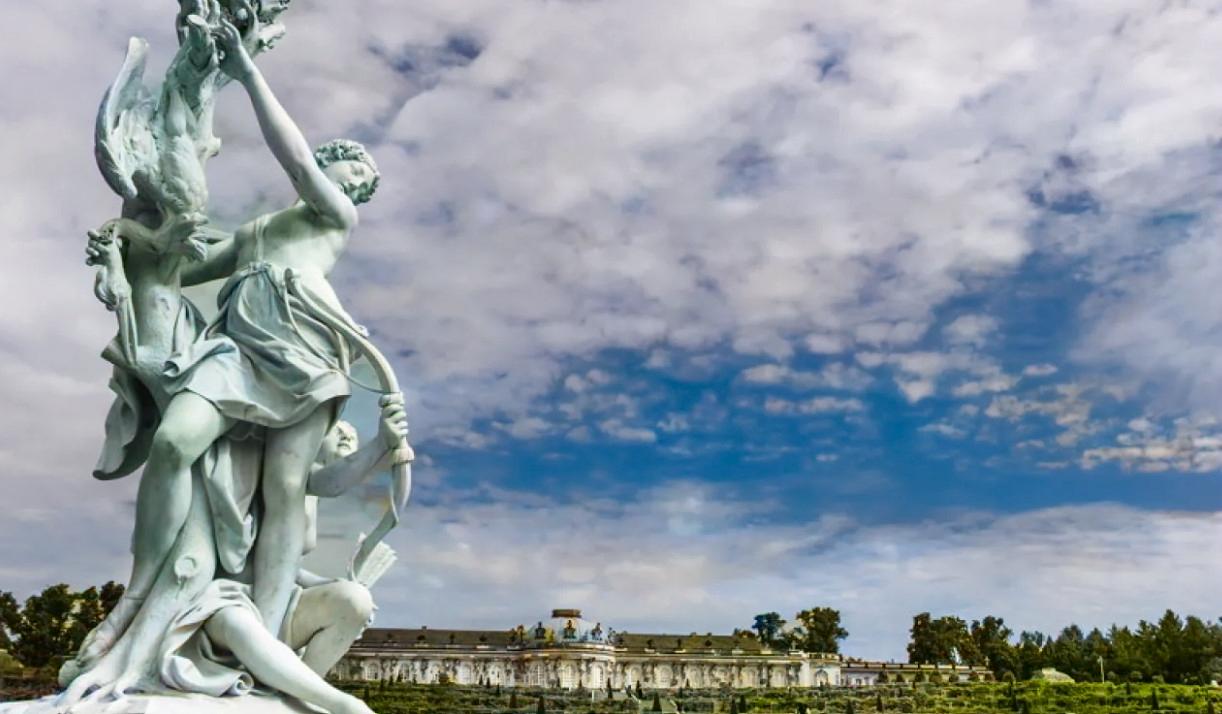}\\
        
        \includegraphics[width=0.166\textwidth]{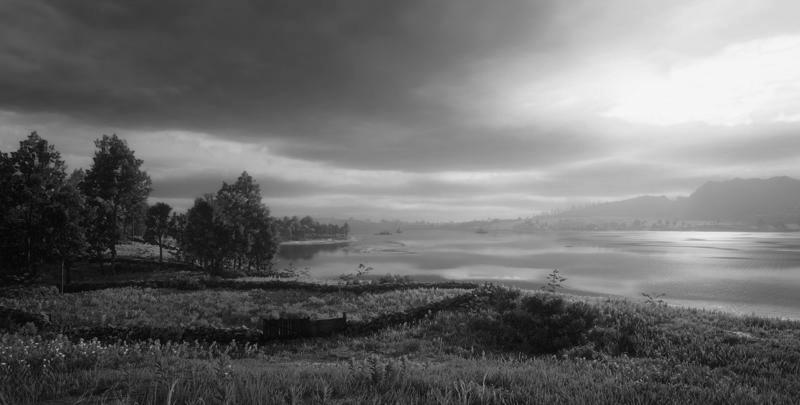} &
        \includegraphics[width=0.166\textwidth]{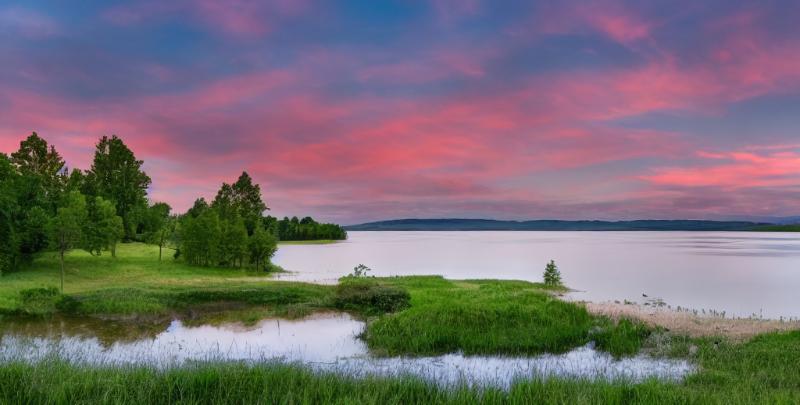} &
        \includegraphics[width=0.166\textwidth]{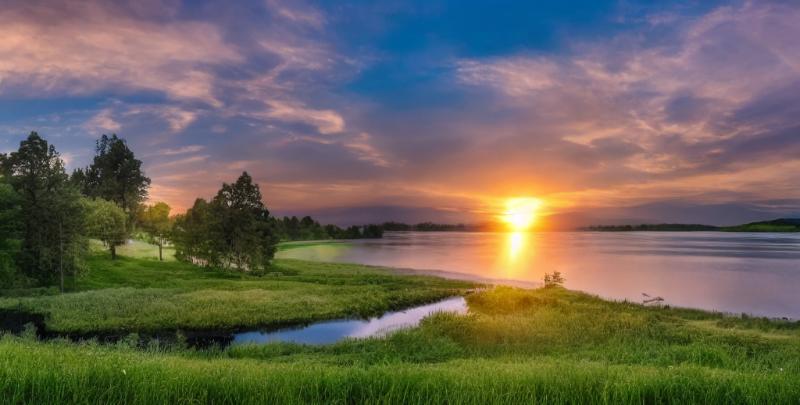} &
        \includegraphics[width=0.166\textwidth]{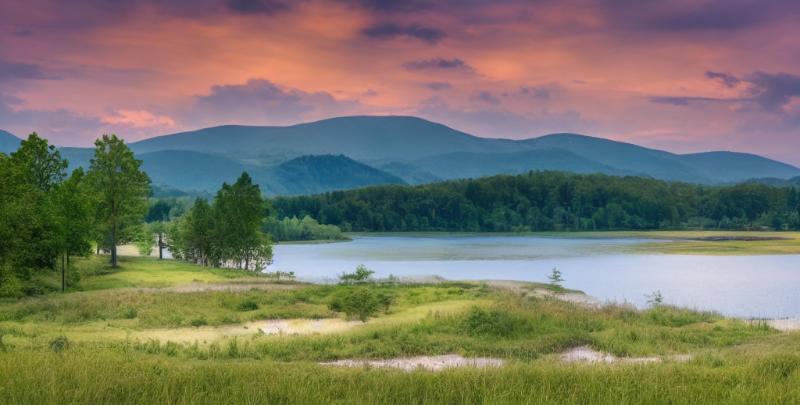} &
        \includegraphics[width=0.166\textwidth]{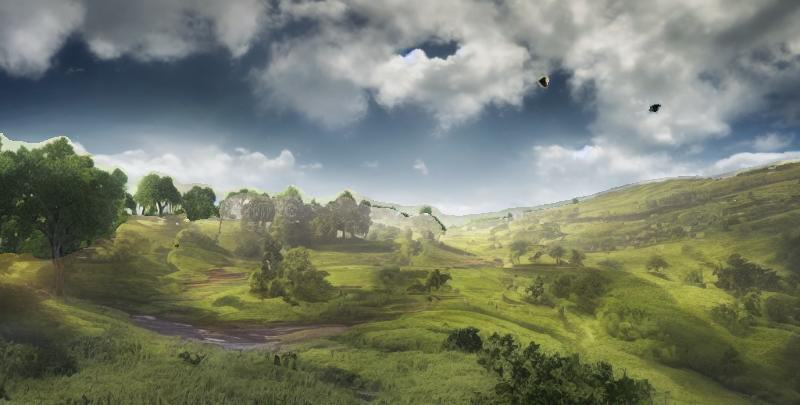} &
        \includegraphics[width=0.166\textwidth]{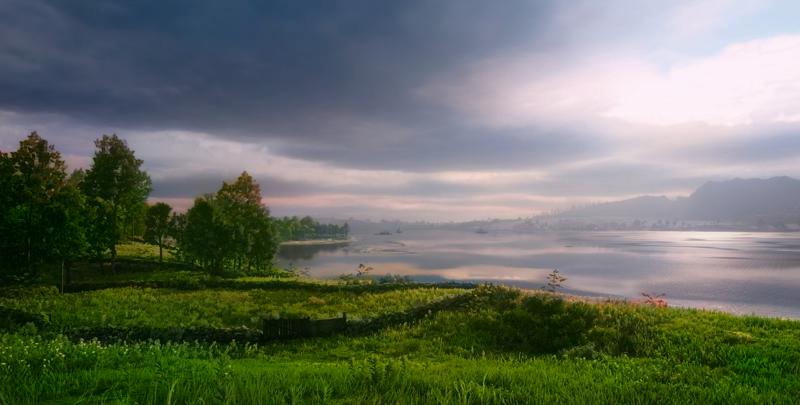}\\

        \includegraphics[width=0.166\textwidth]{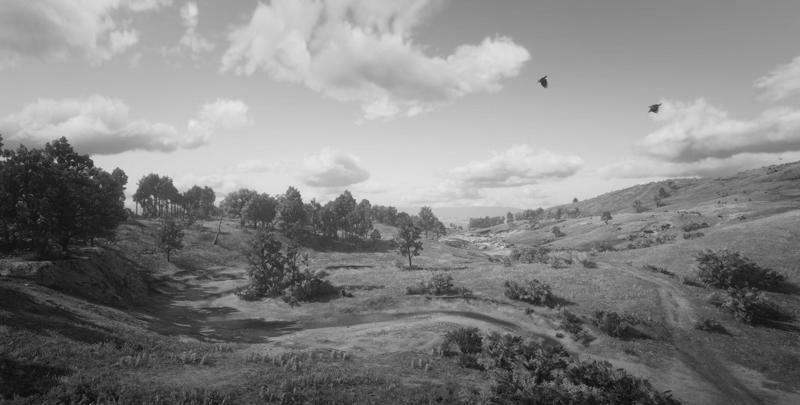} &
        \includegraphics[width=0.166\textwidth]{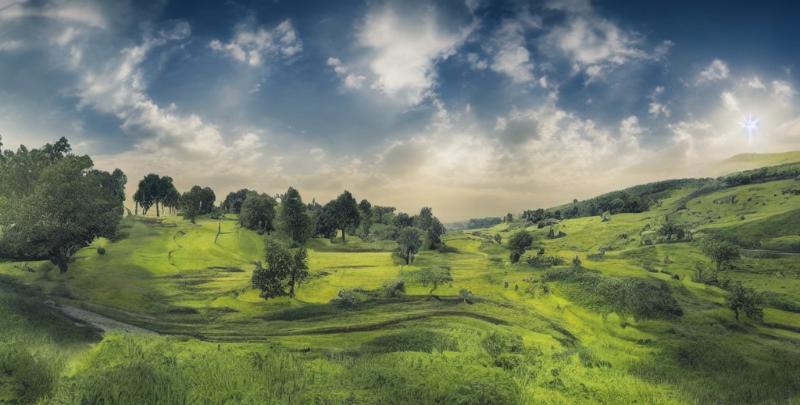} &
        \includegraphics[width=0.166\textwidth]{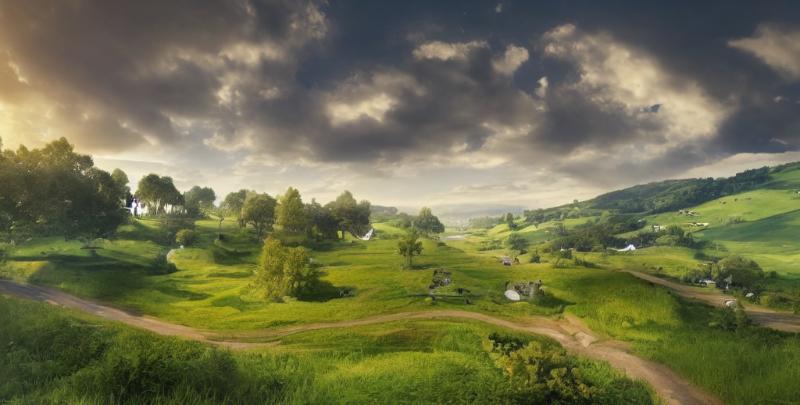} &
        \includegraphics[width=0.166\textwidth]{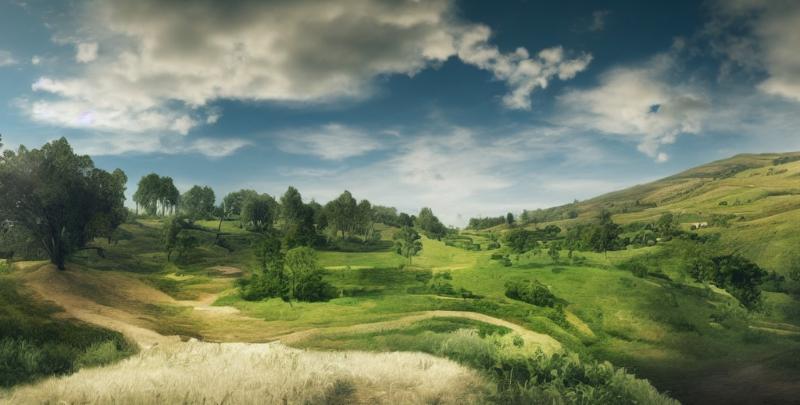} &
        \includegraphics[width=0.166\textwidth]{figs/supp_ref/huangyedabiaoke_002/SAM_00.jpg} &
        \includegraphics[width=0.166\textwidth]{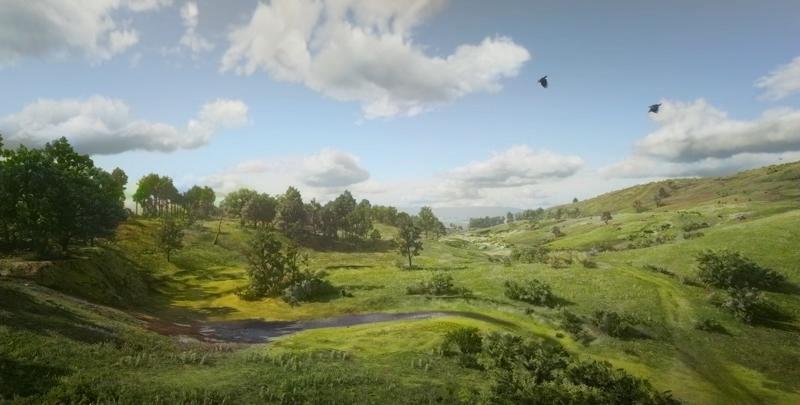}\\

        \includegraphics[width=0.166\textwidth]{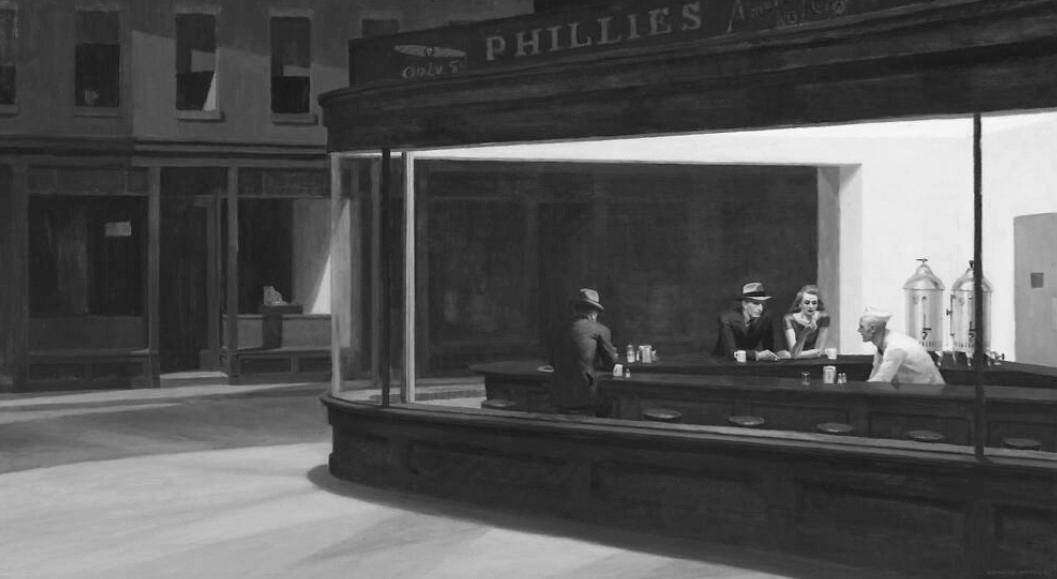} &
        \includegraphics[width=0.166\textwidth]{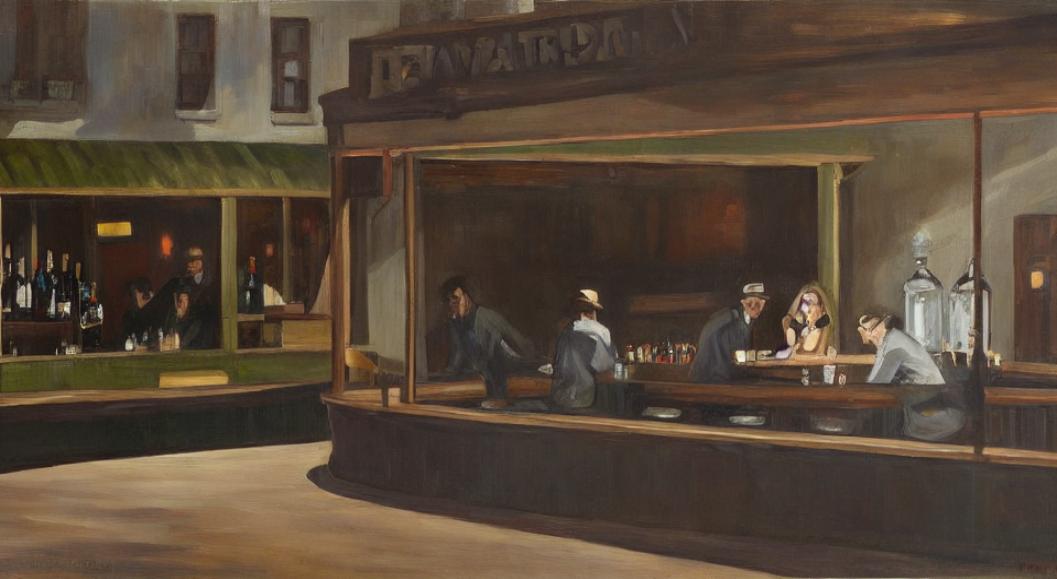} &
        \includegraphics[width=0.166\textwidth]{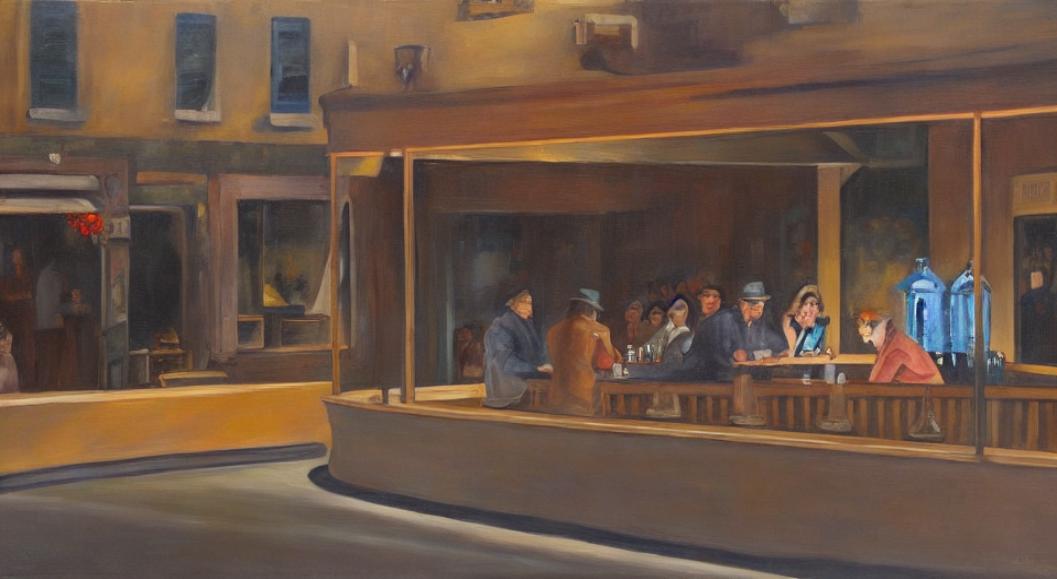} &
        \includegraphics[width=0.166\textwidth]{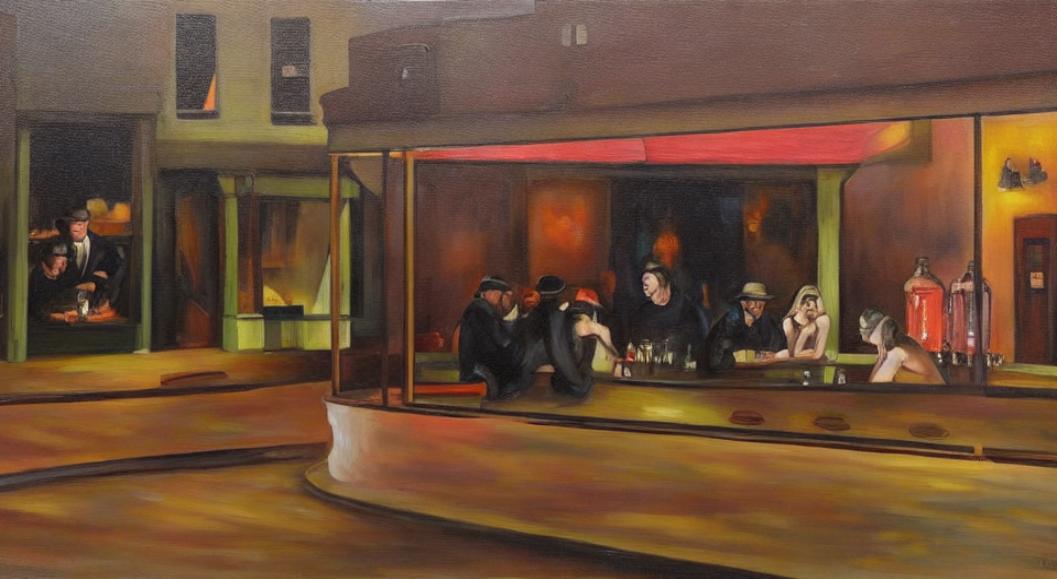} &
        \includegraphics[width=0.166\textwidth]{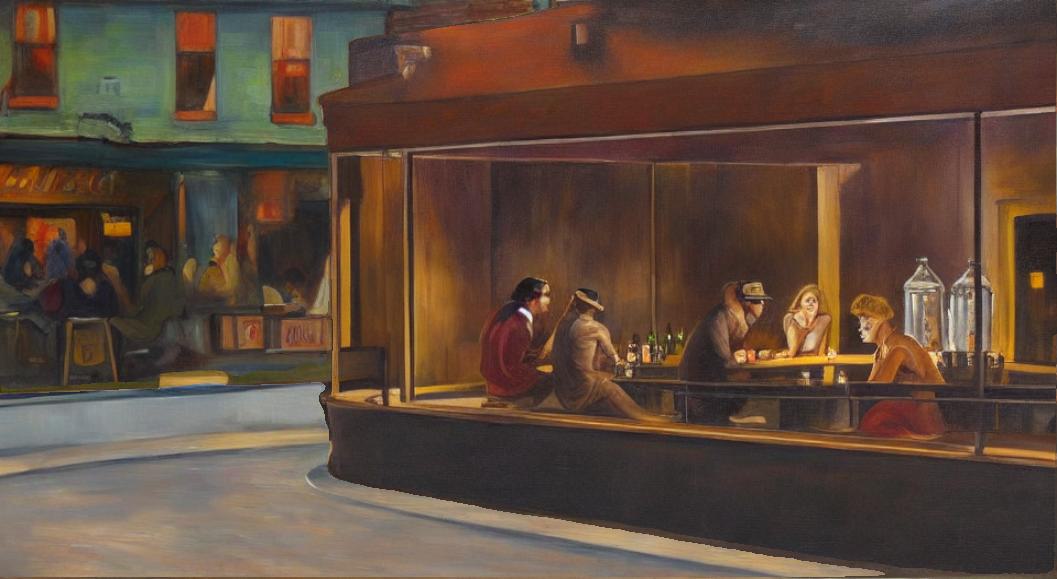} &
        \includegraphics[width=0.166\textwidth]{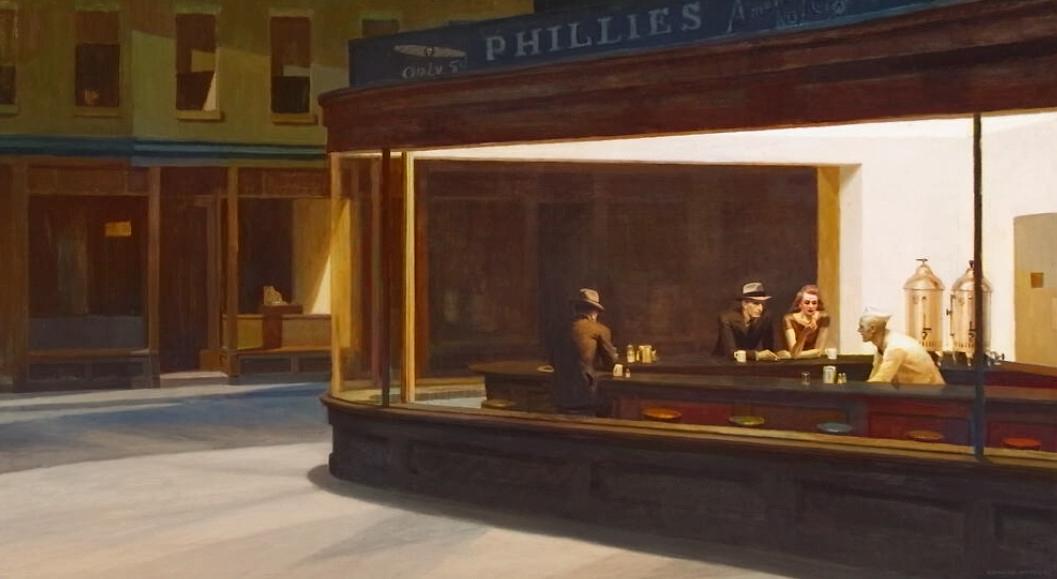}\\

        \includegraphics[width=0.166\textwidth]{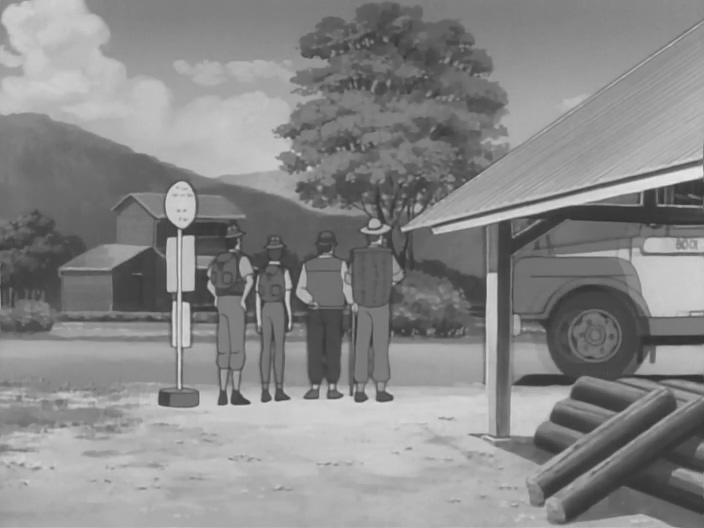} &
        \includegraphics[width=0.166\textwidth]{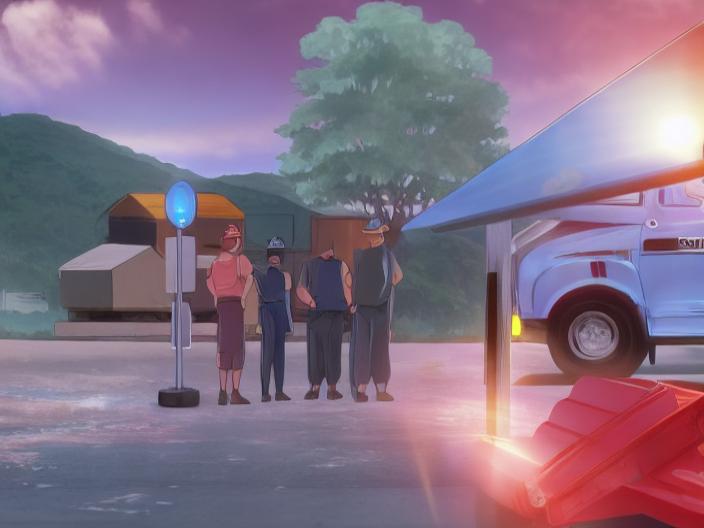} &
        \includegraphics[width=0.166\textwidth]{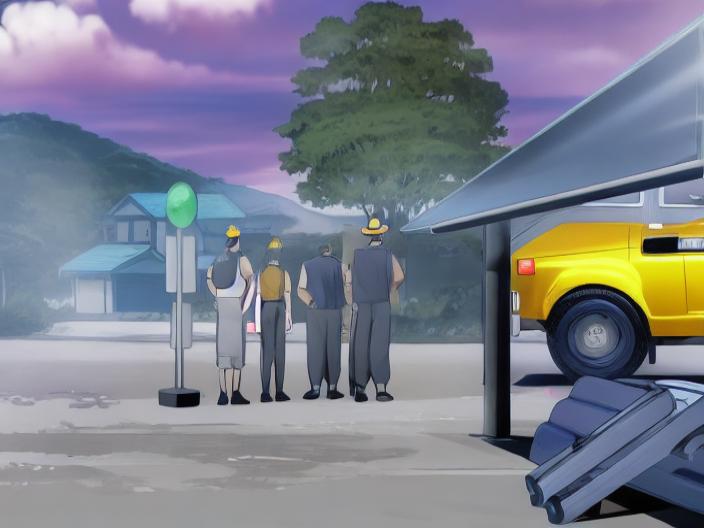} &
        \includegraphics[width=0.166\textwidth]{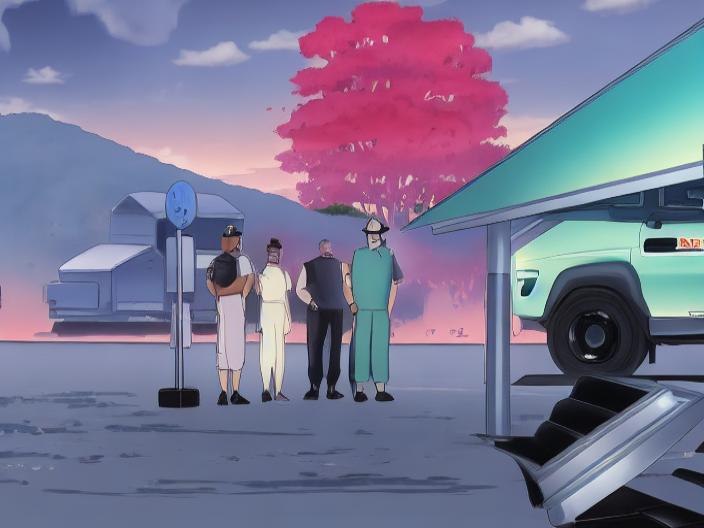} &
        \includegraphics[width=0.166\textwidth]{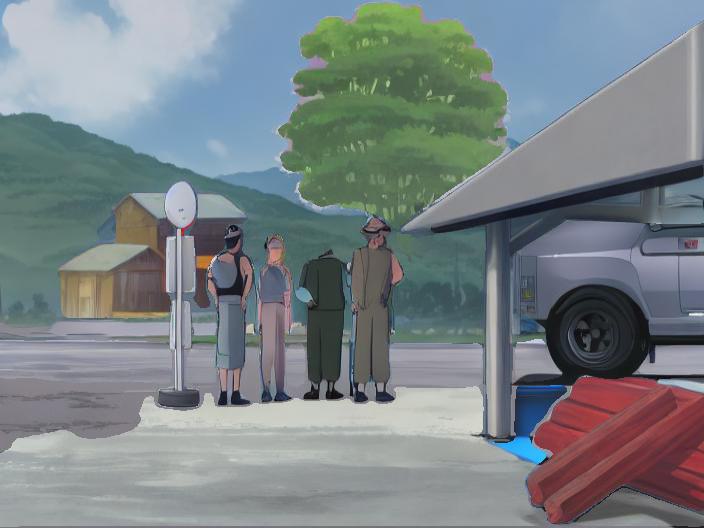} &
        \includegraphics[width=0.166\textwidth]{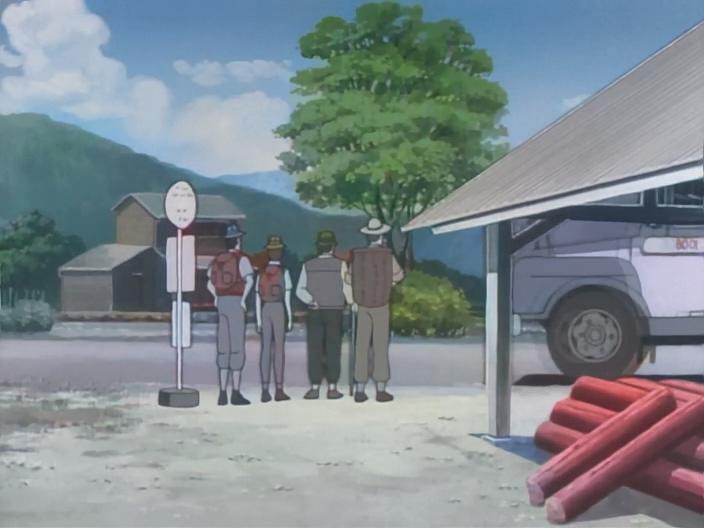} \\

        \includegraphics[width=0.166\textwidth]{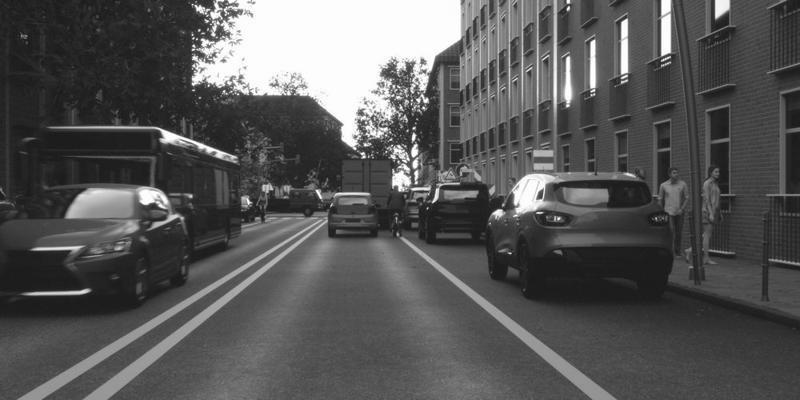} &
        \includegraphics[width=0.166\textwidth]{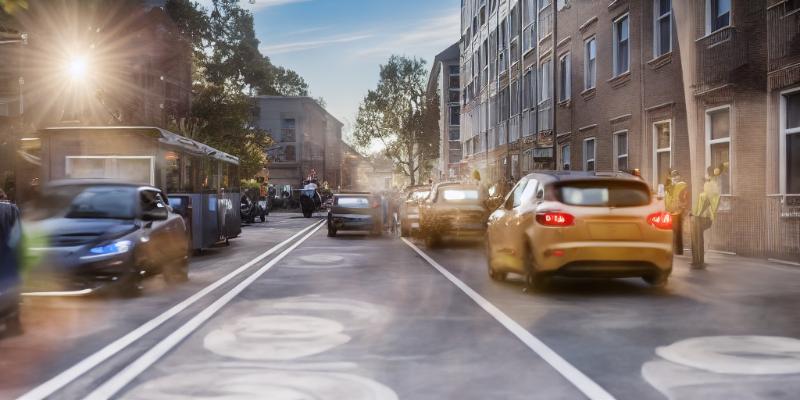} &
        \includegraphics[width=0.166\textwidth]{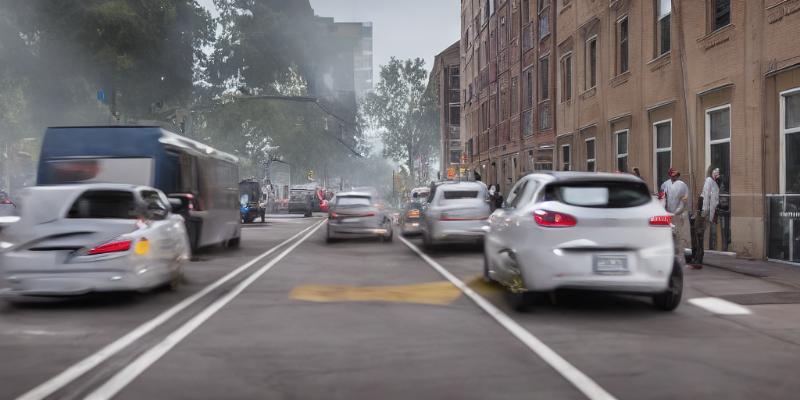} &
        \includegraphics[width=0.166\textwidth]{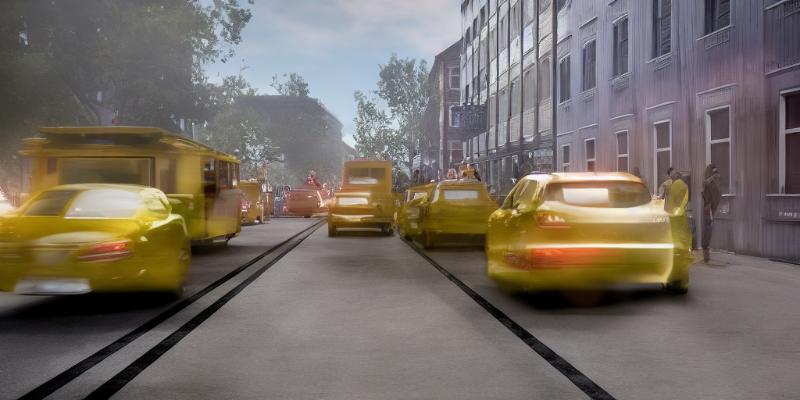} &
        \includegraphics[width=0.166\textwidth]{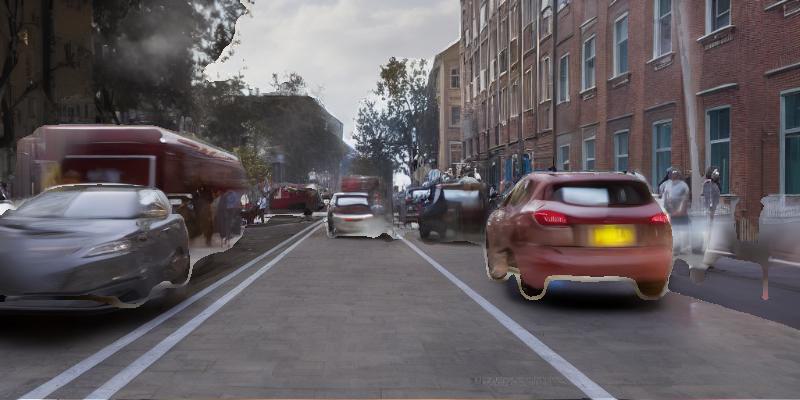} &
        \includegraphics[width=0.166\textwidth]{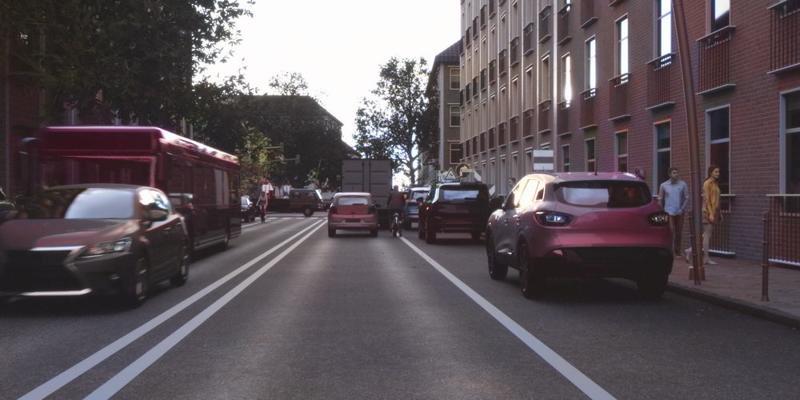}\\

        \includegraphics[width=0.166\textwidth]{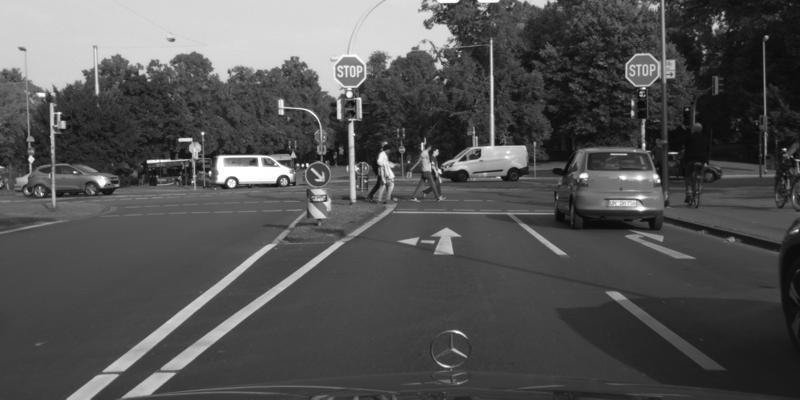} &
        \includegraphics[width=0.166\textwidth]{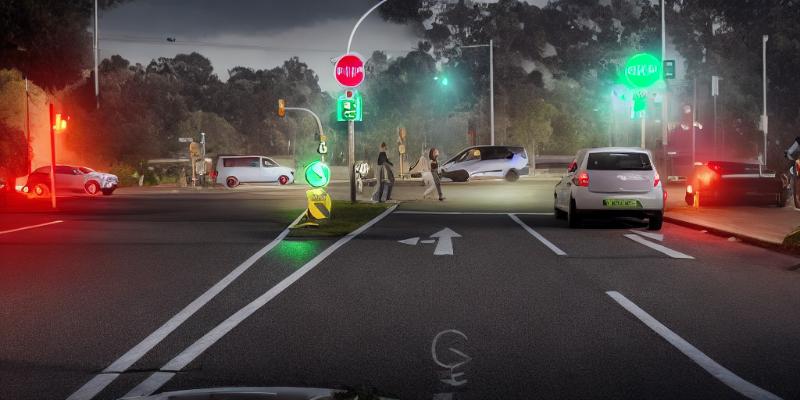} &
        \includegraphics[width=0.166\textwidth]{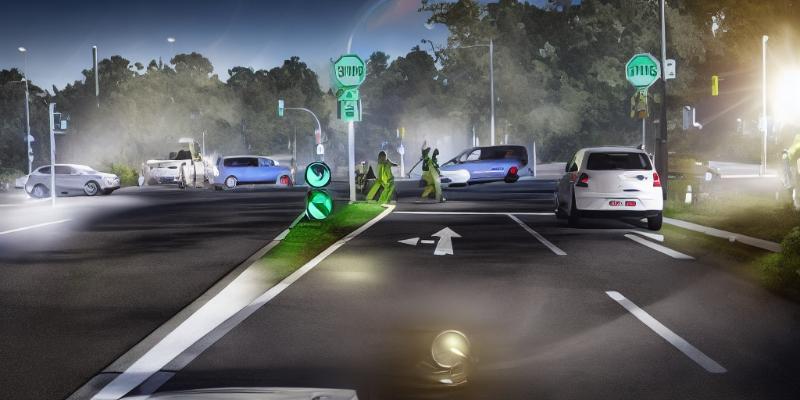} &
        \includegraphics[width=0.166\textwidth]{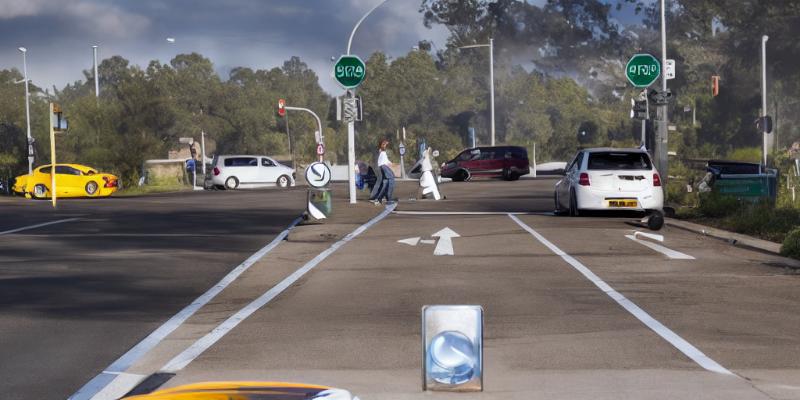} &
        \includegraphics[width=0.166\textwidth]{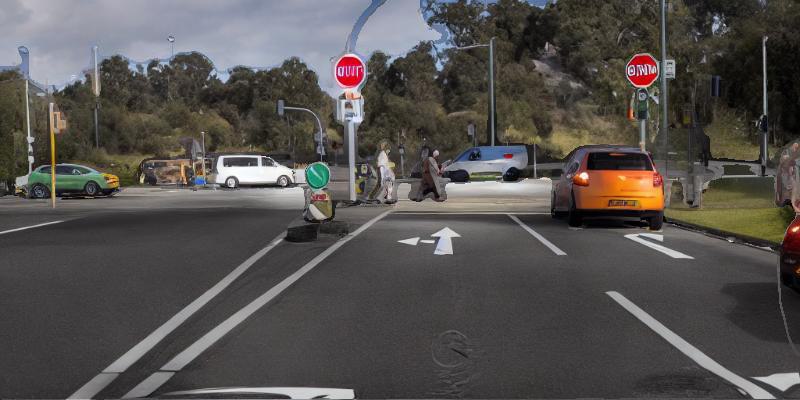} &
        \includegraphics[width=0.166\textwidth]{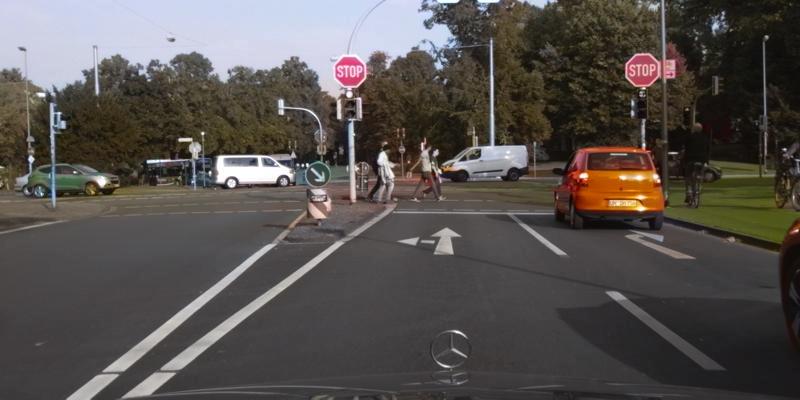}\\
        
        \multicolumn{1}{c}{\small\scalefont{0.7}Input} &
        \multicolumn{1}{c}{\small\scalefont{0.7}Candidate $1$}  &
        \multicolumn{1}{c}{\small\scalefont{0.7}Candidate $2$} &
        \multicolumn{1}{c}{\small\scalefont{0.7}Candidate $3$} &
        \multicolumn{1}{c}{\small\scalefont{0.7}Reference} &
        \multicolumn{1}{c}{\small\scalefont{0.7}Result}

    \end{tabular}
    \caption{\textbf{Visualization of reference synthsis.} For each grayscale input, we present randomly selected 3 reference candidates generated by our imagination module, the best reference synthesized by our reference refinement module and the colorization result. Zoom in for a better view.}
    \vspace{-0.1in}
    \label{fig_supp_ref}
\end{figure*}

We show more visual results of our synthesized reference in Figure \ref{fig_supp_ref}. For each grayscale input, we present randomly selected three reference candidates ($N$ in total), generated from our imagination module, a reference synthesized by our reference refinement module, and the colorization result. Empirically, we set $N=32$ in our experiments. Our imagination module can provide a diverse and abundant color space consisting of multiple semantically similar and structurally aligned reference candidates. However, there may exist some irrational color in each reference candidate, which may misguide our colorization module and yield incorrect colorization and unpleasant artifacts. Thus, we subsequently leverage our reference refinement module to aggregate information among the multiple reference candidates to synthesize the best reference with the help of the grayscale input. By selecting each semantic segment with the most similar DINO feature compared with the same segment in the grayscale input, a semantically similar, structurally aligned, and instance-aware reference is composed for the final colorization.

\section{More qualitative comparisons}
In Figure \ref{fig_supp_compare}, We show more quantitative comparisons results with 5 baselines, BigColor~\cite{Kim2022BigColor}, GCP~\cite{wu2021towards}, UniColor (Automatic)~\cite{huang2022unicolor}, DISCO~\cite{XiaHWW22} and DDColor~\cite{kang2022ddcolor}. All test images are from the COCO-stuff validation set. GAN-based methods, GCP~\cite{wu2021towards} and BigColor~\cite{Kim2022BigColor} tend to show color inconsistency like the bus in row 1 and the giraffe in row 2. Unicolor~\cite{huang2022unicolor} occasionally struggles with grayish colorization like the people in row 5. DISCO~\cite{XiaHWW22} often yields unpleasant over-saturated colorization in row 5 and row 6. DDColor~\cite{kang2022ddcolor} sometimes produces unpleasant artifacts like the buses in row 3 and row 4, and inconsistent colorization within one instance like the sign in row 7. Instead, our method can generate significantly more natural and photo-realistic colorization results. Our method can not only produce semantically reasonable colorization but also maintain color consistency for all instances. Additionally, our method can generate diverse reasonable colors for one semantic instance, such as the buses in row 1, row 3, and row 4 as well as humans in row 2, row 5, and row 6.

\begin{figure*}[b]
    \centering
    \setlength{\tabcolsep}{0pt} 
    \renewcommand{\arraystretch}{0.5} 
    \begin{tabular}{*{7}{p{0.14\textwidth}}} 
        \includegraphics[width=0.14\textwidth]{figs/coco_comparison/000000002006/gray.jpg} &
        \includegraphics[width=0.14\textwidth]{figs/coco_comparison/000000002006/BigColor.jpg} &
        \includegraphics[width=0.14\textwidth]{figs/coco_comparison/000000002006/GCP.jpg} &
        \includegraphics[width=0.14\textwidth]{figs/coco_comparison/000000002006/unicolor.jpg} &
        \includegraphics[width=0.14\textwidth]{figs/coco_comparison/000000002006/DISCO.jpg} &
        \includegraphics[width=0.14\textwidth]{figs/coco_comparison/000000002006/DDColor.jpg} &
        \includegraphics[width=0.14\textwidth]{figs/coco_comparison/000000002006/Ours.jpg} \\

        \includegraphics[width=0.14\textwidth]{figs/coco_comparison/000000130599/gray.jpg} &
        \includegraphics[width=0.14\textwidth]{figs/coco_comparison/000000130599/BigColor.jpg} &
        \includegraphics[width=0.14\textwidth]{figs/coco_comparison/000000130599/GCP.jpg} &
        \includegraphics[width=0.14\textwidth]{figs/coco_comparison/000000130599/unicolor.jpg} &
        \includegraphics[width=0.14\textwidth]{figs/coco_comparison/000000130599/DISCO.jpg} &
        \includegraphics[width=0.14\textwidth]{figs/coco_comparison/000000130599/DDColor.jpg} &
        \includegraphics[width=0.14\textwidth]{figs/coco_comparison/000000130599/Ours.jpg} \\

        \includegraphics[width=0.14\textwidth]{figs/coco_comparison/000000303305/gray.jpg} &
        \includegraphics[width=0.14\textwidth]{figs/coco_comparison/000000303305/BigColor.jpg} &
        \includegraphics[width=0.14\textwidth]{figs/coco_comparison/000000303305/GCP.jpg} &
        \includegraphics[width=0.14\textwidth]{figs/coco_comparison/000000303305/unicolor.jpg} &
        \includegraphics[width=0.14\textwidth]{figs/coco_comparison/000000303305/DISCO.jpg} &
        \includegraphics[width=0.14\textwidth]{figs/coco_comparison/000000303305/DDColor.jpg} &
        \includegraphics[width=0.14\textwidth]{figs/coco_comparison/000000303305/Ours.jpg} \\

        \includegraphics[width=0.14\textwidth]{figs/coco_comparison/000000311909/gray.jpg} &
        \includegraphics[width=0.14\textwidth]{figs/coco_comparison/000000311909/BigColor.jpg} &
        \includegraphics[width=0.14\textwidth]{figs/coco_comparison/000000311909/GCP.jpg} &
        \includegraphics[width=0.14\textwidth]{figs/coco_comparison/000000311909/unicolor.jpg} &
        \includegraphics[width=0.14\textwidth]{figs/coco_comparison/000000311909/DISCO.jpg} &
        \includegraphics[width=0.14\textwidth]{figs/coco_comparison/000000311909/DDColor.jpg} &
        \includegraphics[width=0.14\textwidth]{figs/coco_comparison/000000311909/Ours.jpg} \\

        \includegraphics[width=0.14\textwidth]{figs/coco_comparison/000000373705/gray.jpg} &
        \includegraphics[width=0.14\textwidth]{figs/coco_comparison/000000373705/BigColor.jpg} &
        \includegraphics[width=0.14\textwidth]{figs/coco_comparison/000000373705/GCP.jpg} &
        \includegraphics[width=0.14\textwidth]{figs/coco_comparison/000000373705/unicolor.jpg} &
        \includegraphics[width=0.14\textwidth]{figs/coco_comparison/000000373705/DISCO.jpg} &
        \includegraphics[width=0.14\textwidth]{figs/coco_comparison/000000373705/DDColor.jpg} &
        \includegraphics[width=0.14\textwidth]{figs/coco_comparison/000000373705/Ours.jpg} \\

        \includegraphics[width=0.14\textwidth]{figs/coco_comparison/000000497599/gray.jpg} &
        \includegraphics[width=0.14\textwidth]{figs/coco_comparison/000000497599/BigColor.jpg} &
        \includegraphics[width=0.14\textwidth]{figs/coco_comparison/000000497599/GCP.jpg} &
        \includegraphics[width=0.14\textwidth]{figs/coco_comparison/000000497599/unicolor.jpg} &
        \includegraphics[width=0.14\textwidth]{figs/coco_comparison/000000497599/DISCO.jpg} &
        \includegraphics[width=0.14\textwidth]{figs/coco_comparison/000000497599/DDColor.jpg} &
        \includegraphics[width=0.14\textwidth]{figs/coco_comparison/000000497599/Ours.jpg} \\

        \includegraphics[width=0.14\textwidth]{figs/coco_comparison/000000502910/gray.jpg} &
        \includegraphics[width=0.14\textwidth]{figs/coco_comparison/000000502910/BigColor.jpg} &
        \includegraphics[width=0.14\textwidth]{figs/coco_comparison/000000502910/GCP.jpg} &
        \includegraphics[width=0.14\textwidth]{figs/coco_comparison/000000502910/unicolor.jpg} &
        \includegraphics[width=0.14\textwidth]{figs/coco_comparison/000000502910/DISCO.jpg} &
        \includegraphics[width=0.14\textwidth]{figs/coco_comparison/000000502910/DDColor.jpg} &
        \includegraphics[width=0.14\textwidth]{figs/coco_comparison/000000502910/Ours.jpg} \\
        \multicolumn{1}{c}{\small\scalefont{0.7}Input} &
        \multicolumn{1}{c}{\small\scalefont{0.7}BigColor~\cite{Kim2022BigColor}} &
        \multicolumn{1}{c}{\small\scalefont{0.7}GCP~\cite{wu2021towards}} &
        \multicolumn{1}{c}{\small\scalefont{0.7}Unicolor~\cite{huang2022unicolor}} &
        \multicolumn{1}{c}{\small\scalefont{0.7}DISCO~\cite{XiaHWW22}} &
        \multicolumn{1}{c}{\small\scalefont{0.7}DDColor~\cite{kang2022ddcolor}} &
        \multicolumn{1}{c}{\small\scalefont{0.7}Ours}
        
    \end{tabular}
    \vspace{-0.1in}
    \caption{\textbf{More qualitative comparisons.} Our method can generate more natural and photo-realistic colorization results than baselines. All test images are from the COCO-stuff validation set. Zoom in for a better view.}
    \vspace{-0.1in}
    \label{fig_supp_compare}
\end{figure*}

